%% file: main_CVPR_camera_ready.tex
\documentclass[10pt,twocolumn,letterpaper]{article}

\usepackage{cvpr}              

\usepackage{multirow} 
\usepackage{code_style}
\usepackage[autostyle, english=american]{csquotes}
\MakeOuterQuote{"}
\definecolor{cvprblue}{rgb}{0.21,0.49,0.74}
\usepackage[pagebackref,breaklinks,colorlinks,allcolors=cvprblue]{hyperref}

\def\paperID{43000} 
\def\confName{CVPR}
\def\confYear{2026}

\title{Circuit Mechanisms for Spatial Relation Generation in Diffusion Transformers}

\author{Binxu Wang\thanks{equal contribution}\\
Kempner Institute, Harvard University\\
{\tt\small binxu\_wang@hms.harvard.edu}
\and
Jingxuan Fan\footnotemark[1]\\
Harvard University\\
{\tt\small jfan@g.harvard.edu}
\and
Xu Pan\footnotemark[1]\\
Harvard University\\
{\tt\small xupan@fas.harvard.edu}
}

\begin{document}
\maketitle

\begin{abstract}

Diffusion Transformers (DiTs) have greatly advanced text-to-image generation, but models still struggle to generate the correct spatial relations between objects as specified in the text prompt. In this study, we adopt a mechanistic interpretability approach to investigate how a DiT can generate correct spatial relations between objects. We train, from scratch, DiTs of different sizes with different text encoders to learn to generate images containing two objects whose attributes and spatial relations are specified in the text prompt. We find that, although all the models can learn this task to near-perfect accuracy, the underlying mechanisms differ drastically depending on the choice of text encoder. When using random text embeddings, we find that the spatial-relation information is passed to image tokens through a two-stage circuit, involving two cross-attention heads that separately read the spatial relation and single-object attributes in the text prompt. When using a pretrained text encoder (T5), we find that the DiT uses a different circuit that leverages information fusion in the text tokens, reading spatial-relation and single-object information together from a single text token. We further show that, although the in-domain performance is similar for the two settings, their robustness to out-of-domain perturbations differs, potentially suggesting the difficulty of generating correct relations in real-world scenarios.

\end{abstract}

\section{Introduction} 



Diffusion and flow model \citep{dickstein2015noneq,Dhariwal2021Diffusion,ho2020DDPM,Lipman2023Flow,Albergo2023Stochastic} has been leading the charge in generative modeling across many domains, including image, video, and shape\cite{Ho2022video}, etc. 
Specifically, conditional diffusion transformers (DiT) for text-to-image generation (T2I) have unleashed enormous creativity in both industry and the research community, enabling high-fidelity, diverse image synthesis from natural language prompts \cite{rombach2022latentdiff}. 
However, many current T2I models often fail to follow prompts when composing multiple objects onto a scene \citep{Conwell2022Testing}, particularly in arranging their spatial relations \citep{huang2023t2icompbench,ghosh2023genevalobjectfocusedframeworkevaluating,huang2025t2icompbench++}. While the field is fast advancing in generating accurate attributes for single objects, the improvement of generating correct relations between objects is comparatively slow (Fig.~\ref{fig:eval_T2I}).
Increasing attention has been drawn to this problem, and multiple remedies have been proposed recently, including layout conditioning, cross attention guidance, curriculum learning and finetuning with domain-specific data (\cite{li2023gligenopensetgroundedtexttoimage,
chefer2023attendandexciteattentionbasedsemanticguidance,
chatterjee2024gettingrightimprovingspatial,
han2025progressive}). 
However, few work has approached this problem by understanding the underlying circuit for correct composition of multiple objects. 
Inspired by an emerging research field, named "mechanistic interpretability", that reverse-engineer a model’s internal computations to identify how neurons, attention heads, and weights implement algorithms and produce specific outputs, we study this relation generation problem in a mechanistic fashion, with a goal to understand how T2I models can generate spatial relations under different architectural and hyperparameter configurations, and under what conditions they could fail.

To study this problem in a controlled setting, we construct a text-to-image task and train T2I models from scratch. The task is to generate two objects (with compositional shape and color feature) in the scene with one of eight spatial relations specified in the text. Then we delve into the underlying transformer circuits to achieve this task, and find the actual circuits used to solve this task heavily depend on the choice of text encoders. With random token embedding, the T2I model implements a two-stage circuits with two specialized cross-attention heads for reading relation information and single object information respectively. With T5 text encoder, which produces token embeddings that are semantically mixed, the T2I model indeed reads all the relation and single-object information from a single token. We justify the circuits we found by both ablation and causal manipulation. We further find that though the two circuits mechanism achieves similar task accuracy, their robustness is different upon small perturbation in the text prompt. The accuracy with T5 encoder collapses after perturbing by adding an extra filler token in the prompt.

Our study sheds light on several previously unresolved questions: 
1) It was unclear how neural networks encode and use non-commutative relations between objects \citep{Wattenberg2024Relational}. Our work reveals a concrete circuit in diffusion transformers that image tokens can read and implement the relational information in the text, offering a mechanistic example that may generalize to other relational reasoning tasks.
2) The iterative nature of sampling has been an obstacle that complicates attention map analysis and circuit finding in diffusion models. We provide a systematic approach to summarize attention maps and pinpoint heads underlying certain communication patterns, which could be adapted as a general tool to study DiT. 
3) Previous studies attributed the spatial relation generation failure to particular stages, e.g. cross attention (\cite{chefer2023attendandexciteattentionbasedsemanticguidance,phung2023groundedtexttoimagesynthesisattention}) or text encoding (\cite{zhang2024compassenhancingspatialunderstanding,kang2025clipidealnofix}). In this study, we offer a holistic view that bridges these threads. In our toy setting, the T5-based DiT relies on the information fusion by T5 for spatial relationships while the RTE-based DiT implements its own circuits for generating relations. This suggests that the embedding model could be the bottleneck for generating spatial relations in real-world scenarios, making embedding model improvements potentially more critical than DiT modifications.





\begin{figure}[!htbp]
  \centering
  \vspace{-2pt}
  \includegraphics[width=\linewidth]{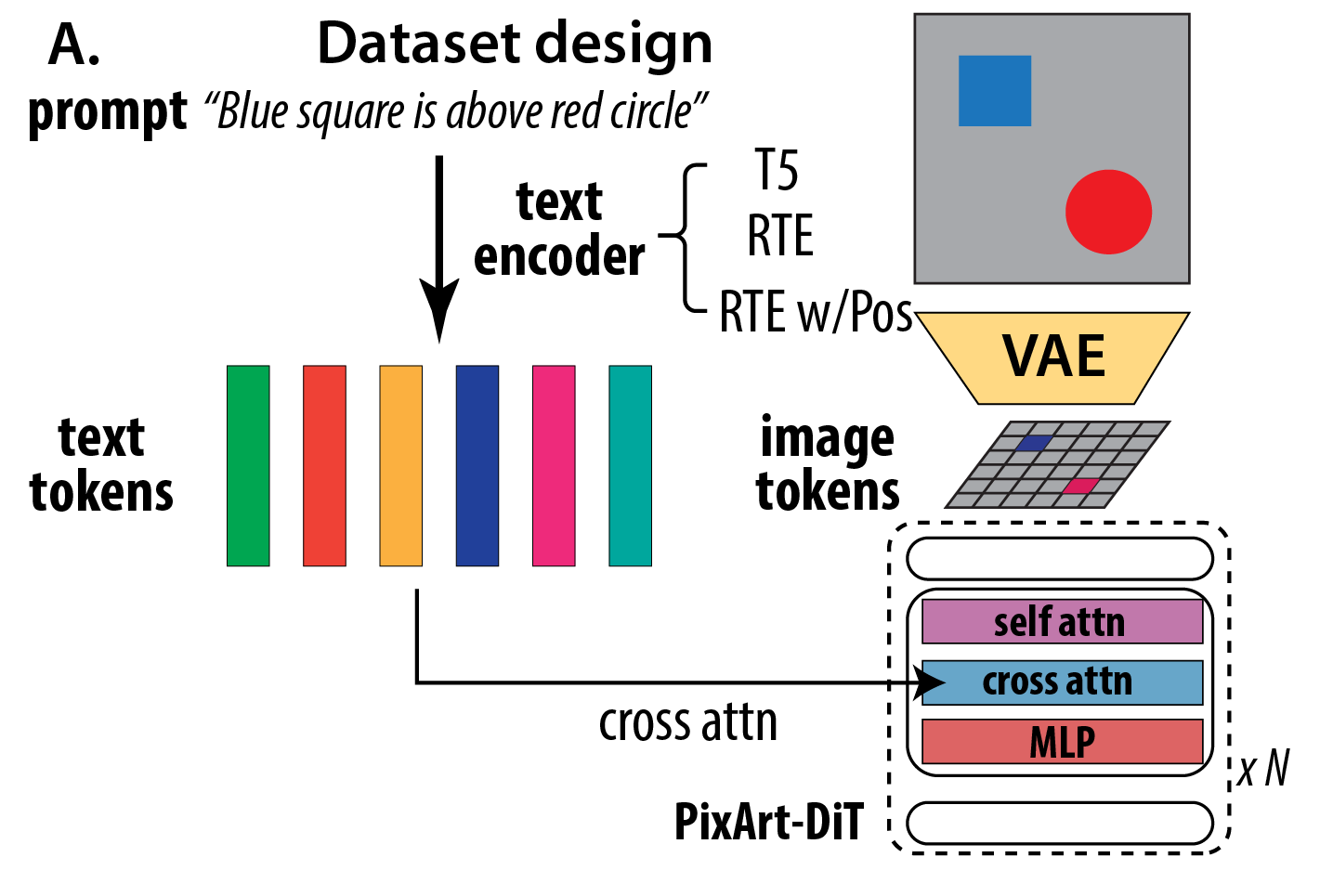}
  \caption{\textbf{Schematics of the model and task}. Our T2I model architecture adopted the design of PixArt \cite{chen2023PixArtAlpha}. There are three main components: the \textbf{text encoder} that processes tokenized natural language prompts into text embeddings, the \textbf{VAE} that processes image inputs into image tokens, and the \textbf{Diffusion Transformer (DiT)} which is the backbone of the denoising diffusion process. The text information routes through the cross attention mechanism in each DiT block and influence the denoising of image tokens. The task is to generate two objects with a specified spatial relation.}
  \label{fig:diagram}
  \vspace{-6pt}
\end{figure}

\section{Background}
\paragraph{Spatial generation failure in T2I models}
Failure in multi-object spatial relation generation has been widely reported in T2I models. While new models improved significantly in generating accurate single object attributes, the improvement of generating correct relations between objects is less impressive (Fig.~\ref{fig:eval_T2I}). One common view is that spatially localized cross-attention grounds object placement. Building on this hypothesis, recent work tackles spatial-relation failures by directly manipulating attention at inference. \cite{chefer2023attendandexciteattentionbasedsemanticguidance} proposes to specifically enhance the cross-attention to subject tokens to prevent catastrophic neglect and improve attribute binding; \cite{phung2023groundedtexttoimagesynthesisattention} optimizes cross- and self-attention using layout-derived losses—via user boxes or LLM-proposed layouts—to enforce multi-object placement and spatial relations. On the other hand, works like \cite{zhang2024compassenhancingspatialunderstanding} and \cite{kang2025clipidealnofix} argue that the poor spatial performance of T2I models stems from the limitations of text encoders. \cite{zhang2024compassenhancingspatialunderstanding} finds that text encoders used across frontier T2I models do not sufficiently preserve spatial relations information in their encoding. Similarly, \cite{kang2025clipidealnofix} argues that multiple image properties including spatial relations cannot be simultaneously encoded by CLIP. Our work is unique in that we unite these two views: we show that different information encoding from text-encoders contribute to distinctive attention patterns and circuits.

\paragraph{Interpretability in diffusion}

The work that shares the most similar interest with us is \citep{Okawa2024Compositional,Park2024Emergence}, which studies the learning dynamics of composition of attributes on a single object in a conditional diffusion model using a minimalist dataset. The key difference is that we focus on the generation of a composition of multiple objects on the scene instead of a single object. Moreover, we study the architecture where the conditioning signal is encoded by multiple word vectors (e.g. T5 encoder) and passed to image tokens via cross attention, instead of a single vector summing all attributes as in \cite{Okawa2024Compositional}. Our setup is more closely related to modern T2I frameworks \citep{chen2023PixArtAlpha,Rombach2022LatentDiffusion,Xie2024SANA}.



\section{Train T2I models to generate spatial relation}

To understand the T2I model's spatial relationship generation mechanism and failure in a controlled manner, we construct a minimal text-image dataset and train DiT-based T2I models of different sizes and text encoders from scratch. We make sure that both single object features and object relation properties are reliably learned and amenable to mechanistic analysis. 
\paragraph{Dataset setup}
We reason that such a dataset should have following properties:
1) multiple objects in the scene with distinct features,
2) objects arranged to satisfy specific spatial relations described by the prompt, and
3) samples are simple enough to be evaluated rigorously. Guided by this principle, we design a dataset of the following format: each sample consists of a prompt in the format \texttt{[descriptor A] [object A] [relation] [descriptor B] [object B]}, e.g., "\textit{red square above and to the left of blue circle}", and a corresponding image with two objects positioned on a gray background (Fig.~\ref{fig:diagram}\textbf{A}). 
We use three shapes (\textit{circle, triangle, square}), two colors (\textit{red, blue}), and eight spatial relations: \textit{left, right, above, below, upper left, upper right, lower left, lower right}. 
The shape and color of the two objects are always distinct, and their positions are arranged to avoid collisions. 
The color descriptors A and B are randomly dropped, and spatial relations are described with multiple paraphrases to add variability (details in App.~\ref{method:dataset}). 


\begin{figure}[!htbp]
  \centering
  \vspace{-5pt}
  \includegraphics[width=\linewidth]{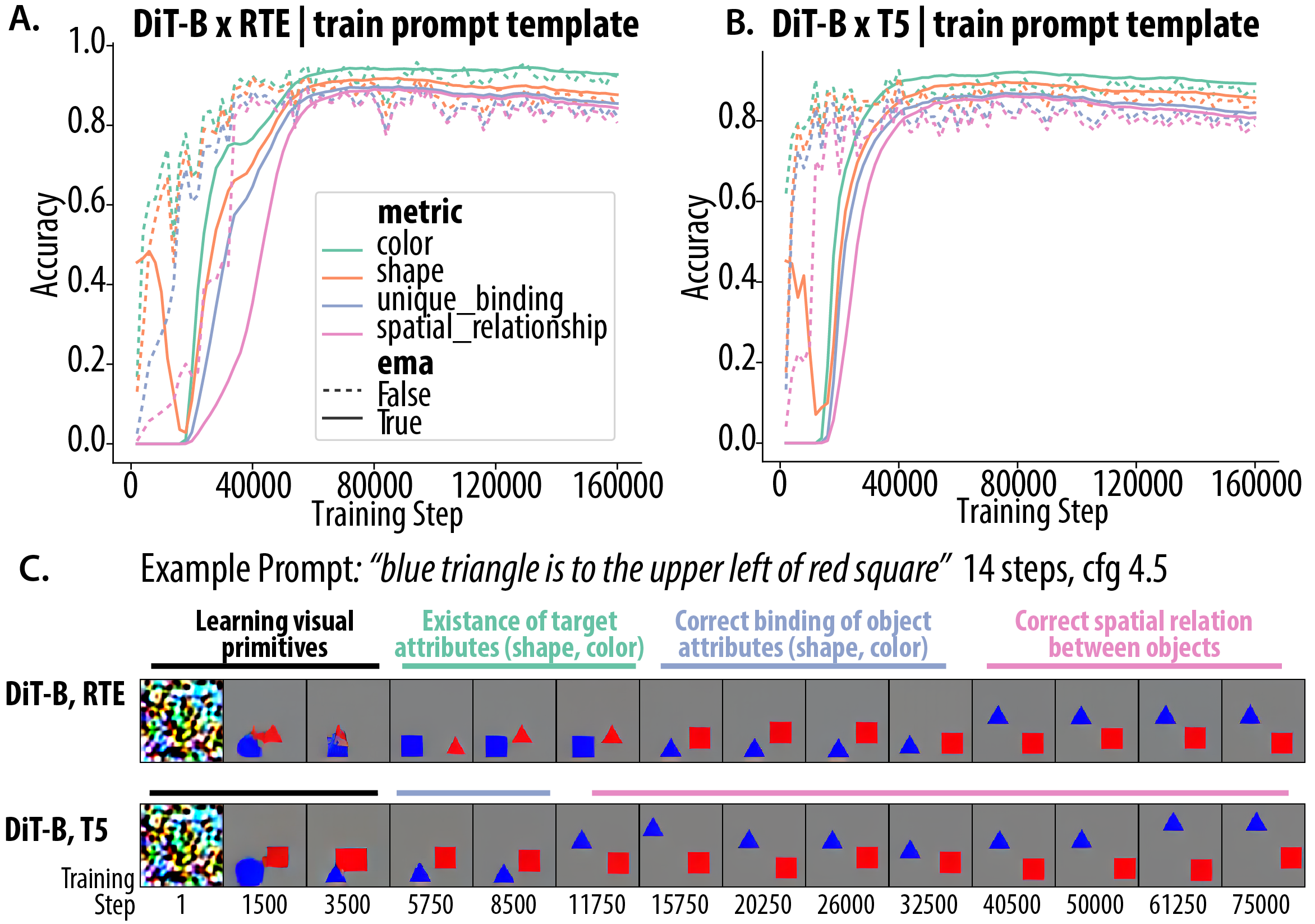}
  \vspace{-10pt}
  \caption{\textbf{Training dynamics of the T2I models (DiT-B)}. \textbf{A.} and \textbf{B.} Both models trained with random token embedding (RTE) and T5 can achieve good accuracy on the task. Solid lines shows the result of model using exponential moving averaged (\texttt{ema}) weights, while dashed line shows the non-averaged weights. 
  \textbf{C.} The task is learned in distinct stages. In both models, they first learn to generate objects but with wrong attributes binding, then they the correct binding of single-object attributes (e.g. red square), finally they learn the correct spatial relation.
  (see App.~\ref{app:eval_learn_dynamics} for other models). 
  }
  \label{fig:dynamics}
  \vspace{-5pt}
\end{figure}
\paragraph{Model architecture}
We use a DiT-based T2I model, following a PixArt-style architecture representative of the state-of-the-art open-source DiT models \citep{chen2023PixArtAlpha}. We train several model sizes with patch size 2: DiT-B (12 layers, 12 heads, 768 latent dimensions), mini (6L, 6H, 384-d), micro (6L, 3H, 192-d), and nano (3L, 3H, 192-d). Following common practice, images are encoded with the VAE pretrained from Stable Diffusion \citep{Rombach2022LatentDiffusion}. 
As for text conditioning, we compare three encoders: (i) T5-XXL \citep{Raffel2023T5Model}; (ii) a random token encoder with sinusoidal positional encoding (RTE); 
(iii) RTE without positional encoding. 
The RTE maps each token ID to a fixed Gaussian-sampled vector whose dimension and norm match those of T5 embeddings. 
This comparison tests whether the diffusion transformer can learn object relations without semantic or contextual structure in the text embeddings, enabling us to localize the model’s relational computations (Fig.~\ref{fig:diagram}\textbf{A}). (details in App.~\ref{method:model_arch}). 

\paragraph{Sampling and evaluation}
Throughout training, we evaluate model generations on 96 prompts spanning 8 spatial relations and 12 object pairs. Generation is performed on multiple random seeds with the default sampler (DPM-Solver++ with 14 steps \citep{lu2022dpm++}) and classifier-free guidance of 4.5 \citep{ho2022classifierFreeGuidance}. We evaluate the consistency of generated images with the prompts using classic segmentation and classification tools from \texttt{cv2} \cite{bradski2000opencv}. 
Specifically, we evaluate four aspects of samples: 1) existence of correct colors on the image, denoted as \texttt{color}, 2) existence of correct shapes on the image, denoted as \texttt{shape} 3) correctness of shape color binding on the two objects, denoted as \texttt{unique\_binding} and 4) correctness of spatial relation between the identified two objects, denoted as \texttt{spatial\_relation}. 
Each metric produces a binary (0,1) score for each sample, which we then average to obtain the model’s accuracy.

\paragraph{Model comparison} 
Accuracy on all four metrics increases with model parameter sizes up to the DiT-mini configuration; accuracy gains from DiT-mini to DiT-B are marginal (Tab. \ref{tab:model_cmp_eval_table}). All trained models at the largest parameter size (DiT-B) show high accuracy in \texttt{color} and \texttt{shape}, but the \texttt{unique\_binding} and\texttt{spatial\_relation} accuracy varies significantly depending on the chosen text encoders (Tab. \ref{tab:model_cmp_eval_table}). 
Both RTE and T5 achieve strong \texttt{unique\_binding} and \texttt{spatial\_relation} accuracy, indicating that \textbf{pretrained semantic structure (T5) is not strictly required for learning object relations}. 
However, RTE without positional encoding is significantly worse on these metrics. 
Since text enters only through the cross attention mechanism, without positional cues, the DiT output is invariant to permutation of text embeddings. For instance, "\textit{red A on top of blue B}" and "\textit{blue B on top of red A}" collapse to the same bag of words, yielding identical outputs. Adding positional information resolves this ambiguity.
\paragraph{Training dynamics}
Having established the end-point performance trends across models, we next examine how these capabilities emerge during training. For both T5-DiT and RTE-DiT models, we evaluate models with the exponential moving average (EMA) weights \citep{Karras2024EDM2Analyzing} following common practice and the accuracy curves averaged across multiple runs are showed. We consistently observe that \texttt{color} accuracy converges first, followed by \texttt{shape} and then \texttt{unique\_binding}. \texttt{spatial\_relation} is learned the slowest (Fig.~\ref{fig:dynamics}\textbf{A.B.}), indicating that relational composition is more challenging to learn than single-object attributes or bindings. Comparing across the two text encoders, we observe that T5-DiT models converge to the final accuracy faster across all four evaluation metrics. Moreover, the temporal gaps between the convergence of different metrics are tighter. We also provide generation examples at different training steps for visual examination (Fig.~\ref{fig:dynamics}\textbf{C.}). The different learning dynamics provide the first hint that the two family of models potentially use different internal mechanisms to accomplish the same generation task. 

\begin{figure*}[!hbtp]
  \centering
  \vspace{-15pt}
  \includegraphics[width=0.99\linewidth]{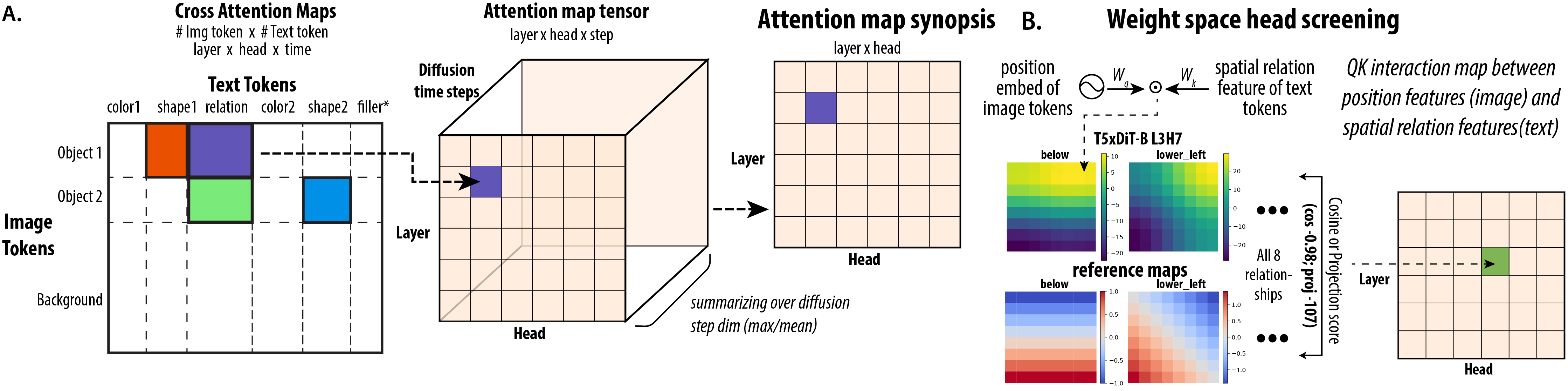}
  \vspace{-8pt}
  \caption{\textbf{Illustration of our methods to find relevant heads} \textbf{A.} \textbf{Attention synopsis}: The giant attention tensor is first reduced to those only between two interested groups of tokens (e.g. the relation token regardless specific words, or an object token regardless where or what it is). Then the reduced attention tensor is averaged over diffusion time steps, resulting in a layer $\times$ head map which we use to pinpoint relevant heads. 
  \textbf{B.} \textbf{Weight space head screening}: We also develop an efficient weight-space screening method to screen for relation heads: computing the QK interaction between image position features and text relation feature, and scoring them by how well these spatial maps align with the reference maps corresponding to the relation.
  }
  \label{fig:synopsis}
  \vspace{-2pt}
\end{figure*}

\section{Relation generation circuits in RTE-DiT}

\subsection{Attention synopsis}
In DiT-based architecture, cross-attention provides the only mechanism by which text prompts can modulate the image tokens during each denoising step. Therefore, we examine the cross-attention patterns to gain insights into how text on single object feature and spatial relations guide correct generations. Given the large number of DiT's cross attention maps ([layers $\times$ heads $\times$ time steps $\times$ condition vs uncondition pass $\times$ number of tokens]), it is impractical to visual inspect them individually. Moreover, simply averaging attention maps over different samples and prompts can obscure specific interactions. Therefore, we develop a scalable paradigm to analyze and quantitatively summarize the cross-attention head patterns, which we term \textit{Attention Synopsis} (Fig.~\ref{fig:synopsis})
Equipped with this method, we efficiently search through over 10 million attention maps to trace stable text-to-image information flow and localize the relevant circuit mechanisms for spatial relations generation. Specifically, we leverage the fact that token categories are identifiable in both image and text (image tokens by object segmentation, text tokens by semantic attribute). We then aggregate attention within categories, yielding interpretable category-to-category interaction patterns. After this aggregation, we reduce the cross-attention map tensor dimension to [num layer $\times$ num head $\times$ num time steps]. 
Observing that the attention maps usually change smoothly across time, we further calculate the mean attention maps over time steps, reducing the tensor to shape \texttt{[num layer, num head]}, which we denote as the \textit{attention map synopsis}. Each entry denotes the mean \textit{attention transmission energy} between a specific text–image token category pair (details in App.~\ref{method:attention_map_synopsis}). 

Many previous works have reported the cross attention communication between the text token of a single object and the corresponding object in the image \citep{tang2022daam}, and this property is leveraged to control generation \citep{hertz2022crossAttnCtrl,liu_towards_2024}. These findings suggest that there is coupling between single object tokens across text and image modality, supporting faithful generation. However, less is known about whether text tokens describing \textit{spatial relations between} objects exhibit similar properties. Therefore, we leverage the \textit{Attention Synopsis} method to examine all category-to-category cross-attention patterns, especially focusing on the spatial relations category. We show results for RTE-DiT in this section and T5-DiT in Section \ref{sec:relation_t5}.

In RTE-DiT, we find a minimal circuit that enables generation of correct objects at correct spatial locations. The circuit consists of two key cross attention heads which we discuss in details below. 

\begin{figure*}[!htbp]
  \centering
  \includegraphics[width=0.95\linewidth]{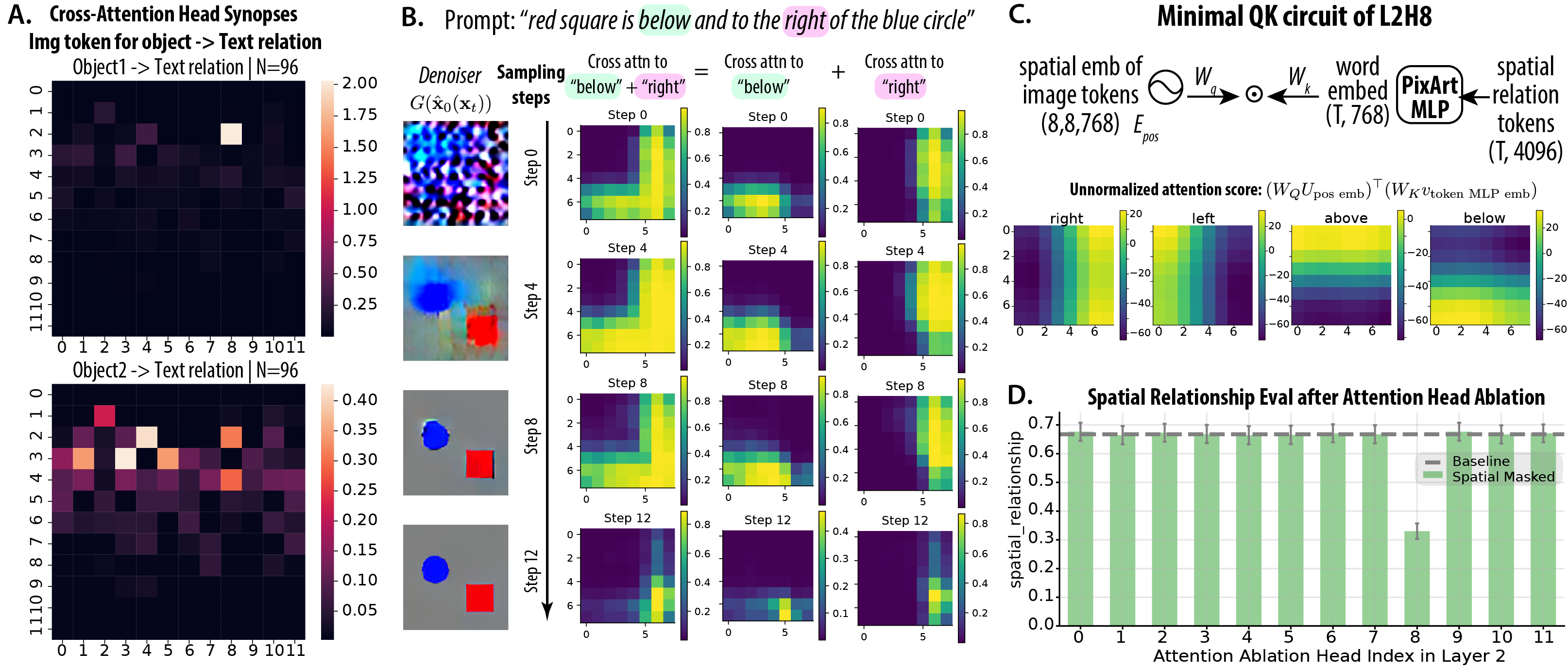}
  \caption{\textbf{The spatial relation heads in random-embedding-based DiT}. \textbf{A.} We find specialized cross attention heads that contributes to the object image tokens (top: the object1 in the text; bottom: the object2 in the text) attending to the relation text tokens. \textbf{B.} We show the activation of this head across images tokens and sampling steps. The map for the composite relation “below and right” decomposes cleanly as the sum of the maps for “below” and “right”, \textbf{C.} The observed attention patterns can be induced by QK interaction of positional embedding and text embedding of relation words. \textbf{D.} Ablating relation head (L2H8) specifically affects relation evaluation.}
  \label{fig:relation_head}
\end{figure*}

\subsection{Spatial relation head}

\begin{figure*}[!htbp]
  \centering
\vspace{-15pt}
\includegraphics[width=0.95\linewidth]{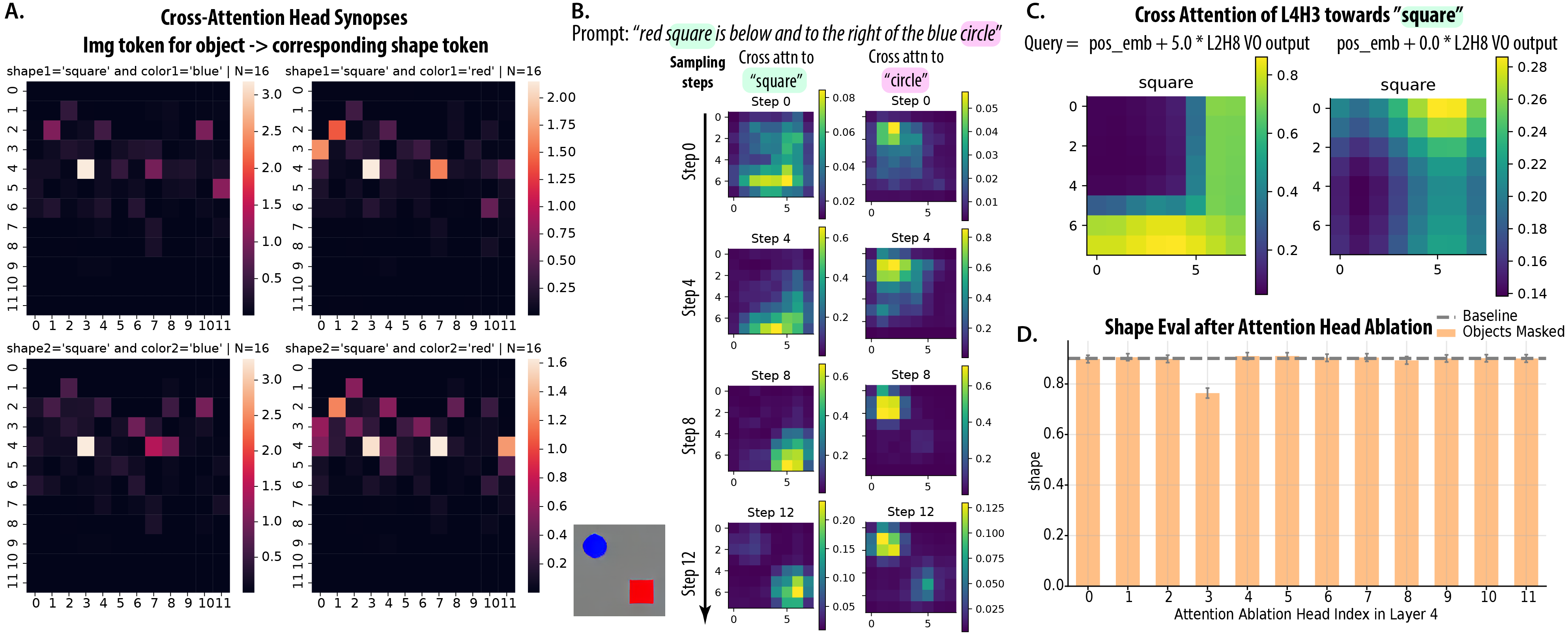}
  \caption{\textbf{The object generation heads in random-embedding-based DiT}. \textbf{A.} Specialized heads emerge in cross-attention synopses, by summarizing strength from image tokens of each object to its own shape token. \textbf{B.} Activation of this head (L4H3) across images tokens and sampling steps for the prompt “\textit{red square is below and to the right of the blue circle}”: tokens at the eventual square location attend to “square,” while the other object attends to “circle”; spatial specificity sharpens from Step 0→12. \textbf{C.} Injecting the VO output of the relation head (L2H8) into positional embeddings is sufficient to elicit selective attention from tagged locations to the “square” token (left); without the "tag", the pattern is weak (right). This indicates the object generation head reads the relational tag generated by the spatial relation head. \textbf{D.} Ablating object head (L4H3) specifically affects shape evaluation.}
  \label{fig:object_head}
\end{figure*}





We first use attention synopses to look for cross attention heads that pass information from relational text tokens to image tokens. We find, indeed, there are specialized heads for this transmission. In the case of ``object1'' (i.e. the first object in the text), there is only one head (L2H8) that dominates, while in the case of ``object2'', there are a small number of heads that have this role (Fig.~\ref{fig:relation_head}A). We also find a similar pattern in models of other sizes (App.~\ref{app_sec:relation_head}). 
We name these heads ``\textbf{spatial relation head}''. Examining their attention score towards relation words across sampling steps (Fig.~\ref{fig:relation_head}B), we find that they activate immediately (at step 0), before any image information is available. Their activation patterns align with the spatial region associated with the relation token and progressively concentrate toward the generated object. These observations imply that the query which maximizes the attention score of this head is potentially aligned with the positional embedding. 
We verify this hypothesis by checking the QK circuit \cite{elhage2021mathematical} of L2H8 (Fig.~\ref{fig:relation_head}C). The head transforms sinusoidal positional embeddings from image tokens into the query space ($Q$) and MLP-transformed relation-word embeddings\footnote{In the PixArt architecture, frozen text embeddings are first passed through a learnable MLP projection before entering the attention layers.} into the key space ($K$) via learned linear layers ($W_q$, $W_k$).
This $Q, K$ transformation aligns specific coordinates in the image grid with the semantics of spatial relation tokens. The resulting inner-product maps (Fig.~\ref{fig:relation_head}\textbf{C}) form smooth gradients whose orientation reflects the spatial relations (e.g., "above" produces a vertical gradient, with highest energy on top). These gradients act as positional tags, marking the regions of the canvas where the first object should be placed. Downstream heads then read these tags to guide accurate object placement and generation. Notably, this mechanism is strikingly similar to the molecular gradients that direct embryonic cells into specific organs, just as tokens "develop" into the proper visual features.

\subsection{Object Generation Head}
The spatial relation head allows differential tagging of image tokens based on relational text tokens. To successfully generate the composite scene, the model also needs to generate the correct object on the tagged canvas. To this purpose, we examine cross attention heads that pass object shape information from text tokens to image tokens. We identify a prominently active head—Layer 4, Head 3 (L4H3) that consistently mediates communication between an object’s image tokens and its corresponding shape word in the prompt (Fig.~\ref{fig:object_head}\textbf{A}). This linkage is invariant to both the object’s position in the sentence (Shape 1 or 2) and the specified spatial relation, indicating that the head transmits shape identity independently of relational context (Fig.~\ref{fig:object_head}\textbf{A}). 
During sampling, in contrast to \textbf{spatial relation head}, this communication channel is active later in sampling (step 4-8), linking each objects to their corresponding shape tokens in the prompt (Fig.~\ref{fig:object_head}\textbf{B}).  

\subsection{Ablation and causal manipulation} 
To test whether the heads discussed above have a causal role in correct spatial relation and object shape generation, we perform both ablation and casual manipulation experiments.  

First, we perform head-specific ablation of cross-attention. Specifically, we remove the contribution from a specific cross-attention head on either the spatial relation (Fig.~\ref{fig:relation_head}\textbf{D}) or object (Fig.~\ref{fig:object_head}\textbf{D}) text tokens. Ablating spatial-relation attention specifically in L2H8 reduces relational accuracy from 67\% to 33\%, while other heads show negligible effects (Fig.~\ref{fig:relation_head}\textbf{D}). 
This confirms L2H8’s critical role in implementing the correct spatial layout. On the other hand, ablating cross-attention to the object text in specifically L4H3 decreases the object shape generation accuracy from 90\% to 76\%, while the ablation of other layer head combinations shows minimal effect (Fig.~\ref{fig:object_head}\textbf{D}). This emphasized the critical value of such a head in robustly generating the correct object shape. Although the effect size of object shape ablation is smaller than that of spatial relations ablation, the effects in both cases are confined only to the two previously identified heads, L2H8 and L4H3, suggesting a highly concentrated circuit.

We reason that L2H8 and L4H3 functions in sequence to generate a correct object at the correct spatial location. To test this hypothesis, we use the VO output of the relation head (L2H8) along with positional embeddings as the `image tokens' of the object generation head (L4H3). 
This manipulation is sufficient to elicit selective attention from tagged locations to the “square” token (Fig.~\ref{fig:object_head}\textbf{C}). Without the injected VO inputs, no obvious attention pattern is observed (right), showing that the downstream object shape head indeed reads the relational tag. 




\paragraph{Consistency across model sizes} 
Finally, we assess whether this circuit mechanism generalizes across RTE-DiT models of different sizes (App.~\ref{app_sec:relation_head}). 
We find the spatial relation heads consistently emerge in DiT of three scales (DiT-B, mini, micro). 
In contrast, the smallest model (DiT-nano) shows no such head and performs poorly on spatial relation (accuracy 5\%) (Tab. \ref{tab:model_cmp_eval_table}). 

\begin{figure}[!htbp]
\vspace{-8pt}
  \centering
\includegraphics[width=0.9\linewidth]{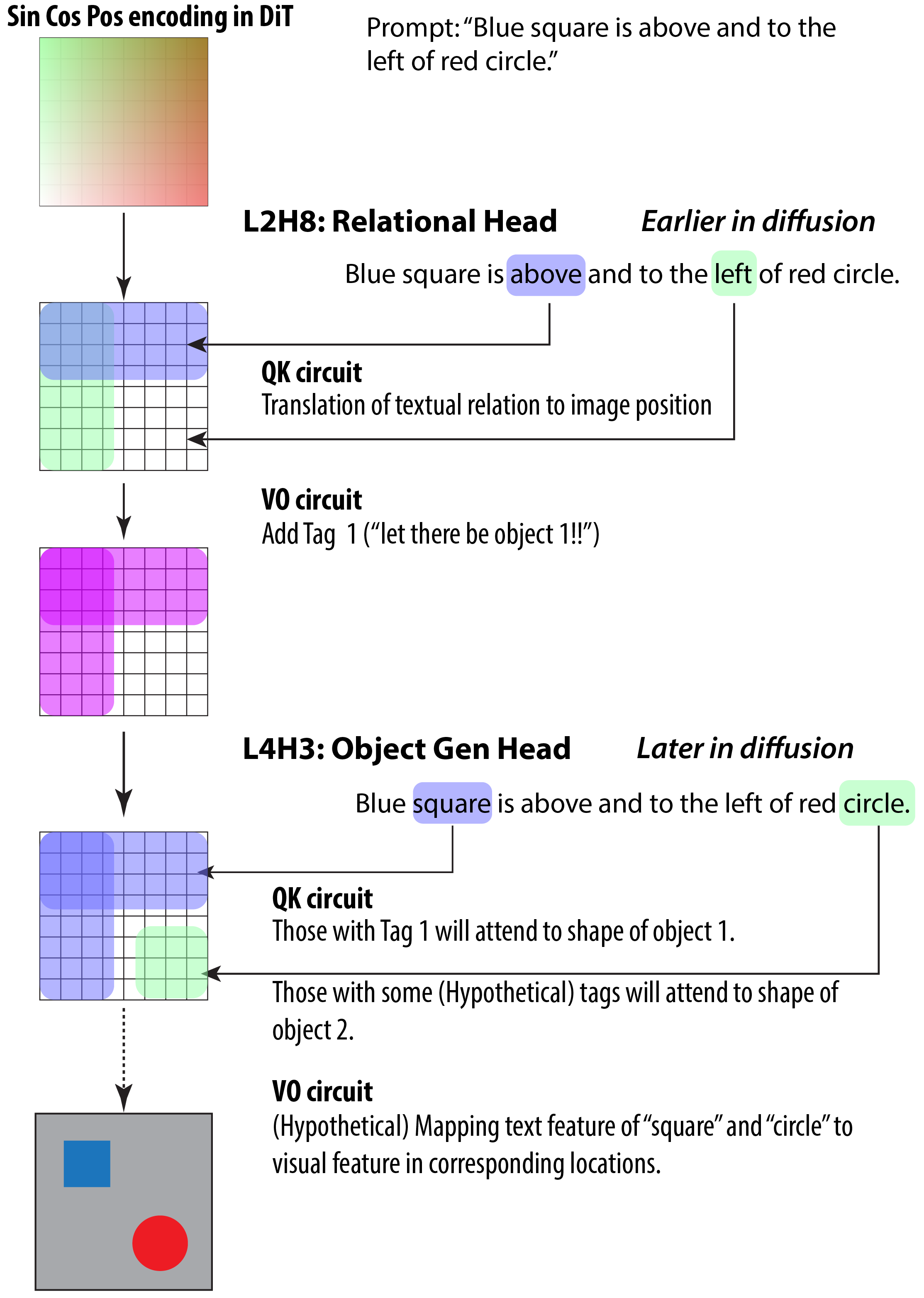}
\vspace{-5pt}
  \caption{\textbf{Schematics of the object relation circuit in DiT trained with random embedding}}
  \label{fig:summary}
\vspace{-5pt}
\end{figure}

\paragraph{Summary}
Taking it together, for DiTs trained with random token embedding (RTE), relational object generation unfolds in two stages (Fig.~\ref{fig:summary}): 
the \textbf{``spatial relation head''} reads relational text tokens (e.g., "above," "left") via the QK circuit and interact with the sinusoidal positional encoding of image tokens, producing spatial gradients for each relation. 
The VO circuit writes positional tags (e.g., Tag~1) onto image tokens, marking where the object (e.g. 1st in sentence) should appear.
In the \textbf{``object generation head''}, tokens with matching tags attend to shape token of the corresponding object. 
The VO circuit maps these text features (e.g., "square", "circle") into visual features at tagged locations, generating the object via denoising.


This modularizes operation where relation heads laying the ground and object heads assigning attributes provides a clean and disentangled mechanism for robust composition of relation and object combination. 

\section{Relation circuits in T5-DiT} 
\label{sec:relation_t5}

\begin{figure*}[!htbp]
  \centering
\vspace{-15pt}
\includegraphics[width=0.99\linewidth]{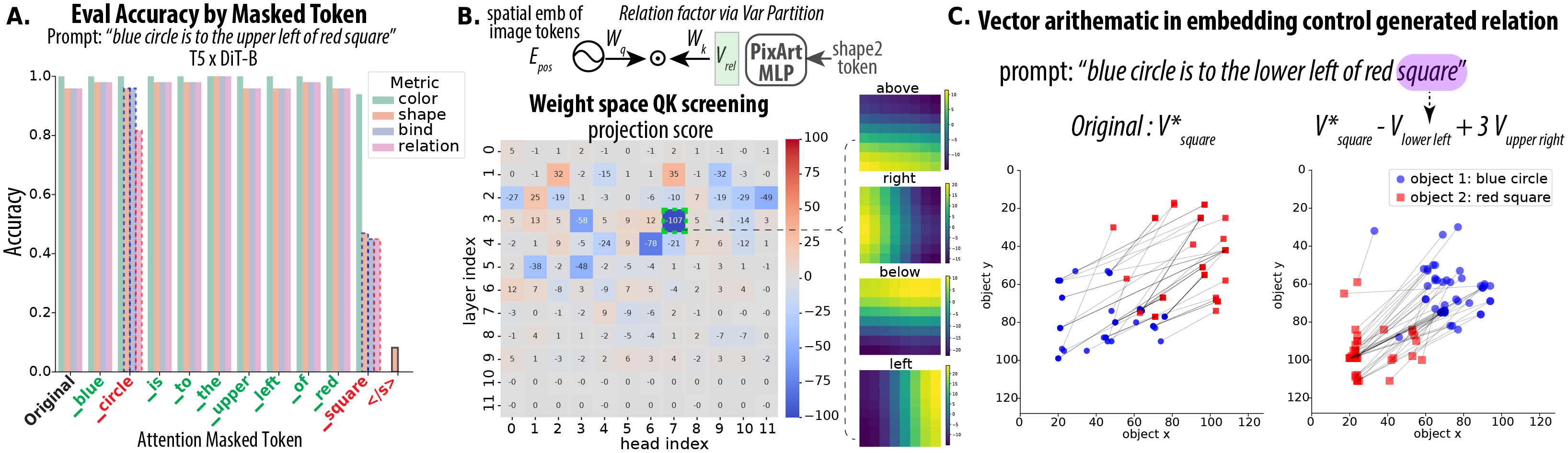}
  \caption{\textbf{Mechanism for relational generation in T5-DiT. }
\textbf{A.} T5-based DiT is robust to attention
ablation of relation word, but most sensitive to \texttt{shape2} and \texttt{EOS}. 
\textbf{B.} Weight space screening for spatial relation heads via projection score, and its corresponding spatial gradients (L3H7). 
\textbf{C.} Vector arithmetic on factorized word embedding causally affects generated object relation.} 
  \label{fig:ablation2}
  \label{fig:causal2}
\end{figure*}




For T5-DiT, applying the same \textit{Attention Synopsis} method reveals no clear pattern for spatial relation heads (App.~\ref{app_sec:T5_DiT_attn_synops}), suggesting a different underlying mechanism.

\paragraph{T5-DiT extracts relations from other tokens}
T5's self-attention allows each text token to integrate information from the entire sentence, so image tokens may derive spatial relation cues from non-relation tokens. To test this, we masked individual text tokens and measured their impact on relational object generation (Fig.~\ref{fig:ablation2}\textbf{A}). 
Surprisingly, the relation words, filler and color words have little effect on the generation performance (Fig.~\ref{fig:ablation2}\textbf{A}). 
Ablating \texttt{<end\_of\_sentence>} disrupts the denoising process and collapses all metrics. Masking \texttt{shape1} reduces relation accuracy by 15\%, while masking \texttt{shape2} lowers all shape, binding, and relation accuracies by 50\%.

This result suggests a different mechanism: T5's self-attention spreads the complete sentence context into many text tokens, so DiT only needs to decode the relevant information from \texttt{shape2} (and \texttt{shape1}) without relying on relation words. 

\paragraph{Structure of relation representation in T5 embedding}
Visualization of the T5 contextual embedding of the second shape token ($V^*_{\text{shape2}}$) using UMAP, tSNE, and PCA reveals clear separation by spatial relation and, within each relation, by object2's color (Fig.~\ref{fig:T5_context_wordvec_UMap}), indicating that both object identities and spatial relation are jointly encoded in this space.

We further dissect this representation by variance partitioning (App.~\ref{method:variance_partitioning}). We model the contextual embedding of the 2nd shape token as a linear combination of four factor vectors:
$V^*_{\text{shape2}}=V_{\text{shape2}}+V_{\text{color2}}+V_{\text{shape1}}+V_{\text{rel}}$.
This analysis enables us to identify a specific vector corresponding to each value of a factor (for example, a vector for "upper right" versus "left"), which in turn allows us to systematically manipulate the embeddings by adding or subtracting these vectors.
Variance partitioning confirms this factorization (Tab.~\ref{tab:varpart_shape2}). 
In the T5 embedding, \texttt{shape2} explains most variance ($\sim$37.5\% partial $R^2$), with \texttt{relation} contributing substantially ($\sim$12\%). After DiT MLP projection, \texttt{relation} becomes dominant ($\sim$21\%), while \texttt{shape2} decreases, suggesting DiTs amplify relation information in the \texttt{shape2} token.

\begin{table}[!h] 
\centering
\scriptsize
\renewcommand{\arraystretch}{0.9}
\begin{tabular}{lcccccc}
\toprule
  & & $V_{\text{shape2}}$ & $V_{\text{color2}}$ & $V_{\text{shape1}}$ & $V_{\text{rel}}$ & tot. $R^2$ \\ 
\midrule
\multirow{2}{*}{T5 emb}
 & part. $R^2$ & 37.5\% & 4.7\% & 5.0\% & 12.1\% & \multirow{2}{*}{73.2\%} \\
 & marg. $R^2$ & 51.4\% & 4.7\% & 18.9\% & 12.1\% &  \\
\midrule
\multirow{2}{*}{DiT MLP}
 & part. $R^2$ & 14.9\% & 8.0\% & 7.2\% & 21.3\% & \multirow{2}{*}{53.4\%} \\
 & marg. $R^2$ & 16.9\% & 8.0\% & 9.0\% & 21.3\% &  \\
\bottomrule
\end{tabular}
\caption{Variance partitioning of T5 embedding and DiT-MLP projection of \texttt{shape2} token.}\label{tab:varpart_shape2}
\vspace{-7pt}
\end{table}

\paragraph{Relation heads produce spatial gradients from the relation factor in \texttt{shape2}}
Given these results, we hypothesized that certain cross-attention heads read from the \texttt{shape2} token and specifically interact with its relation factor. To evaluate this, we calculated the QK interaction between the positional encodings of image tokens and the learned relation factor vectors $V_{rel}$, i.e., $(W_q E_{pos})^\top(W_k V_{rel})$. To be useful for relational object generation, such a head should yield a spatially modulated attention map (e.g. a gradient) that aligns with the semantics of the relation vector. We screened all heads by comparing the resulting QK inner product maps to ideal spatial gradient across the eight relations (Fig.~\ref{fig:synopsis}\textbf{B}, method detailed in App.~\ref{method:head_screening}). Notably, we identified one or two cross-attention heads that produce strong spatial gradients matching the intended relation (L3H7, Fig.~\ref{fig:ablation2}\textbf{B}). 
Compared to the RTE-DiT’s relation head, T5-DiT achieves a similar effect with a variation: the QK interaction generates a spatial gradient using the relation factor embedded in \texttt{shape2}. 
Overall, this approach provides a more efficient, weight-space method for screening potential spatial relation heads via direct hypothesis testing at scale. We applied it to DiTs with RTE, T5 and CLIP encoders and found similar relation heads (\ref{app_sec:rel_head_screening}). 
As a side note, a similar in depth dissection of relation circuit CLIP-DiT is presented in \ref{app_subsec:CLIP_DiT}, yielding comparable results to T5-DiT. 

\paragraph{Causal evidence for relation encoding in object tokens}
Building on the observation that relational information is embedded in the contextual representation of \texttt{shape2}, we directly manipulated the T5 embedding to test its effect on generation. Specifically, we performed targeted vector arithmetic in the 4096-dimensional prompt embedding space: for the embedding of \texttt{shape2} (e.g., $V^*_{\text{square}}$ in the prompt "\textit{blue circle is to the lower left of red square}"), we subtracted the factor vector corresponding to the original relation (e.g., $V_{\text{lower left}}$) and added a scaled vector for an alternative relation (e.g., $3V_{\text{lower right}}$ or $3V_{\text{upper right}}$). As illustrated in Fig.~\ref{fig:causal2}\textbf{C}, this manipulation consistently shifts the generated object positions to reflect the new relation, while preserving object shape and color. These results provide direct, causal evidence that spatial relational geometry is encoded within the embedding of \texttt{shape2}, and that simple linear operations on this embedding suffice to control spatial relationships in generated images.


\paragraph{Robustness to prompt perturbation} 
Although RTE- and T5-based DiTs achieve similar in-distribution spatial-relation accuracy, their different circuitry mechanisms suggest potential different generalization behaviors especially when the text tokens are perturbed. As detailed in \ref{app_sec:robustness}, small prompt changes, such as inserting filler word "the" before the object, sentence reversion, or using color synonym all cause a large drop in T5-DiT relation accuracy, while RTE-DiT remains much more stable (Fig.~\ref{fig:robustness_the}). 
This failure in generalization can be well explained by its relation circuits: since DiT decodes spatial relations from the contextual embedding of \texttt{shape2}, nuisance words introduce perturbations to the \texttt{shape2} representation, which DiT interprets as biases toward certain relations, resulting in incorrect outputs. 
This suggests that the T5-DiT circuit, which decodes both relational and object information from a single contextual word embedding, is brittle and easily perturbed.




\paragraph{Transferability to pre-trained models} 

We tested whether these circuit-analysis tools transfer to a pre-trained model, PixArt-Sigma. As shown in \ref{app_sec:pretrained}, PixArt-Sigma has only weak spatial-relation generation ability: among 30 object pairs, only 8 show nontrivial object and relation accuracy. Nevertheless, token ablation indicates that spatial information is present in both the relation word and object words, which is consistent with our findings on the synthetic dataset. Head screening reveals a small number of heads that are strongly associated with spatial generation (\cref{suppfig:pretrained}). Thus, even if spatial relation performance is limited, the same analysis still identifies sparse, interpretable spatial circuits in a pre-trained model.

\section{Discussion}

We find that Diffusion Transformers can learn to generate scenes with correct object–relation composition when trained on a minimalistic relational dataset, but the underlying mechanisms depend strongly on the choice of text encoder. Both T5 and RTE encoding models achieve high relational accuracy, yet they differ markedly in their attention circuits, their generalization behavior, and their sensitivity to perturbations. 

RTE-DiTs are highly sensitive to relational words ablation yet robust to filler word changes. In contrast, T5-DiTs are robust to relational word ablation but sensitive to small lexical variations, indicating that explicit relation tokens play a limited role in their generation. These differences in generation properties can be linked to their distinctive attention circuit mechanisms. RTE-DiTs use a modularized strategy: a specialized "relational heads" directly read and translate relation tokens into spatial gradients across image tokens, guiding object placement before identity assignment; then a separate object generate head place the objects according to the spatial gradients. T5-DiTs instead use an object-centric strategy: contextual embeddings from T5 absorb relational information into object tokens (especially the last shape words), allowing both spatial layout and single object attribute to be decoded from one or two token embeddings. 
This approach is more compact but less disentangled, and sensitive to small perturbation of the two token embedding, even due to addition of nuisance words. 


These results highlight that text embedding design shapes not only performance but also inductive bias and interpretability. 
RTE enables explicit, object-invariant, disentangled relation-to-location pathways, offering clearer control mechanisms, while T5 embeddings promote integration of relational cues into object features, potentially aiding linguistic generalization but obscuring intermediate relational processing. The trade-off between interpretability, prompt robustness, and reliance on explicit relational tokens should inform future text-to-image model design.

\clearpage

\subsubsection*{Acknowledgments}
We gratefully acknowledge support from the Kempner Research Fellowship (B.W.) and the Schwartz Fellowship (X.P.), as well as computing resources provided by the Kempner Institute cluster. 
We thank Martin Wattenberg, Yonatan Belinkov, and Thomas Fel for their careful reading and valuable feedback on earlier versions of this work. We also appreciate the insightful comments and discussion from participants and reviewers of the New England Mechanistic Interpretability (NEMI) Workshop and the Mechanistic Interpretability Workshop at NeurIPS 2025.



{
    \small
    \bibliographystyle{ieeenat_fullname}
    \bibliography{diffusion_prune}
}


\clearpage
\onecolumn
\appendix
\section{Extended Results}
\subsection{Evaluation and Benchmark}
\begin{figure}[!htp]
    \centering
    \includegraphics[width=0.85\linewidth]{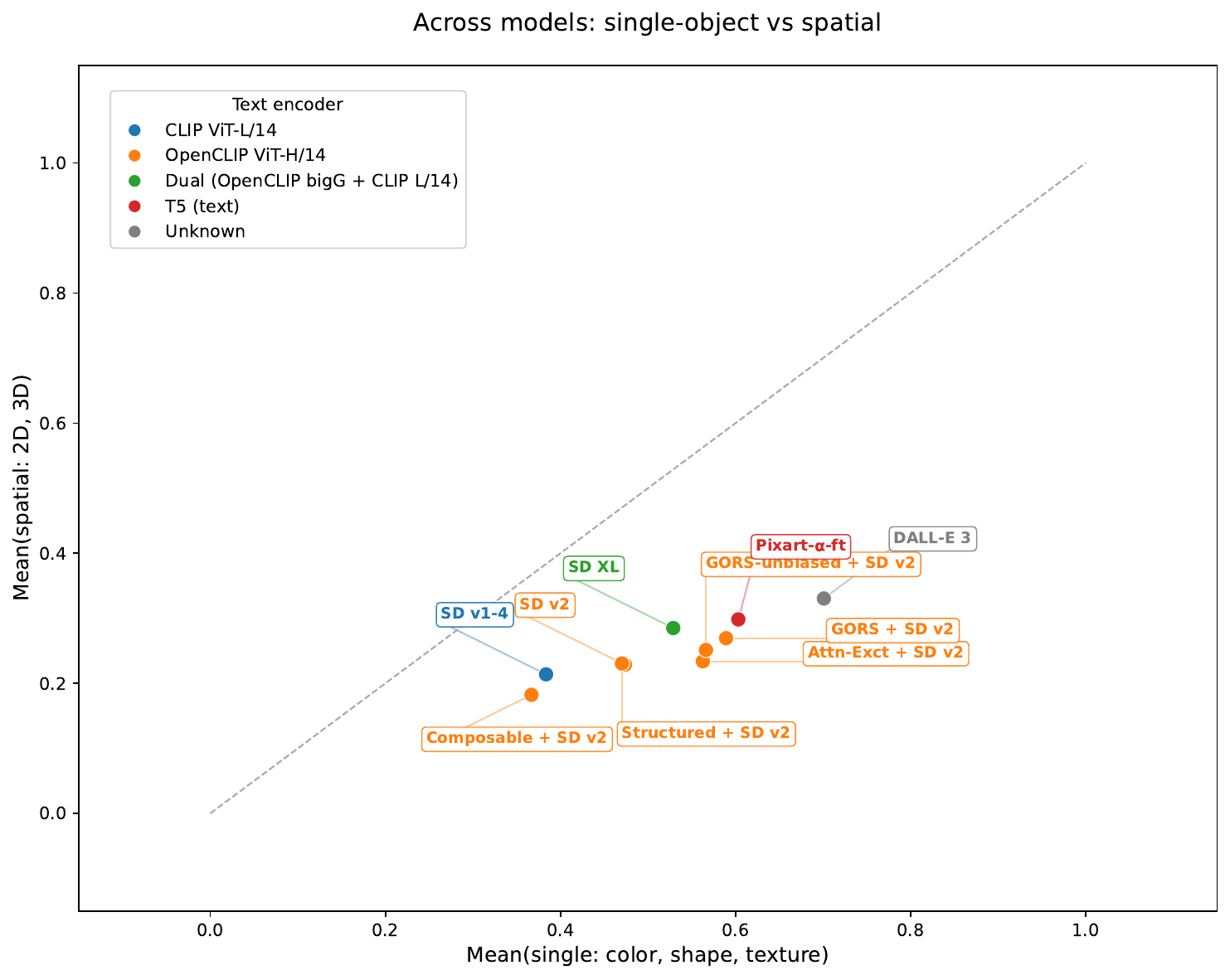}
    \caption{\textbf{Benchmark scores of spatial relationship and object feature attributes of open- and closed- source models.} Color of dots denote the text encoder.}
    \label{fig:eval_T2I}
\end{figure}

\begin{table}[!htp]
    \centering
    \vspace{-25pt}
        \caption{\textbf{Comprehensive evaluation of models on prompt template variations.} \\
         \textbf{Abbreviations}: \textit{WD}: weight decay, \textit{RTE}: random embedding plus position encoding, \textit{RTE w/pos}: random embedding without position encoding. \textit{bind}: unique and correct attribute binding. \textit{sp rel}: spatial relation correctness (loose). \textit{sp rel+}: spatial relation correctness (stringent). \textit{Dx}, \textit{Dy}: difference of coordinates between the two identified objects (with target attributes) $x_1-x_2$, $y_1-y_2$, with the unit pixel (128 pixel total).  \\
         All statistics are averaged from 264 prompts, covering all 8 relations and all object combinations, each drawing 50 samples. 
         Thus, the non-zero value in Dx, Dy suggests systematic bias in spatial relation. Evaluated the checkpoint at 4000 epochs with EMA. }
    \label{tab:model_cmp_eval_table}
    \small
    \begin{tabular}{llrrrrrrr}
\toprule
 &  & shape & color & bind & sp rel & sp rel+ & Dx & Dy \\
model name & template &  &  &  &  &  &  &  \\
\midrule
\multirow[t]{5}{*}{RTE DiT-B} & O1 is Rel O2 & 0.877 & 0.928 & 0.855 & 0.843 & 0.758 & -0.4 & -0.9 \\
 & O1 is Rel the O2 & 0.900 & 0.942 & 0.877 & 0.862 & 0.717 & -0.3 & -0.4 \\
 & O1 Rel O2 & 0.858 & 0.909 & 0.833 & 0.823 & 0.752 & -1.2 & -1.6 \\
 & O1 Rel the O2 & 0.877 & 0.925 & 0.853 & 0.842 & 0.759 & -0.3 & -1.2 \\
 & the O1 is Rel the O2 & 0.895 & 0.946 & 0.868 & 0.833 & 0.614 & -0.9 & 0.5 \\
\cline{1-9}
\multirow[t]{5}{*}{RTE DiT-mini} & O1 is Rel O2 & 0.865 & 0.914 & 0.838 & 0.828 & 0.644 & 0.8 & 0.4 \\
 & O1 is Rel the O2 & 0.871 & 0.931 & 0.847 & 0.834 & 0.613 & 1.1 & 1.5 \\
 & O1 Rel O2 & 0.778 & 0.845 & 0.743 & 0.737 & 0.621 & 1.5 & -0.4 \\
 & O1 Rel the O2 & 0.799 & 0.879 & 0.770 & 0.762 & 0.616 & 1.8 & 0.0 \\
 & the O1 is Rel the O2 & 0.767 & 0.912 & 0.721 & 0.680 & 0.471 & -0.1 & 1.8 \\
\cline{1-9}
\multirow[t]{5}{*}{RTE DiT-micro} & O1 is Rel O2 & 0.726 & 0.683 & 0.508 & 0.489 & 0.315 & -0.2 & 0.2 \\
 & O1 is Rel the O2 & 0.738 & 0.705 & 0.520 & 0.501 & 0.312 & 0.3 & 0.2 \\
 & O1 Rel O2 & 0.626 & 0.604 & 0.395 & 0.386 & 0.270 & 0.1 & -1.5 \\
 & O1 Rel the O2 & 0.649 & 0.639 & 0.410 & 0.401 & 0.269 & -0.2 & -1.2 \\
 & the O1 is Rel the O2 & 0.665 & 0.724 & 0.432 & 0.403 & 0.234 & 2.1 & -0.7 \\
\cline{1-9}
\multirow[t]{5}{*}{RTE DiT-nano} & O1 is Rel O2 & 0.360 & 0.531 & 0.195 & 0.090 & 0.049 & 3.2 & -0.1 \\
 & O1 is Rel the O2 & 0.372 & 0.539 & 0.205 & 0.096 & 0.051 & 2.4 & -1.6 \\
 & O1 Rel O2 & 0.270 & 0.568 & 0.146 & 0.069 & 0.037 & 5.1 & -0.4 \\
 & O1 Rel the O2 & 0.279 & 0.581 & 0.151 & 0.071 & 0.036 & 3.2 & -2.4 \\
 & the O1 is Rel the O2 & 0.399 & 0.632 & 0.193 & 0.082 & 0.047 & -3.6 & -3.5 \\
\cline{1-9}
\multirow[t]{5}{*}{RTE w/pos DiT-B} & O1 is Rel O2 & 0.859 & 0.899 & 0.415 & 0.207 & 0.192 & -0.1 & 0.1 \\
 & O1 is Rel the O2 & 0.863 & 0.903 & 0.416 & 0.207 & 0.190 & -0.0 & -0.0 \\
 & O1 Rel O2 & 0.856 & 0.893 & 0.412 & 0.205 & 0.191 & 0.0 & -0.0 \\
 & O1 Rel the O2 & 0.860 & 0.902 & 0.415 & 0.206 & 0.191 & -0.0 & 0.1 \\
 & the O1 is Rel the O2 & 0.866 & 0.910 & 0.417 & 0.207 & 0.188 & 0.1 & 0.0 \\
\cline{1-9}
\multirow[t]{5}{*}{T5 DiT-B} & O1 is Rel O2 & 0.857 & 0.892 & 0.820 & 0.808 & 0.749 & -0.8 & -0.5 \\
 & O1 is Rel the O2 & 0.915 & 0.931 & 0.894 & 0.498 & 0.306 & -33.7 & -24.9 \\
 & O1 Rel O2 & 0.853 & 0.871 & 0.825 & 0.608 & 0.493 & -4.5 & -16.6 \\
 & O1 Rel the O2 & 0.941 & 0.958 & 0.925 & 0.400 & 0.217 & -35.1 & -37.0 \\
 & the O1 is Rel the O2 & 0.917 & 0.935 & 0.896 & 0.529 & 0.309 & -18.4 & -20.7 \\
\cline{1-9}
\multirow[t]{5}{*}{T5 DiT-mini} & O1 is Rel O2 & 0.856 & 0.889 & 0.822 & 0.810 & 0.659 & -0.4 & -0.6 \\
 & O1 is Rel the O2 & 0.877 & 0.922 & 0.855 & 0.487 & 0.259 & -35.8 & -24.1 \\
 & O1 Rel O2 & 0.816 & 0.844 & 0.772 & 0.559 & 0.399 & -12.7 & -18.2 \\
 & O1 Rel the O2 & 0.895 & 0.946 & 0.878 & 0.391 & 0.184 & -38.7 & -37.6 \\
 & the O1 is Rel the O2 & 0.906 & 0.947 & 0.885 & 0.537 & 0.272 & -17.1 & -19.3 \\
\cline{1-9}
\multirow[t]{5}{*}{T5 DiT-B WD} & O1 is Rel O2 & 0.183 & 0.114 & 0.033 & 0.033 & 0.031 & -1.2 & 1.3 \\
 & O1 is Rel the O2 & 0.169 & 0.104 & 0.030 & 0.017 & 0.013 & -39.7 & -22.2 \\
 & O1 Rel O2 & 0.164 & 0.110 & 0.032 & 0.025 & 0.023 & -5.0 & -15.4 \\
 & O1 Rel the O2 & 0.181 & 0.122 & 0.037 & 0.016 & 0.011 & -40.4 & -34.8 \\
 & the O1 is Rel the O2 & 0.160 & 0.100 & 0.028 & 0.017 & 0.013 & -15.0 & -18.4 \\
\cline{1-9}
\multirow[t]{5}{*}{T5 DiT-mini WD} & O1 is Rel O2 & 0.894 & 0.942 & 0.866 & 0.854 & 0.667 & -0.4 & 1.0 \\
 & O1 is Rel the O2 & 0.911 & 0.967 & 0.886 & 0.521 & 0.265 & -42.0 & -19.1 \\
 & O1 Rel O2 & 0.843 & 0.886 & 0.804 & 0.596 & 0.429 & -9.1 & -16.7 \\
 & O1 Rel the O2 & 0.911 & 0.975 & 0.888 & 0.414 & 0.189 & -47.1 & -35.3 \\
 & the O1 is Rel the O2 & 0.911 & 0.965 & 0.887 & 0.514 & 0.249 & -22.4 & -13.8 \\
\cline{1-9}
\multirow[t]{5}{*}{CLIP DiT-B} & O1 is Rel O2 & 0.806 & 0.900 & 0.772 & 0.759 & 0.701 & -0.7 & -0.7 \\
  & O1 is Rel the O2 & 0.832 & 0.925 & 0.805 & 0.782 & 0.676 & +0.2 & +1.1 \\
  & O1 Rel O2 & 0.837 & 0.936 & 0.816 & 0.762 & 0.614 & -6.3 & -1.2 \\
  & O1 Rel the O2 & 0.839 & 0.940 & 0.822 & 0.768 & 0.619 & -5.9 & +0.8 \\
  & the O1 is Rel the O2 & 0.858 & 0.957 & 0.842 & 0.794 & 0.620 & +2.3 & +2.7 \\
\cline{1-9}
 \multirow[t]{5}{*}{CLIP DiT-mini} & O1 is Rel O2 & 0.865 & 0.888 & 0.828 & 0.814
& 0.664 & -0.0 & -0.9 \\
  & O1 is Rel the O2 & 0.900 & 0.918 & 0.865 & 0.819 & 0.615 & +4.3 & +5.4 \\
  & O1 Rel O2 & 0.876 & 0.885 & 0.839 & 0.816 & 0.644 & -3.0 & -3.2 \\
  & O1 Rel the O2 & 0.907 & 0.919 & 0.876 & 0.844 & 0.630 & +2.0 & +4.2 \\
  & the O1 is Rel the O2 & 0.911 & 0.930 & 0.878 & 0.822 & 0.598 & +0.3 & +6.6 \\
\cline{1-9}
\cline{1-9}
\bottomrule
\end{tabular}

\end{table}
\clearpage
\subsection{Evaluation Training Dynamics} 
\label{app:eval_learn_dynamics}
\begin{figure}[!htp]
\begin{center}
\includegraphics[width=0.9\textwidth]{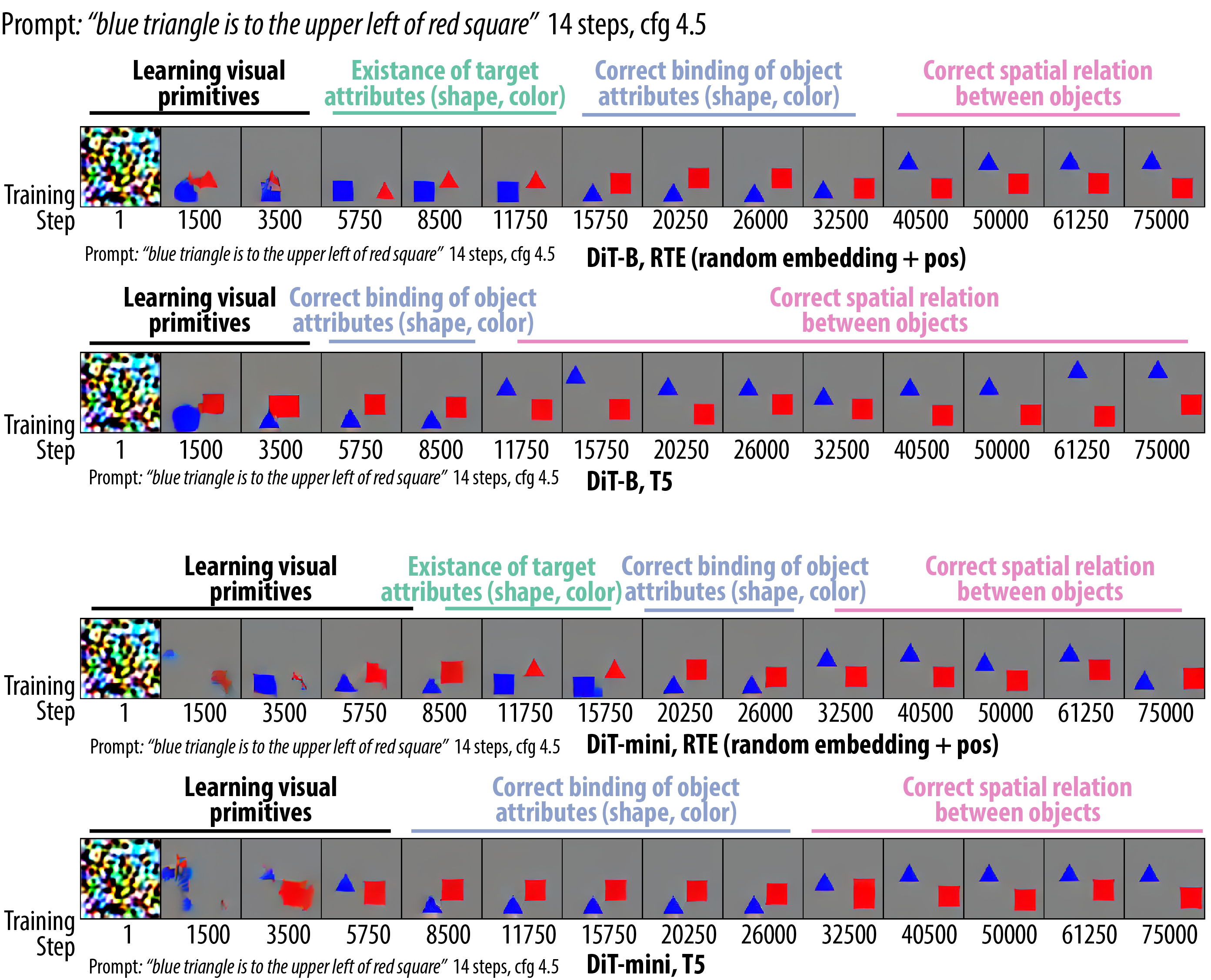}
\end{center}
\caption{
\textbf{Comparison of training dynamics of DiT models with different text encoding and scale.} Specific evaluation prompt used was \enquote{\textit{blue triangle is to the upper left of red square}}, sampled with 14 steps at cfg 4.5, sampled from the same noise seed. 
Further, T5 models immediately learn to achieve object attribute binding after learning attributes themselves, while random embedding model (RTE) gradually learn the correct attribute binding and then the correct spatial relation.   
Across scales, generally, larger scale models train faster.  
}
\end{figure}

\begin{figure}[!htbp]
\begin{center}
\includegraphics[width=0.99\textwidth]{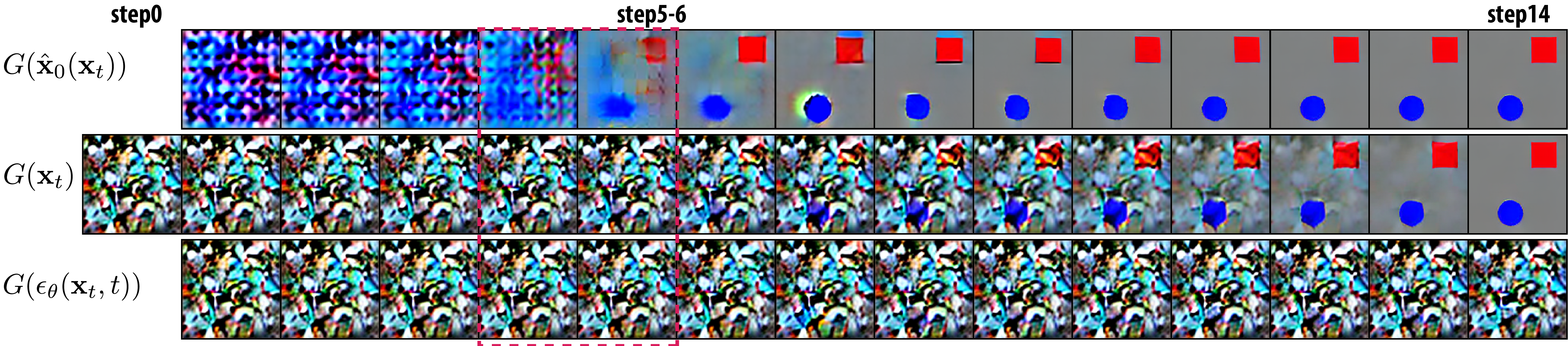}
\end{center}
\label{fig:train_dynamics_vis}
\caption{\textbf{Observation on sampling dynamics} Specific evaluation prompt used was \enquote{\textit{the red square is above and to the right of the blue circle}}, sampled with 14 steps at cfg 4.5. Model used is RTE x DiT-B. 
A transition can be seen at step 4-6, where the two object at their final positions can be clearly seen from the expected outcome $G(\hat{\mathbf{x}}_0(\mathbf{x}_t))$. }
\end{figure}

\clearpage
\subsection{RTE-DiT Spatial Relation Head Additional Evidence} \label{app_sec:relation_head}

\begin{figure}[!htbp]
\begin{center}
\includegraphics[width=0.99\textwidth]{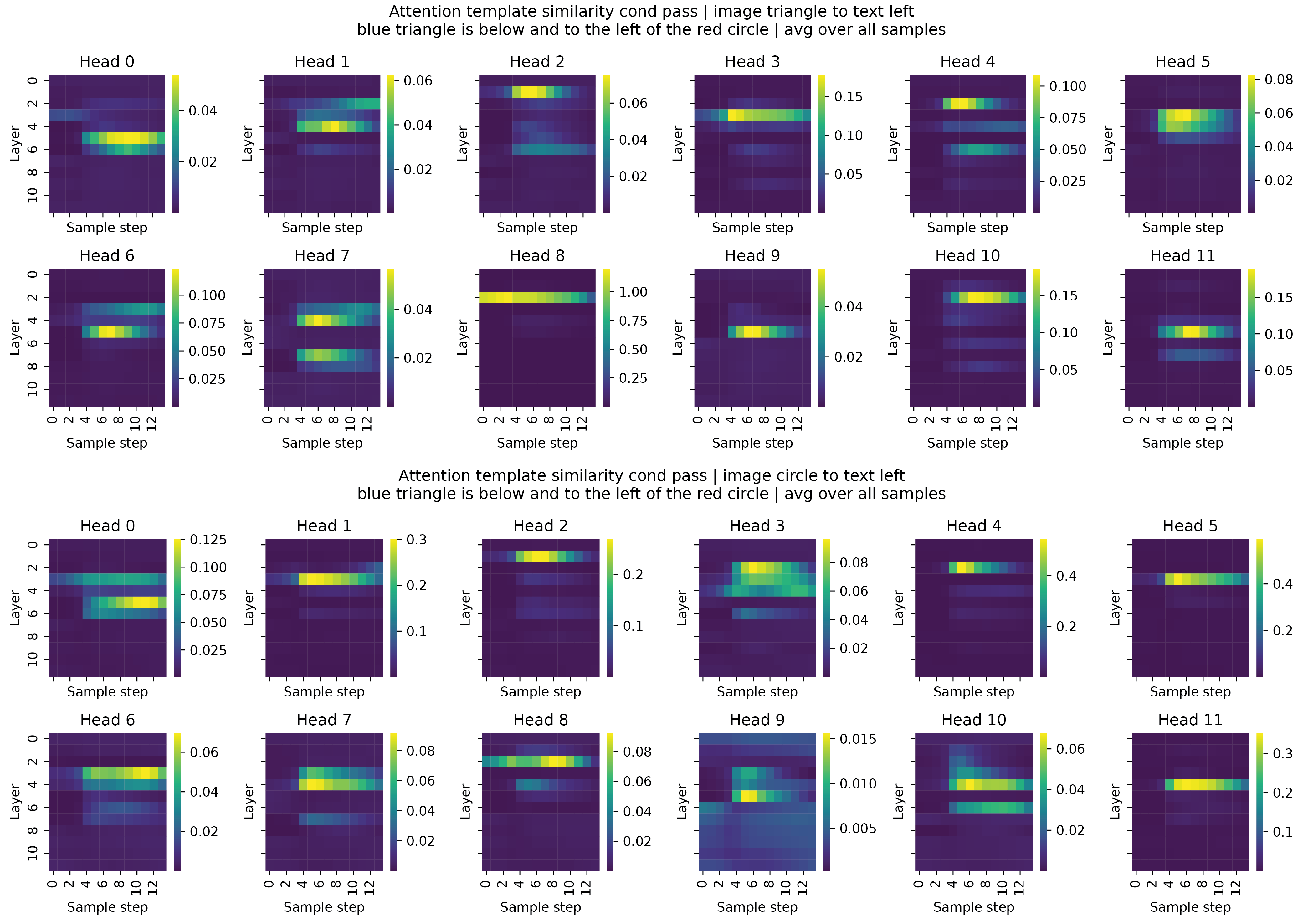}
\end{center}
\label{suppfig:attn_dynamics_demo}
\caption{\textbf{Cross attention energy during sampling dynamics.} Highlighting the smooth varying attention strength, and the salient contribution of L2H8 head from first object to relation words in RTE x DiT-B.}
\end{figure}
\begin{figure}[!htbp]
\begin{center}
\includegraphics[width=0.7\textwidth]{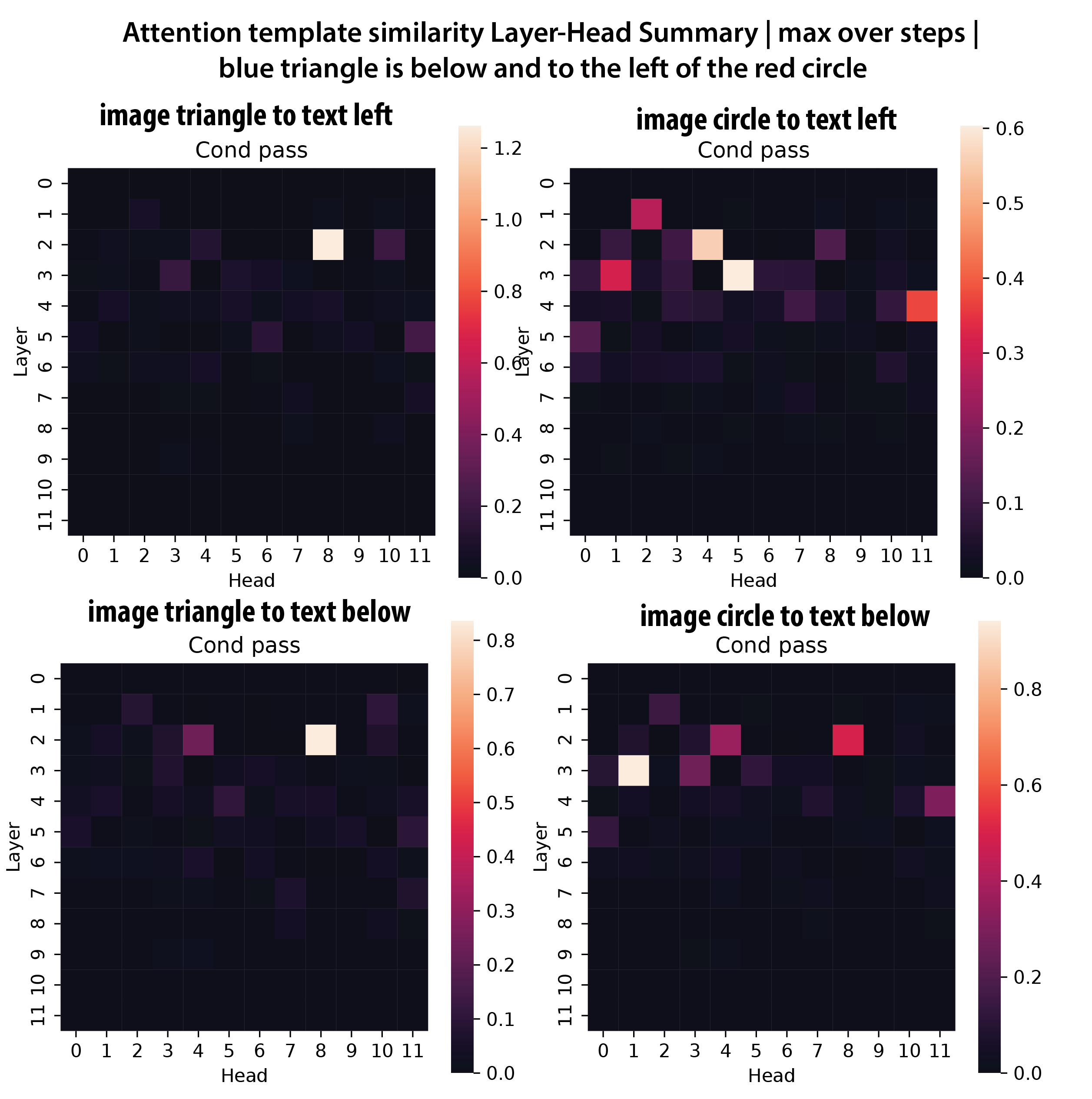}
\end{center}
\label{suppfig:attn_dynamics_demo_summary}
\caption{\textbf{Cross attention energy summary (max over time) for the specific prompt above}}
\end{figure}

\begin{figure}[!htbp]
\begin{center}
\includegraphics[width=0.85\textwidth]{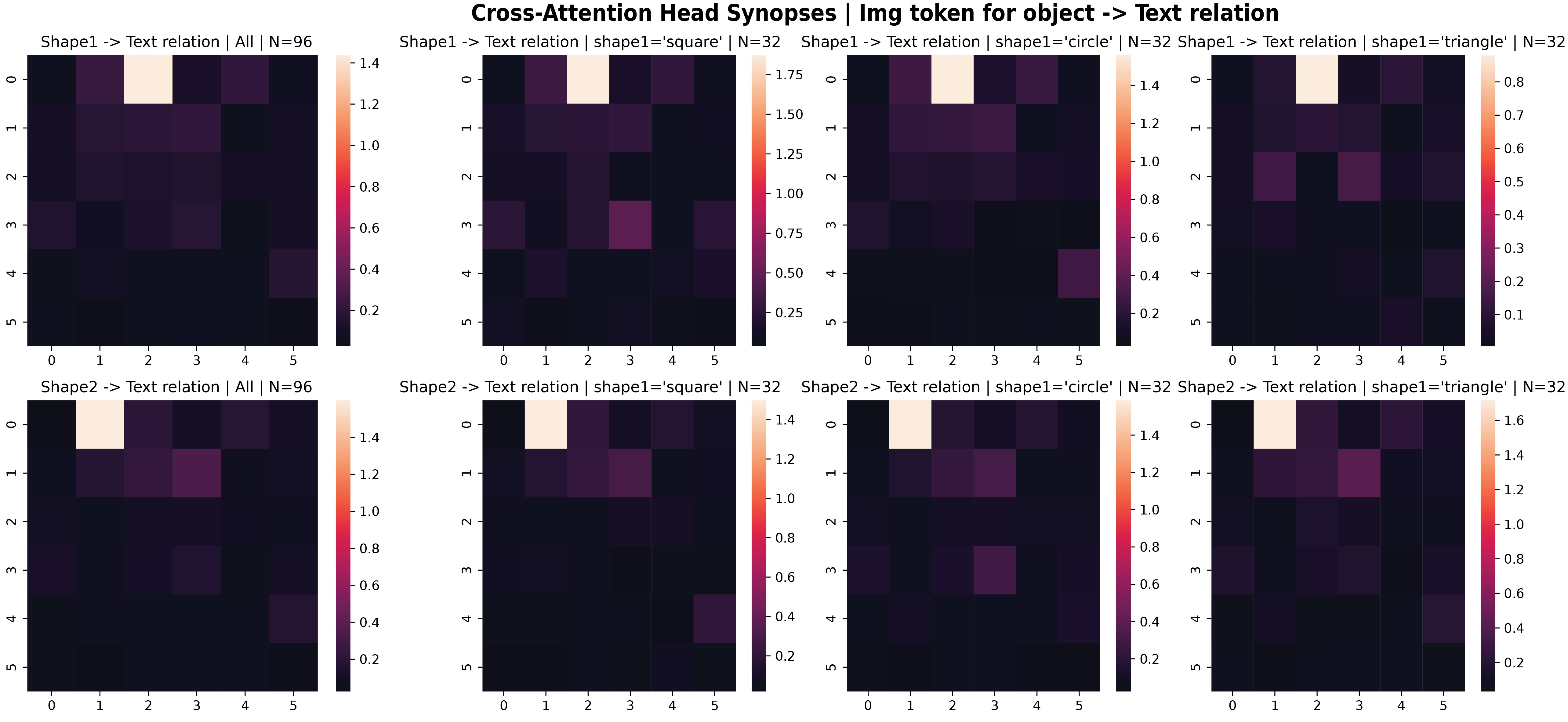}
\end{center}
\label{suppfig:DiT-mini_rel}
\caption{\textbf{Attention Synopsis for Shape to Relation word for RTE x DiT-mini, showing it's invariant to the specific shape of object1.}}
\end{figure}

\begin{figure}[!htbp]
\begin{center}
\includegraphics[width=1\textwidth]{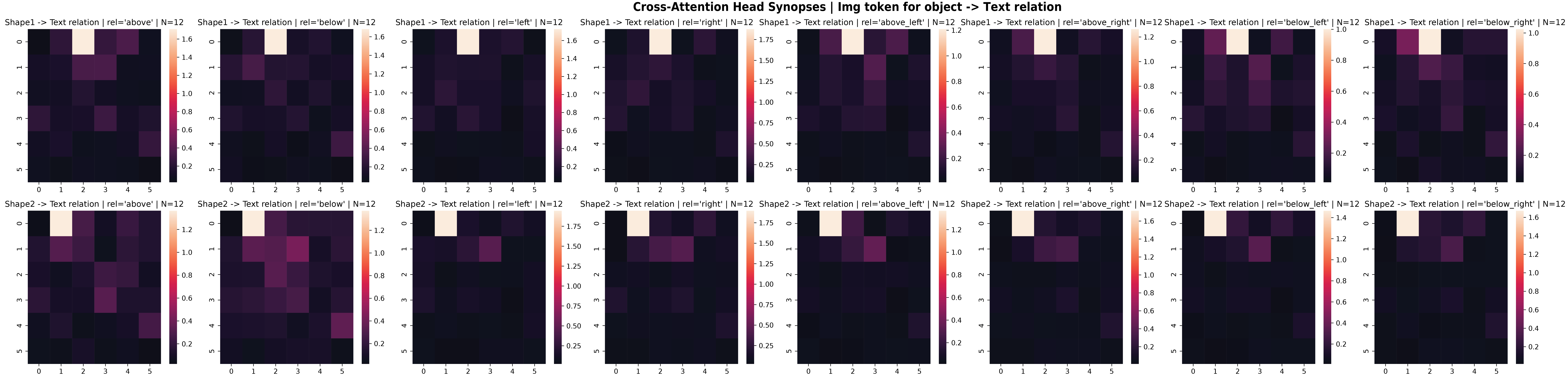}
\end{center}
\label{suppfig:DiT-mini_rel_inv}
\caption{\textbf{Attention Synopsis for Shape to Relation word for RTE x DiT-mini, showing it's invariant to the specific spatial relation and phrasing.}}
\end{figure}

\begin{figure}[!htbp]
\begin{center}
\includegraphics[width=0.85\textwidth]{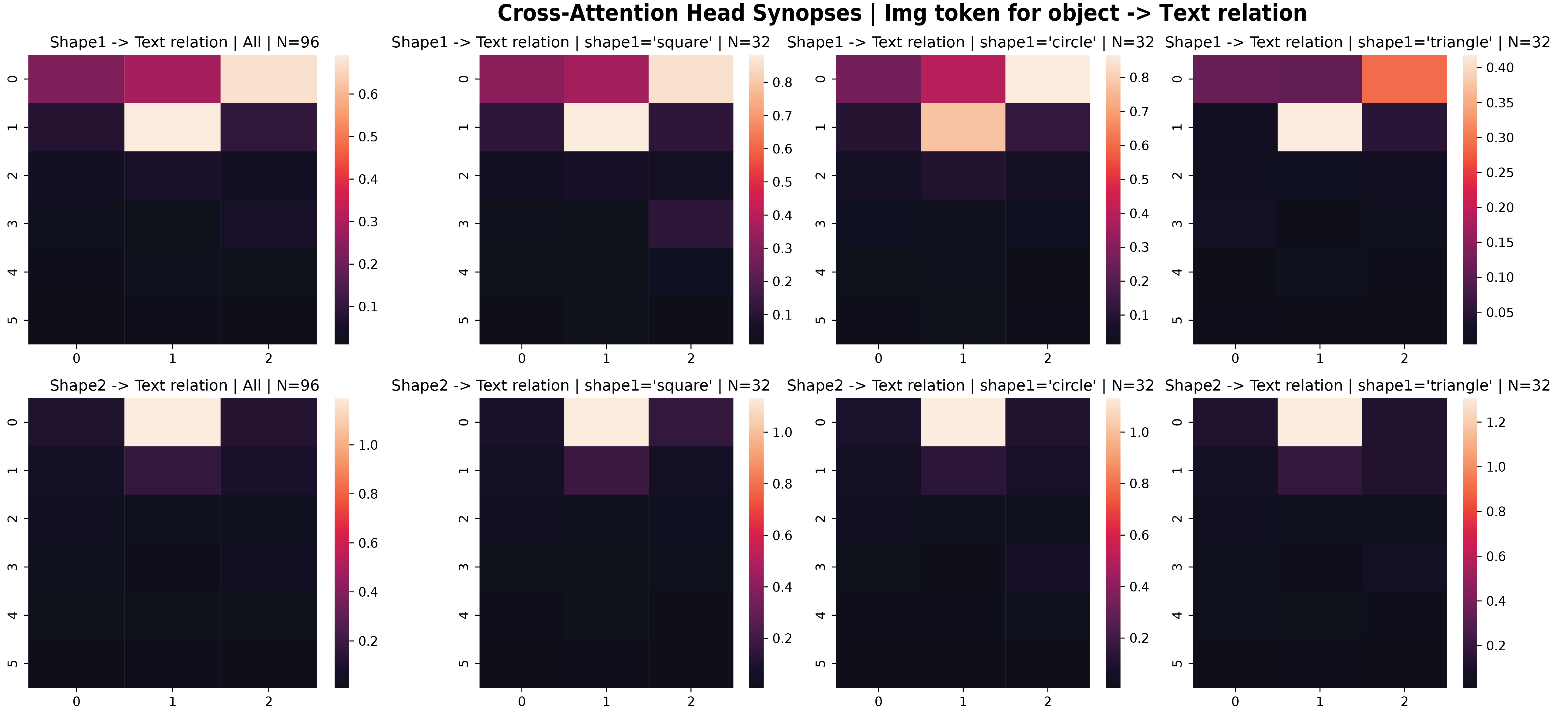}
\end{center}
\label{suppfig:DiT-mirco_rel}
\caption{\textbf{Attention Synopsis for Shape to Relation word for RTE x DiT-micro, showing it's invariant to the specific shape of object1.}}
\end{figure}

\begin{figure}[!htbp]
\begin{center}
\includegraphics[width=1\textwidth]{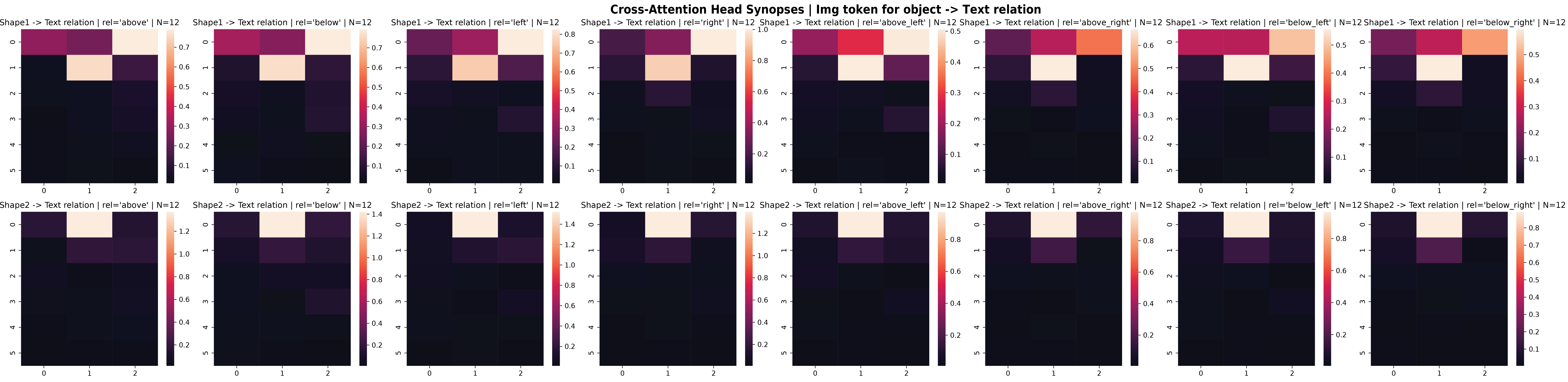}
\end{center}
\label{suppfig:DiT-micro_rel_inv}
\caption{\textbf{Attention Synopsis for Shape to Relation word for RTE x DiT-micro, showing it's invariant to the specific spatial relation and phrasing.}}
\end{figure}

\clearpage
\subsection{RTE-DiT Object Head Additional Evidence} 
\label{app_sec:object_head}

\begin{figure}[!htbp]
\begin{center}
\includegraphics[width=0.99\textwidth]{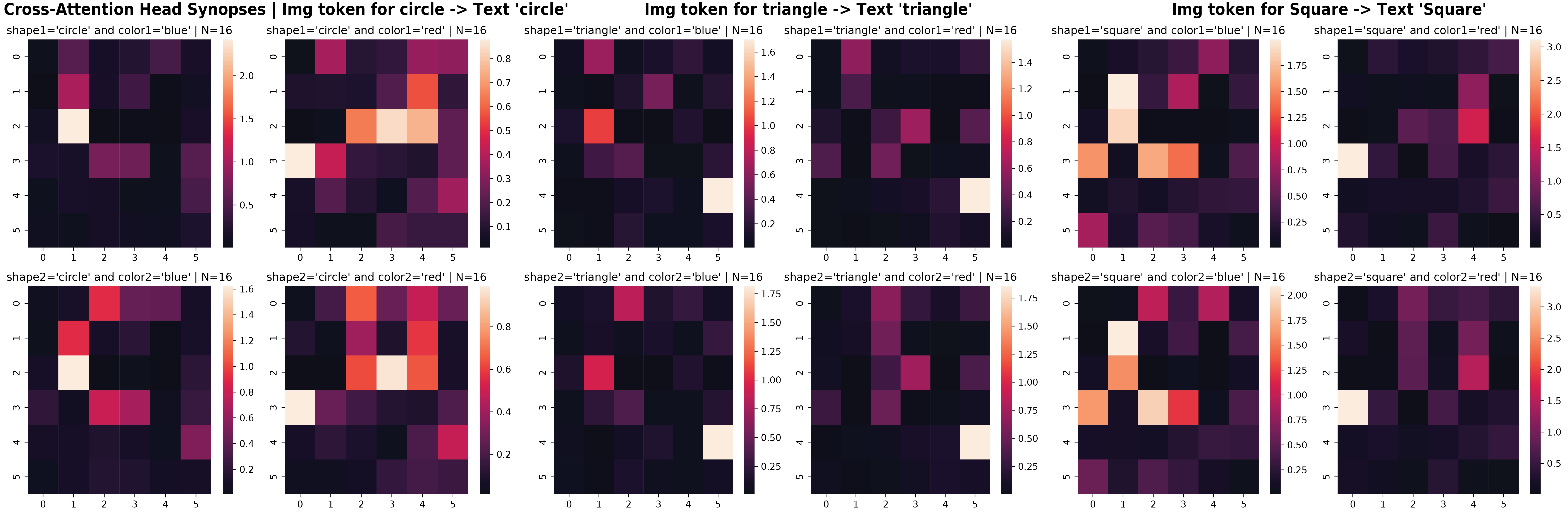}
\end{center}
\label{suppfig:DiT-mini_obj}
\caption{\textbf{Attention Synopsis for Shape to corresponding shape word for RTE x DiT-mini.}}
\end{figure}

\begin{figure}[!htbp]
\begin{center}
\includegraphics[width=0.99\textwidth]{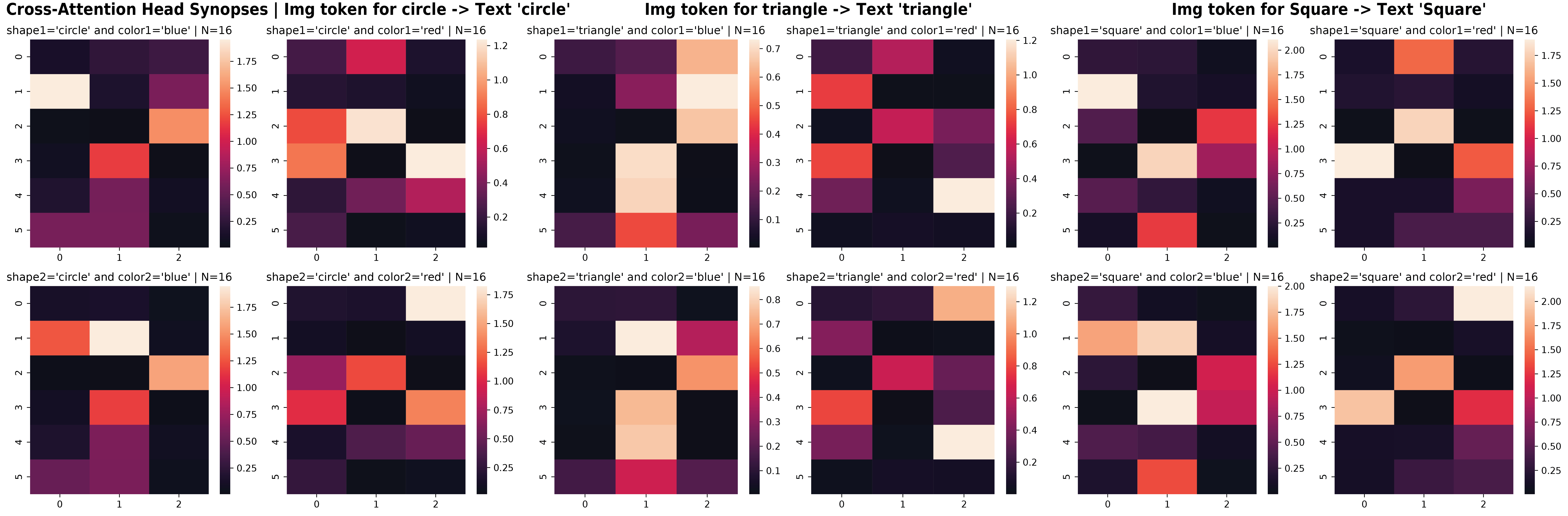}
\end{center}
\label{suppfig:DiT-mirco_obj}
\caption{\textbf{Attention Synopsis for Shape to corresponding Shape word for RTE x DiT-micro.}}
\end{figure}

\clearpage
\subsection{T5-DiT Attention Synopsis} 
\label{app_sec:T5_DiT_attn_synops}

\begin{figure}[!htbp]
\begin{center}
\includegraphics[width=0.85\textwidth]{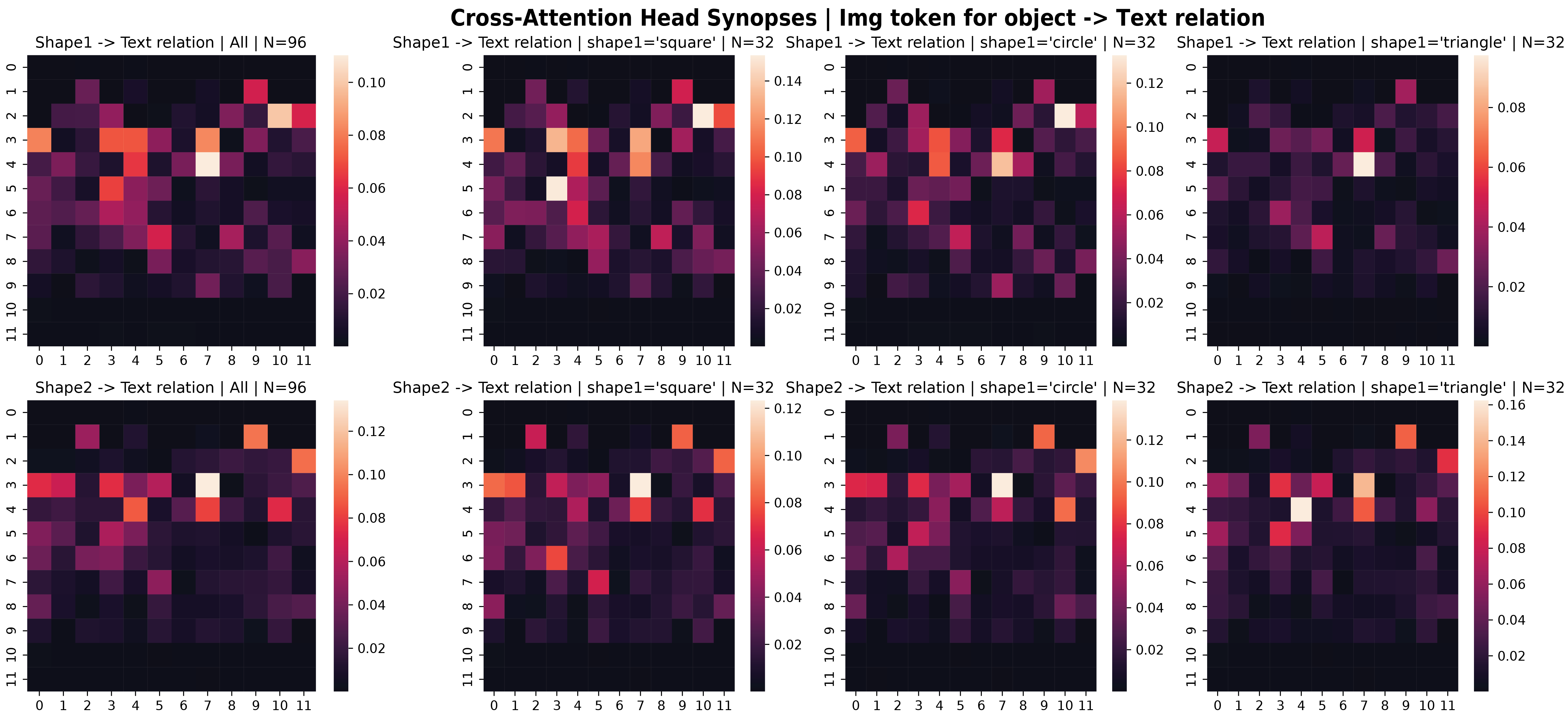}
\end{center}
\label{suppfig:DiT-B_T5_rel}
\caption{\textbf{Attention Synopsis for Shape to Relation word for T5 x DiT-B. The pattern is much less clear than RTE.}}
\end{figure}

\begin{figure}[!htbp]
\begin{center}
\includegraphics[width=1\textwidth]{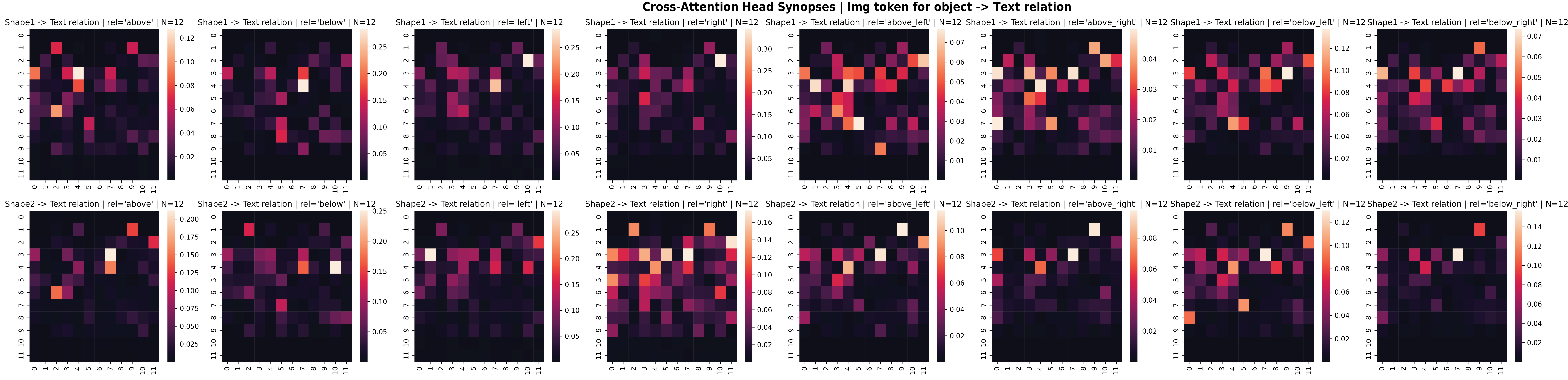}
\end{center}
\label{suppfig:DiT-B_T5_rel_inv}
\caption{\textbf{Attention Synopsis for Shape to Relation word for T5 x DiT-B. The pattern is much less clear than RTE.}}
\end{figure}


\clearpage
\subsection{T5-DiT Representation Analysis} 
\label{app_sec:T5_DiT_Repr_analysis}

\begin{figure}[!htp]
\begin{center}
\includegraphics[width=0.99\textwidth]{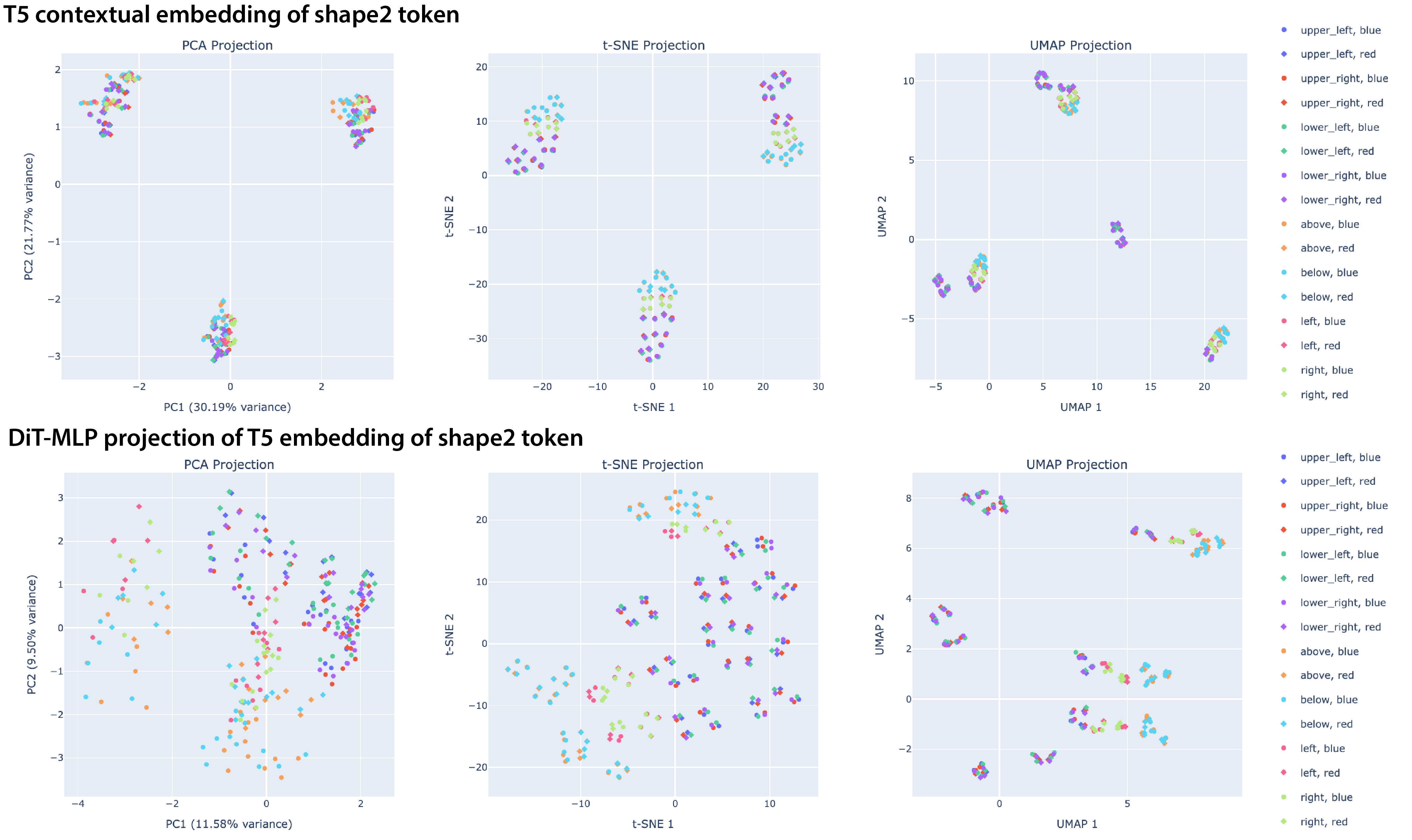}
\end{center}
\caption{
\textbf{Dimension reduction visualization of \texttt{shape2} token representation (PCA, tSNE, UMAP).} Top row: T5 contextual embedding (4096d), Bottom row: Caption projection (784d) using MLP from T5 x DiT-B. }\label{fig:T5_context_wordvec_UMap}
\end{figure}

\clearpage
\subsection{Large scale weight-space relation head screening}\label{app_sec:rel_head_screening}
Here we show the results of large scale weight-space relation head screening for the three family of DiT models with text encoders (RTE, T5, CLIP). 

\begin{figure}[!htp]
\begin{center}
\includegraphics[width=0.95\textwidth]{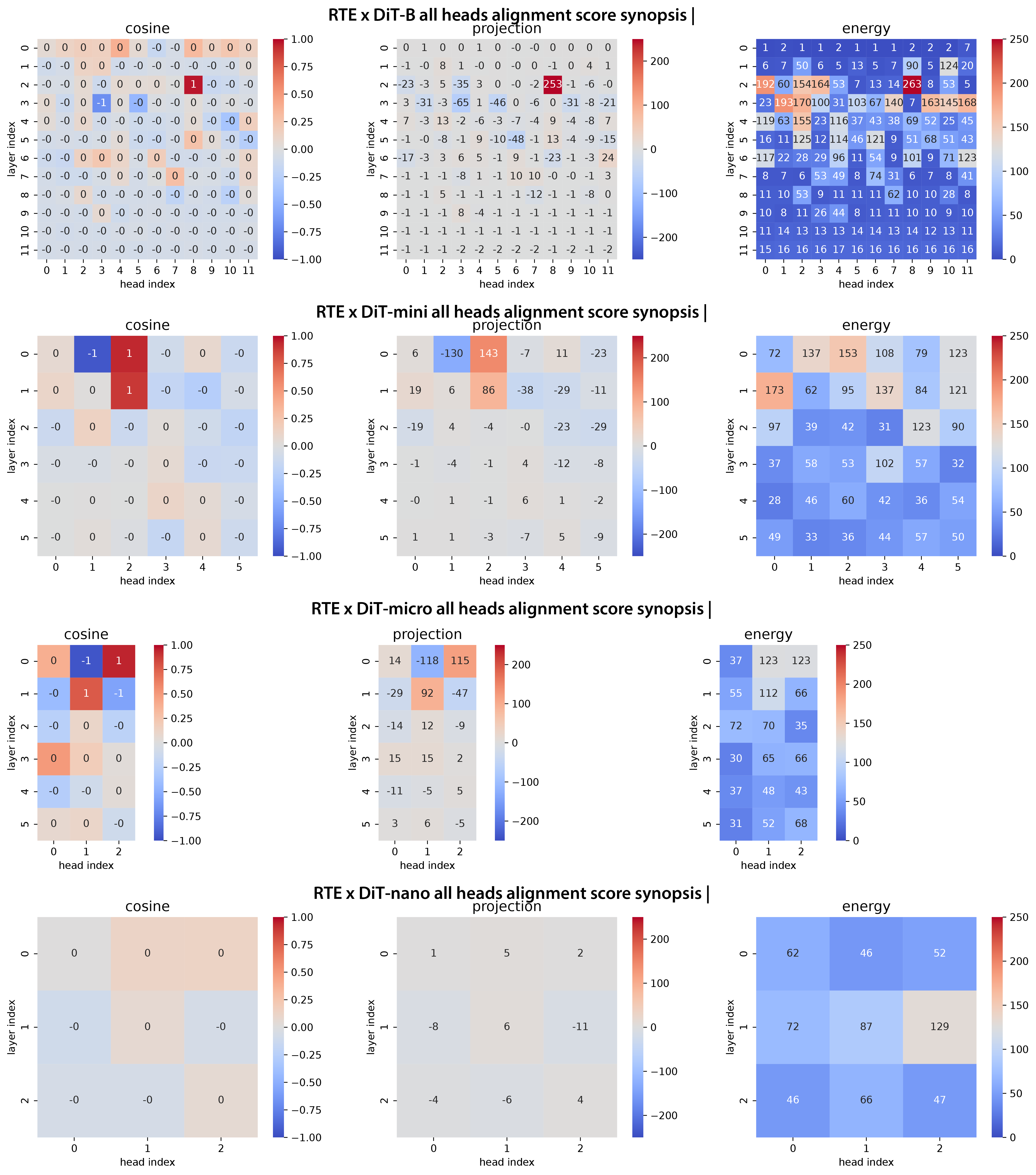}
\end{center}
\caption{
\textbf{Weight-space relation head screening for RTE-DiT (B, mini, micro, nano).} Each column shows a different alignment metric (cosine, projection, energy); each row shows a different model.
}
\label{suppfig:HeadScreening_RTE-DiT}
\end{figure}

\begin{figure}[!htp]
\begin{center}
\includegraphics[width=0.95\textwidth]{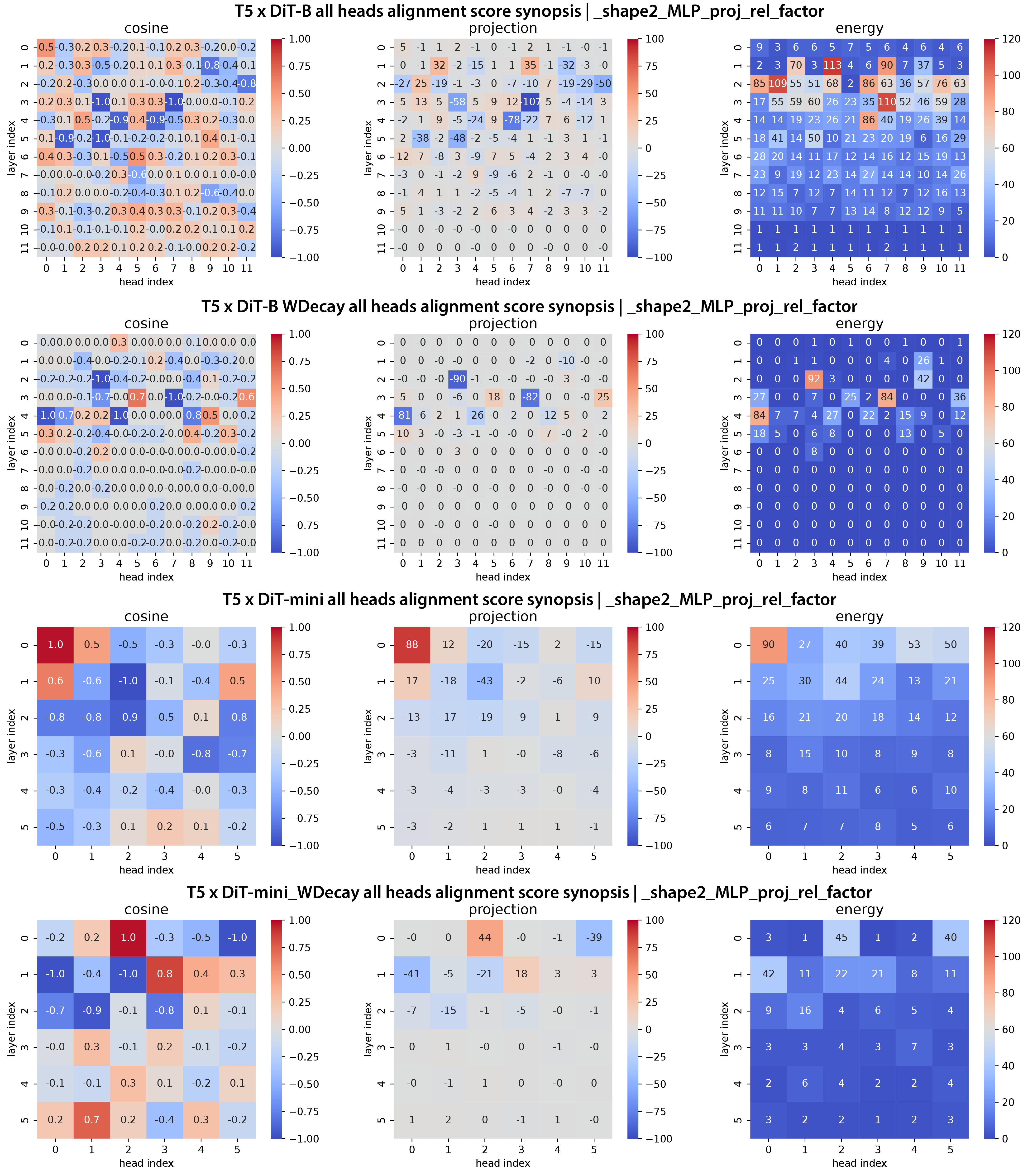}
\end{center}
\caption{
\textbf{Weight-space relation head screening for T5-DiT (\texttt{shape2} token) (B, mini, with/without weight decay).} Each column shows a different alignment metric (cosine, projection, energy); each row shows a different model.
Note this one uses the spatial feature vectors from variance partitioning on \texttt{shape2} token.
}
\label{suppfig:HeadScreening_T5-DiT_shape2}
\end{figure}

\begin{figure}[!htp]
    \begin{center}
    \includegraphics[width=0.95\textwidth]{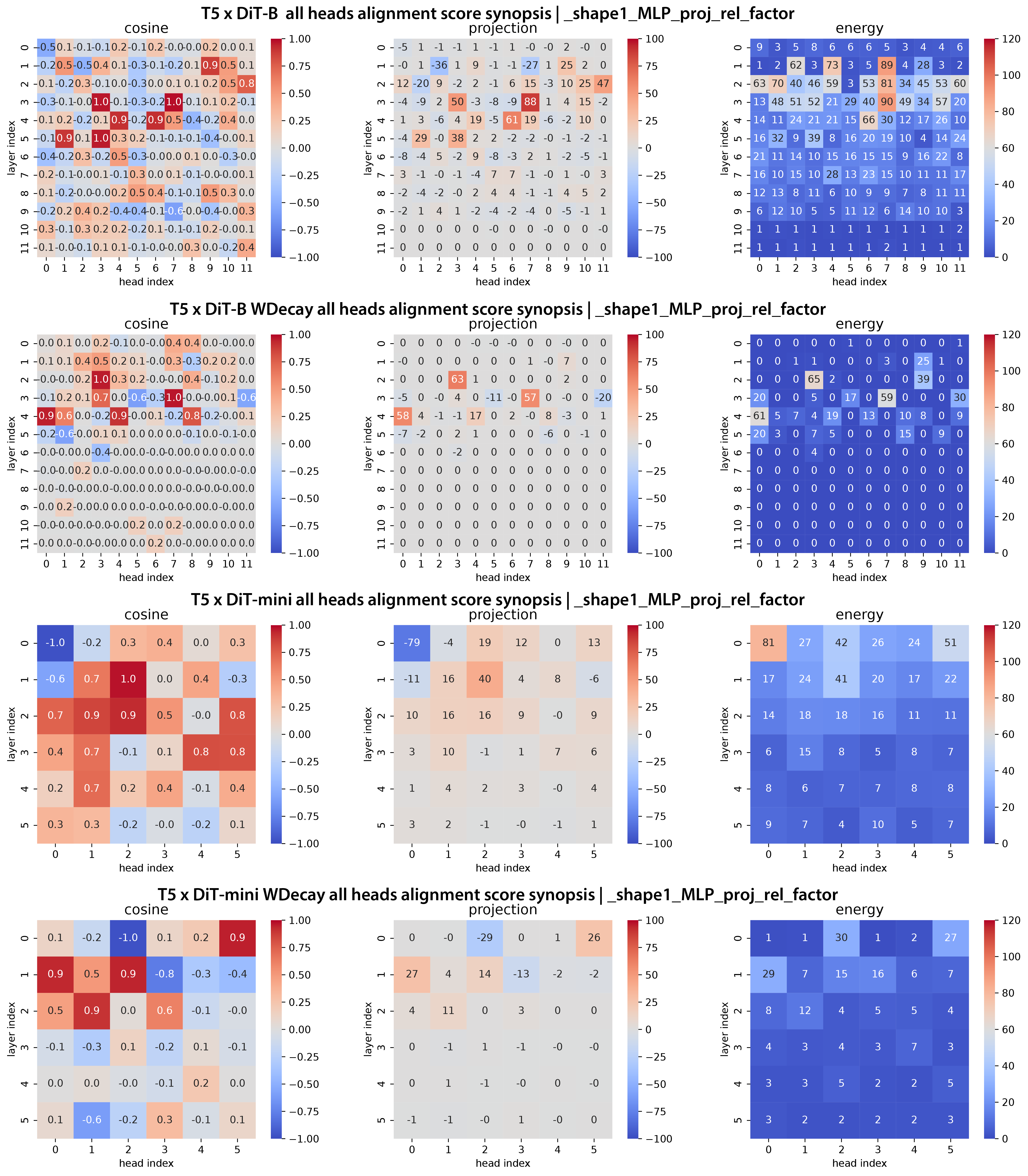}
    \end{center}
    \caption{
    \textbf{Weight-space relation head screening for T5-DiT (\texttt{shape1} token) (B, mini, with/without weight decay).} Each column shows a different alignment metric (cosine, projection, energy); each row shows a different model. Note this one uses the spatial feature vectors from variance partitioning on \texttt{shape1} token.
    }
    \label{suppfig:HeadScreening_T5-DiT_shape1}
\end{figure}
    
\begin{figure}[!htp]
    \begin{center}
    \includegraphics[width=0.95\textwidth]{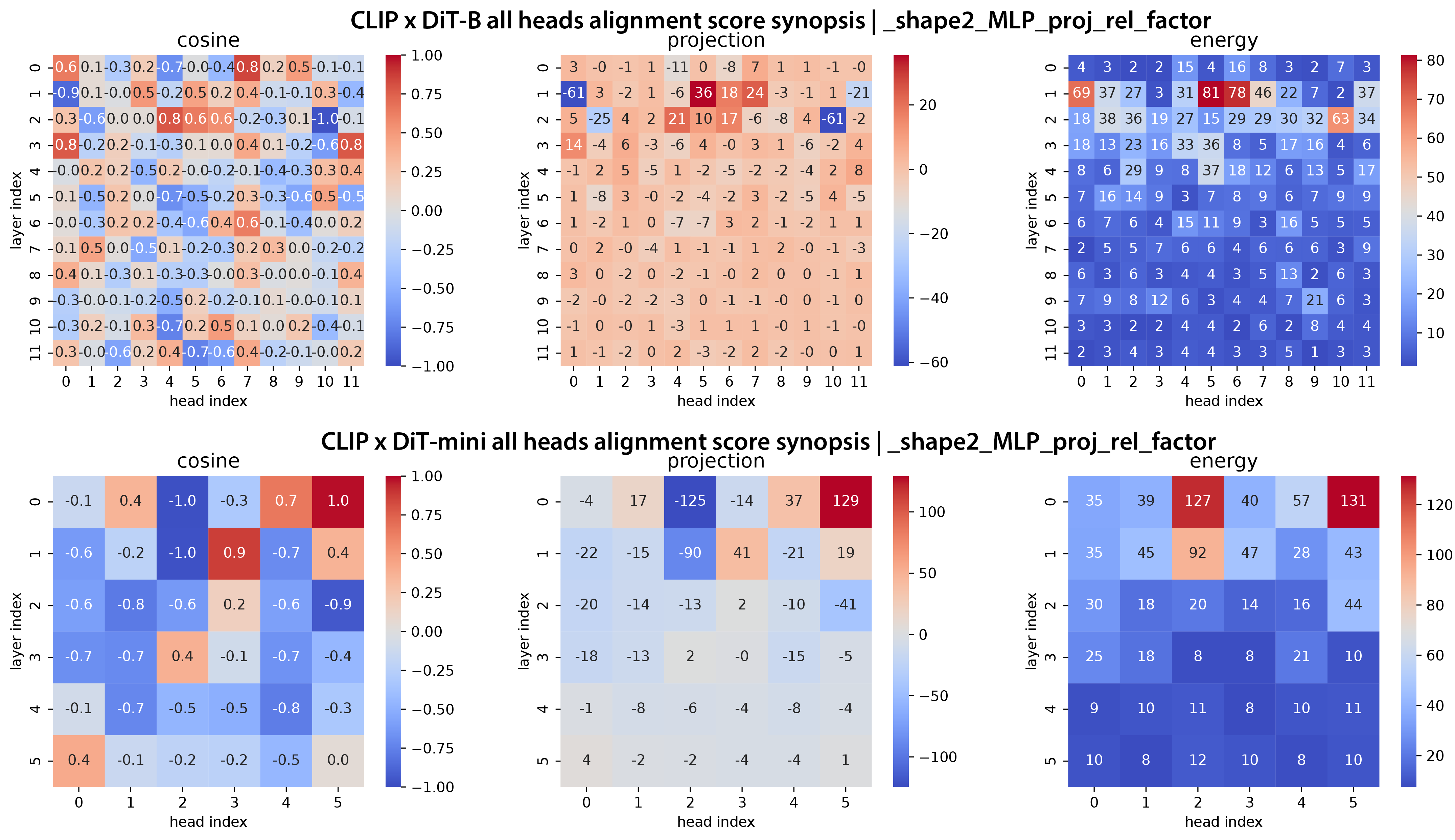}
    \end{center}
    \caption{
    \textbf{Weight-space relation head screening for CLIP-DiT (B, mini).} Each column shows a different alignment metric (cosine, projection, energy); each row shows a different model.
    }
    \label{suppfig:HeadScreening_CLIP-DiT}
\end{figure}

\clearpage
\subsection{Robustness of RTE-based and T5-based circuits}
\label{app_sec:robustness}
\begin{figure}[!htp]
\begin{center}
\vspace{-6pt}
\includegraphics[width=0.6\linewidth]{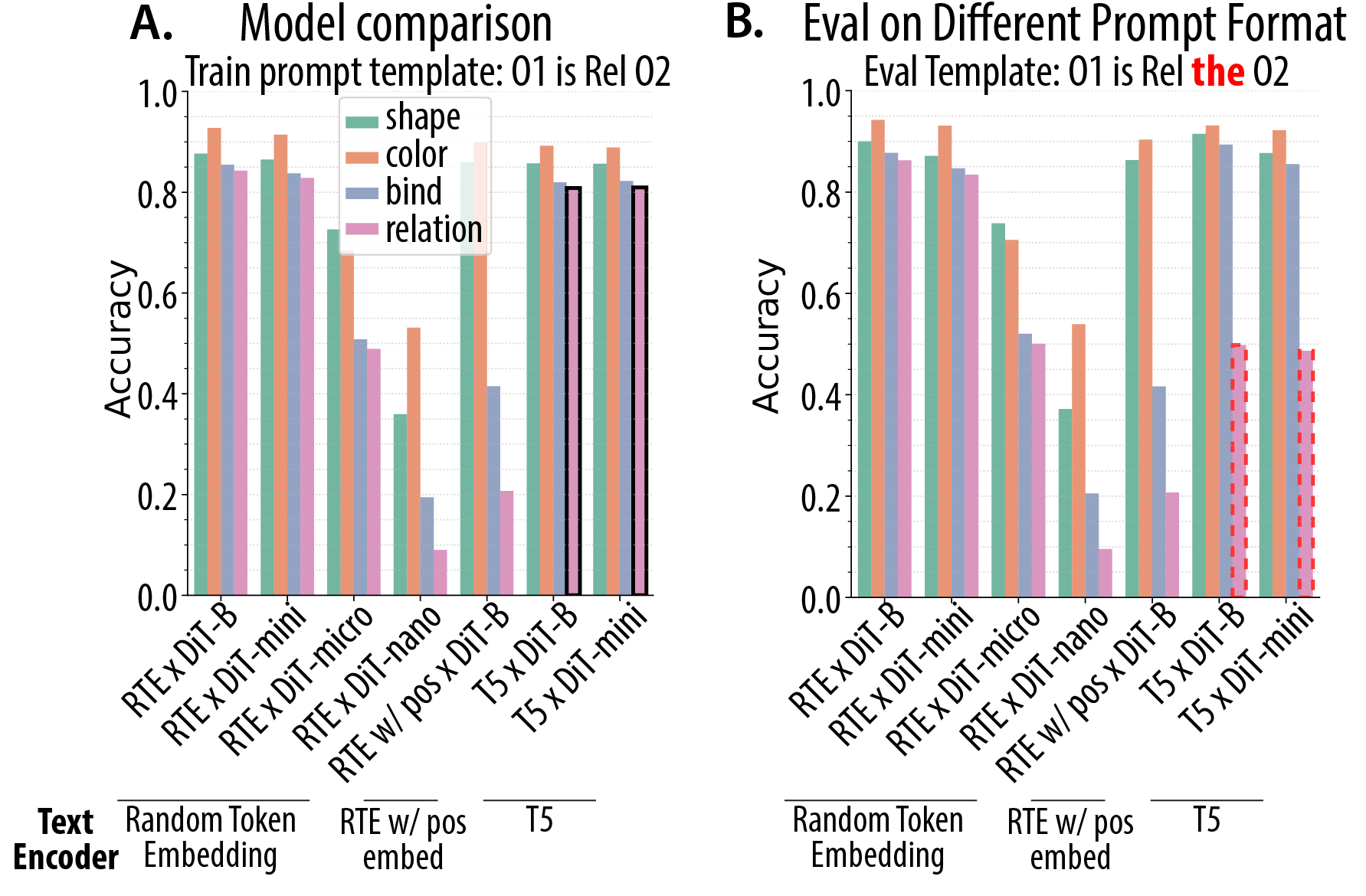}
\vspace{-8pt}
\end{center}
\caption{
\textbf{Evaluation of model performance on trained and generalized prompt template.} 
}\label{fig:eval_bench_bars}
\vspace{-2pt}
\end{figure}

\begin{figure}[!htp]
  \centering
\vspace{-4pt}
  \includegraphics[width=0.85\linewidth]{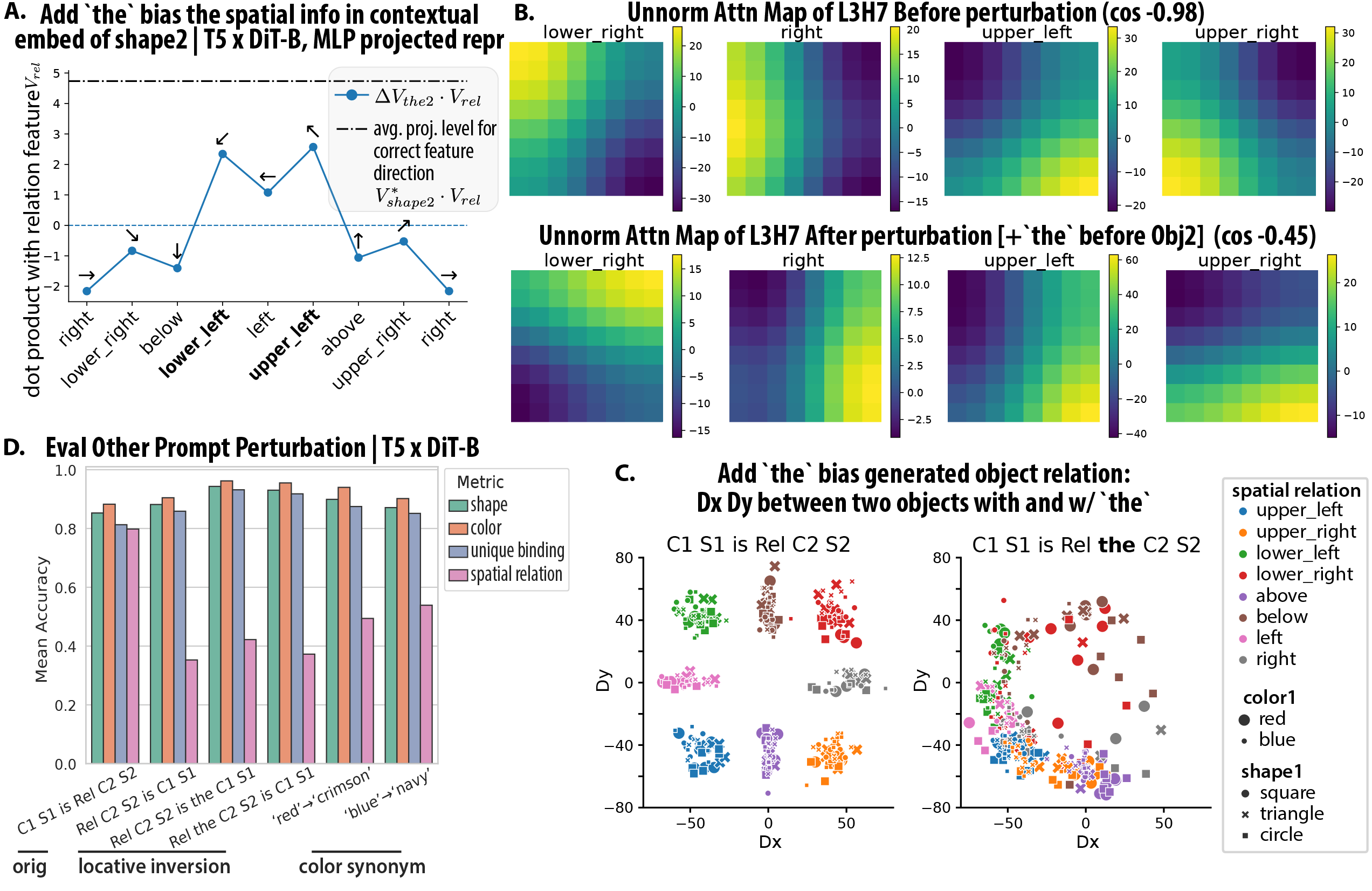}
   \caption{\textbf{T5-DiT Robustness and Circuit Analysis} \textbf{(A)} The change of contextual embedding of the \texttt{shape2} token when the word "the" is added interfere with the 8 relation feature directions, esp. positively with lower left and upper right feature. \textbf{(B)} Unnormalized attention maps of the relation head (layer 3, head 7) before and after the prompt perturbation. \textbf{(C)} The generated spatial relationship, i.e. the displacement (Dx, Dy) between the two objects was biased systematically by adding "the", consistent with the interference with spatial feature in \textbf{A}. 
   \textbf{(D)} Evaluation of model accuracy under other prompt perturbations, such as locative inversion and color synonyms. The spatial relation is particularly non-robust towards these perturbations.} 
   \label{fig:robustness_the}
\end{figure}

When evaluating on prompt with the exact format as the training ones, RTE- and T5-trained models have comparably high performance on spatial relation. However, slight prompt variation breaks the tie, i.e. adding \textit{the} to the prompt reduces the relational accuracy of T5-DiT model by around 40\% (Fig.~\ref{fig:eval_bench_bars}\textbf{B.}, Tab. \ref{tab:model_cmp_eval_table}). This suggests that even though the task accuracies are similar between RTE-based and T5-based T2I models, their robustness to small perturbations in the text is different. The T5-based models are more sensitive to the perturbation, we hypothesize that this is because perturbed context shifts token embeddings away from the learned directions used by its cross-attention and MLP layers.

To test this idea, we further analyze how adding or editing text tokens affects the T5 model's attention pattern (Fig.~\ref{fig:robustness_the}). Using the MLP-projected contextual embeddings, we computed the average shift in the \texttt{shape2} token's representation when "the" is inserted, defined as $\Delta V_{the2}:=\mathbb{E}[V_{shape2,the}^{*}-V_{shape2}^{*}]$. This shift aligns strongly with the spatial-relationship factors obtained from variance partitioning, showing as positive for lower-left/upper-left and negative for right (Fig.~\ref{fig:robustness_the}\textbf{A}). This perturbation on word representation propagates to the spatial gradients created by the relation head (L3H7) in DiT (Fig.~\ref{fig:robustness_the}\textbf{B}), where average gradient alignment (cosine) drops from -0.98 to 0.45, substantially distorting spatial generation. This explains why adding "the" before the second object biases generation toward the lower-left (Fig.~\ref{fig:robustness_the}\textbf{C}). We also tested other prompt perturbations, such as locative inversion and color paraphrase, which resulted in severely decreased spatial relation accuracy while single object attribute accuracy remained almost intact. This demonstrates the non-robustness of relational features and decoding in DiT-T5 (Fig.~\ref{fig:robustness_the}\textbf{D}).

\begin{figure}[!htp]
    \begin{center}
    \includegraphics[width=0.99\textwidth]{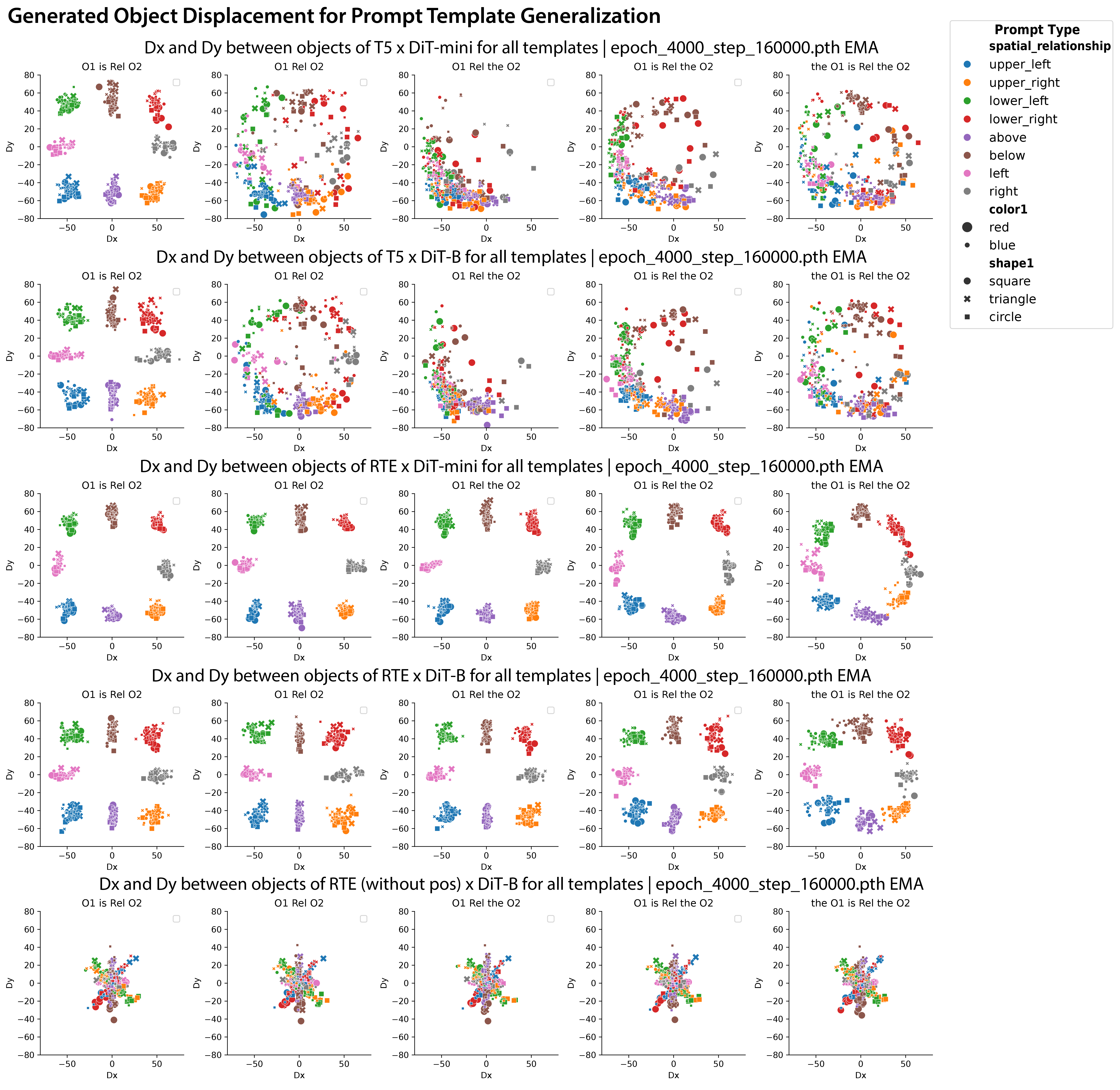}
    \vspace{-10pt}
    \end{center}
    \caption{\textbf{Prompt generalization behavior of RTE-DiT and T5-DiT.} 
    Each row shows a different model, each column shows a different prompt template. 
    In each panel, we plot the displacement between the two objects (Dx, Dy) parsed from the generated images; prompts are represented by the color (relationship), size (color1) and marker type (shape1). 
    One can see the addition of "the" at different positions (before first or second object) biases the generation towards different relations in T5-DiT, but not in RTE-DiT.
    }
    \label{suppfig:prompt_generaization_cmp}
\end{figure}

\clearpage
\subsection{Circuit dissection of PixArt with CLIP text encoder}
\label{app_subsec:CLIP_DiT}
\begin{figure}[!htp]
    \centering
    \vspace{-5pt}
    \includegraphics[width=0.90\linewidth]{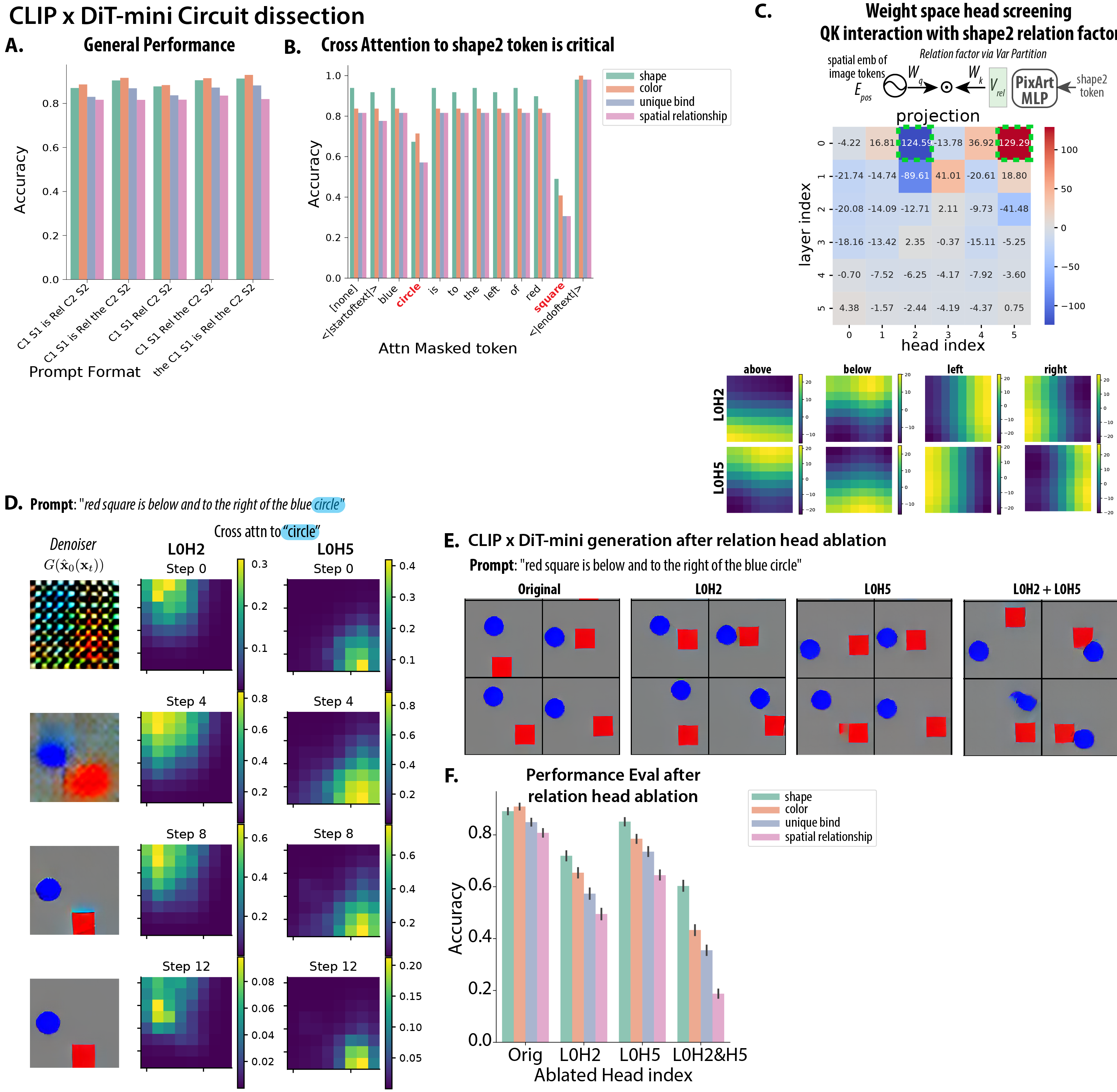}
    \vspace{-5pt}
    \caption{\textbf{Circuit dissection of PixArt with CLIP text encoder}
        \textbf{(A)} General evaluation of CLIP-DiT-mini model on training and generalized prompt
        templates, showing it is relatively robust to prompt perturbation, comparing to T5-DiT. 
        \textbf{(B)} Attention masking analysis showing CLIP-DiT is robust to attention masking of
        most words (including the relation word), but is most sensitive to \texttt{shape2} and
        slightly to \texttt{shape1}, similar to T5-DiT, suggesting both architectures extract
        spatial information primarily from object tokens.
        \textbf{(C)} Weight-space head screening via QK interaction with the spatial relationship factor in \texttt{shape2} computed from variance partitioning. Projection score identifies two strongly aligned heads: L0H2 and L0H5, positively and negatively aligned with the reference gradient maps. 
        The lower panels display the inner product maps for L0H2 and L0H5 from spatial position embedding towards the spatial relation factors, revealing that these heads translates the textual relation features into aligned spatial attention gradients.
        \textbf{(D)} Visual examples of cross-attention maps to the \texttt{shape2} token (i.e. \texttt{circle}) for L0H2 and L0H5 across denoising generation timesteps (steps 0, 4, 8, 12) for the prompt ``\textit{red square is below and to the right of the blue circle}.'' Both heads progressively localize attention to the spatial region of the target object as denoising proceeds.
        \textbf{(E)} Generated examples following ablation of the identified relation heads (L0H2, L0H5, and combined). Individual head ablation degrades spatial layout, while ablating both heads simultaneously causes severe failure, producing random object relations. 
        \textbf{(F)} Quantitative evaluation of relation head ablation confirms a substantial drop in spatial relationship accuracy, while performance on shape and color
        attributes remains largely intact, demonstrating that L0H2 and L0H5 are specifically
        responsible for encoding spatial relational information in the CLIP-DiT circuit.
    }
     \label{suppfig:CLIP_DiT}
  \end{figure}

  To test the generality of our circuit dissection framework beyond the T5 text encoder, we applied the same analysis pipeline to a PixArt model trained with the CLIP text encoder and tokenizer (CLIP-DiT-mini), which is a popular choice for T2I models (\cref{fig:eval_T2I}). Despite architectural differences in text encoding, we find that the resulting spatial reasoning circuit shares striking structural parallels with T5-DiT. 
  
  The CLIP-DiT-mini model achieves strong performance across prompt templates and is relatively robust to prompt format perturbations (\cref{suppfig:CLIP_DiT}A). Attention masking analysis (\cref{suppfig:CLIP_DiT}B) reveals a highly similar pattern as in T5-DiT: masking most tokens, including the relation word, has minimal effect, while the model is most sensitive to \texttt{shape2} and slightly to \texttt{shape1}; 
  one difference from T5-DiT is that masking attention to \texttt{<end of sentence>} token does not collapse the evaluation metrics, but even improves their performance. 
  This confirms that both architectures extract spatial information primarily from object tokens rather than explicit relation words.
  
  Applying our weight-space head screening method (\cref{suppfig:CLIP_DiT}C), we identify two heads---L0H2 and L0H5---as strongly aligned with spatial relation encoding, with opposite signs. Their inner product maps (lower panels) show that these heads translate textual relation features into complementary spatial attention gradients aligned with each relation direction (above, below, left, right). Visualizing cross-attention maps to the \texttt{shape2} token across denoising steps (\cref{suppfig:CLIP_DiT}D) confirms that both heads progressively localize attention to the target object's spatial region as generation proceeds.
  
  Targeted ablation experiments establish their causal role (\cref{suppfig:CLIP_DiT}E,F): ablating either head individually degrades spatial layout, while ablating both causes severe failures with random object relations. We note that the shape, color and binding accuracy are somewhat also affected by the ablation, but the spatial relation accuracy is severely compromised and degrades towards chance level (1/8). It's less clean than what is found in RTE-DiT.
  
  In summary, these results confirm that our circuit dissection framework generalizes across text encoder architectures and the similar spatial gradient mechanisms in T5-DiT and CLIP-DiT.

  \clearpage
  \subsection{Pre-trained model}
  \label{app_sec:pretrained}
  \begin{figure}[!htp]
    \centering
    \includegraphics[width=0.93\linewidth]{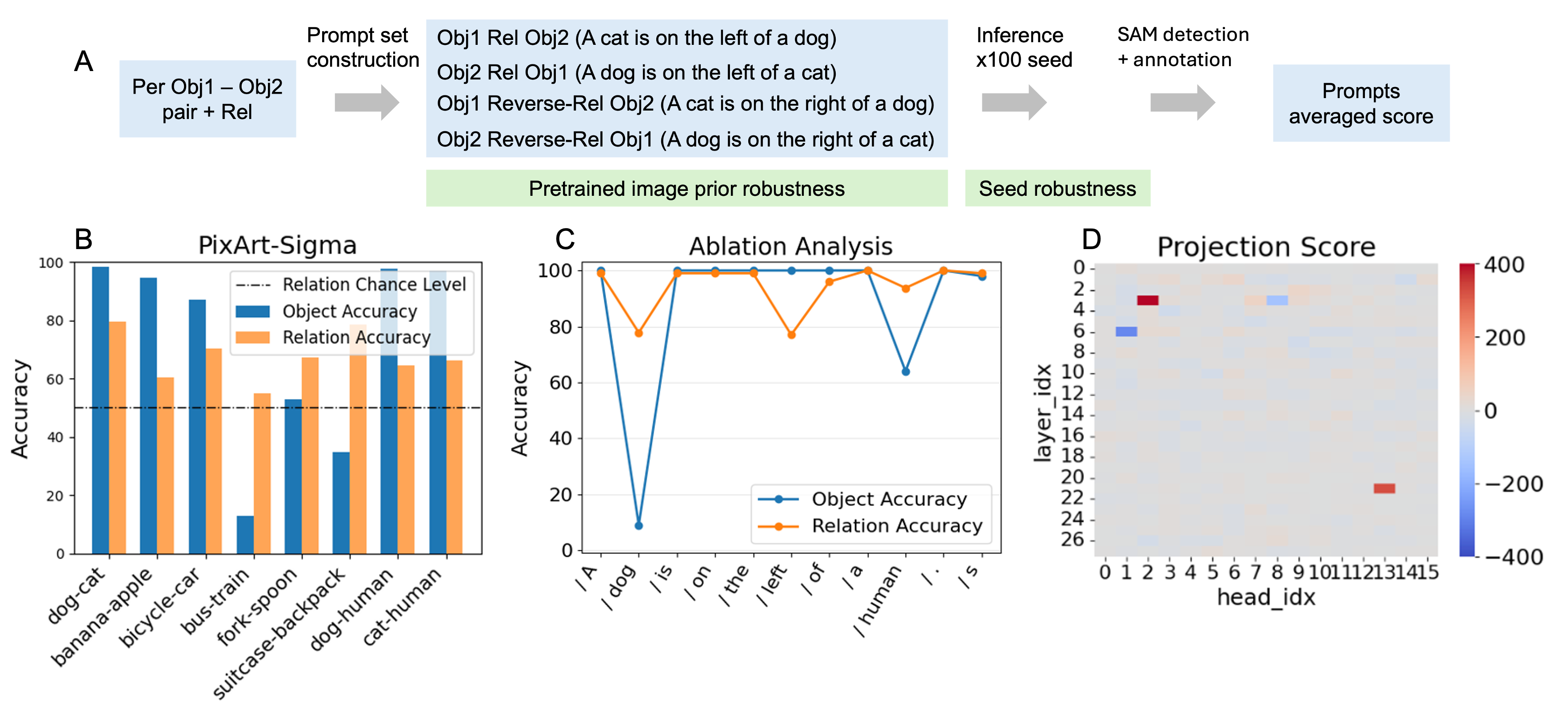}
     \caption{\textbf{Pre-trained T2I models} \textbf{(A)} The prompt set construction and evaluation pipeline. \textbf{(B)} Object and relation accuracy across various object pairs for the PixArt-Sigma model. \textbf{(C)} A text token ablation analysis demonstrating how masking specific tokens affects object and relation accuracy. \textbf{(D)} Projection scores used to identify salient spatial relation heads within the model's layers}
     \label{suppfig:pretrained}
  \end{figure}
  
  To evaluate pre-trained models under real prompts, we applied circuits analysis to the PixArt-Sigma model. We found that despite being the best model in the PixArt family, its spatial accuracy is very low when considering robustness to seed and image priors (e.g., a certain object is highly likely to appear on the left regardless of the relation in the prompt). Among 30 object-pairs, only 8 exhibited nontrivial object and relation accuracy when conditioned on the correct object (Fig.~\ref{suppfig:pretrained}\textbf{B}). We conducted an attention masking analysis on the highest-accuracy prompt, "\textit{A dog is on the left of a human}", by masking each token. This revealed that spatial information is mostly located in the relation token, but it is also distributed across other nouns (Fig.~\ref{suppfig:pretrained}\textbf{C}). Using the quantification method described in App.~\ref{method:head_screening}, we identified three salient heads relevant to spatial relation generation (Fig.~\ref{suppfig:pretrained}\textbf{D}). 
  We believe this method is also applicable to other T2I models such as SDXL, but we leave this for future work due to complications like the U-net architecture and the lack of explicit spatial encoding.
  
\clearpage

\section{Detailed Method}\label{app:method}
\subsection{Object-Relation Dataset Construction}\label{method:dataset}

To study spatial reasoning under controlled conditions, we generate a synthetic dataset of paired images and natural–language captions. Each sample contains two object rendered on a uniform gray background at a resolution of $128\times 128$ pixels. Here we describe the generative process of the image and text pair, basically, we first sample the images and then generate the captions accordingly. For details of generation logic, see the pseudo code below. 

\paragraph{Image Rendering.}
For each image, we randomly sample two \textit{distinct} shapes from \{circle, square, triangle\} with order, and call them shape1 and shape2. Consequently, there are 6 possible combinations of (shape1, shape2). 
Then we place them at uniformly sampled coordinates
\[
(x_i, y_i) \sim \mathcal{U}(r+1,\, 128-r-1), \qquad i=1,2,
\]
where $r=16$ is the radius (or half-side length) controlling object scale. Circles and squares are drawn using their canonical analytic outlines, while triangles are equilateral with side length $2r$. 
To disambiguate the objects, we set them as different colors, one blue and one red. Without loss of generality, the first-selected shape is rendered in \textcolor{red}{red} and the second in \textcolor{blue}{blue}. 
To produce controlled occlusion events, the drawing order is determined by a fair coin flip; the last-drawn object is considered visually ``in front.''

\paragraph{Occlusion Detection.}
We compute axis-aligned bounding boxes $B_1$ and $B_2$ for the two shapes and evaluate their intersection area
\[
A_{\text{overlap}} = \operatorname{area}(B_1 \cap B_2).
\]
If the normalized overlap exceeds a $5\%$ threshold,
\[
\frac{A_{\text{overlap}}}{\min(\operatorname{area}(B_1),\,\operatorname{area}(B_2))} > 0.05,
\]
we label the objects as occluding. In such cases, the spatial relationship is determined by the rendering order: the topmost shape is ``in front of'' the other.

\paragraph{Spatial–Relation Annotation.}
When no significant occlusion occurs, spatial relations are assigned analytically from the relative positions of the two shape centers. 
Using a $5$-pixel tolerance, we assign a spatial relation by comparing the centers $(x_1,y_1)$ and $(x_2,y_2)$ of the two shapes. If $|x_1 - x_2| \le 5$, the relation is classified as \emph{above} or \emph{below} depending on the sign of $y_1 - y_2$. If $|y_1 - y_2| \le 5$, it is classified as \emph{left of} or \emph{right of}. Otherwise, the relation falls into one of the four diagonal categories (\emph{upper-left}, \emph{upper-right}, \emph{lower-left}, \emph{lower-right}), determined jointly by the signs of $(x_1 - x_2)$ and $(y_1 - y_2)$.

\paragraph{Caption Generation.}
For each image, we generate a natural language caption that describes the relationship of the shape1 relative to the shape2 using the geometric and occlusion labels described above. 

If the two shapes are occluding, we select either an ``in front of'' or ``behind'' relation according to the rendering order. A paraphrase variant is then sampled uniformly at random from the corresponding set of templates (e.g., ``in front of,'' ``overlapping and in front of,'' ``behind,'' ``overlapped by'', see Tab.\ref{tab:spatial_paraphrases}).

When no occlusion is present, the captioning module uses the categorical relation (above/below/left/right/diagonal) derived above from $(x_1,y_1)$ and $(x_2,y_2)$ and uniformly maps it to one of several paraphrases (Tab. \ref{tab:spatial_paraphrases}). 
To introduce controlled linguistic variation while preserving semantic structure, we optionally prepend color modifiers to each shape. 
Concretely, the first and second shapes are associated with fixed colors (\textcolor{red}{red} and \textcolor{blue}{blue}), and for each caption we randomly decide whether to include the color adjective or omit it for each shape (50\% chance).

The final caption is produced in a template of the form
\[
\text{``[shape1] is [relation] [shape2]''},
\]
with optional color modifiers and paraphrased relational expressions
\[
\text{``[color1] [shape1] is [relation] [color2] [shape2]''},
\]
. 
Example captions produced by the system include:
\begin{itemize}
    \item \text{``square is in front of triangle.''}
    \item \text{``red triangle is above blue circle.''}
    \item \text{``red circle is overlapped by blue triangle.''}
    \item \text{``red square is to the lower left of blue triangle.''}
    \item \text{``square is left of triangle.''}
    \item \text{``square is to the right of blue triangle.''}
    \item \text{``circle is diagonally up and left from square.''}
    \item \text{``circle is below and to the left of triangle.''}
    \item \text{``red triangle is diagonally down and left from square.''}
    \item \text{``circle is higher than blue triangle.''}
\end{itemize}

\paragraph{Dataset Output.}
For each of the 10{,}000 generated samples, we store:
(i) the rendered RGB image,
(ii) shape identities,
(iii) pixel coordinates,
(iv) occlusion state and depth order,
(v) the categorical spatial–relation label,
and (vi) the corresponding caption.  
This dataset provides a controlled setting for evaluating multi-object scene generation.

\begin{table}[!hb]
\centering
\small
\begin{tabular}{ll}
\toprule
\textbf{Relation} & \textbf{Paraphrase Variants} \\
\midrule
\texttt{upper\_left} 
& to the upper left of; above and to the left of; diagonally up and left from \\
\texttt{upper\_right} 
& to the upper right of; above and to the right of; diagonally up and right from \\
\texttt{lower\_left} 
& to the lower left of; below and to the left of; diagonally down and left from \\
\texttt{lower\_right} 
& to the lower right of; below and to the right of; diagonally down and right from \\
\texttt{above} 
& above; directly above; higher than \\
\texttt{below} 
& below; directly below; lower than \\
\texttt{left} 
& to the left of; left of \\
\texttt{right} 
& to the right of; right of \\
\texttt{in\_front} 
& in front of; overlapping and in front of \\
\texttt{behind} 
& behind; overlapped by \\
\bottomrule
\end{tabular}
\caption{Paraphrase templates for each spatial relation used in caption generation.}
\label{tab:spatial_paraphrases}
\end{table}
\newpage
\begin{RoundedListing}[basicstyle=\fontsize{8}{9}\selectfont\ttfamily\color{atomForeground}]
def generate_sample(canvas_size=128, radius=16):
    """Generate one synthetic example: image, labels, and caption."""
    shapes = ["triangle", "circle", "square"]
    colors = {1: "red", 2: "blue"}  # shape1, shape2
    # 1. Sample two distinct shapes
    shape1, shape2 = sample_two_distinct(shapes)
    # 2. Sample valid positions (inside canvas margins)
    x1, y1 = sample_position(canvas_size, radius)
    x2, y2 = sample_position(canvas_size, radius)
    # 3. Randomize drawing order (decides which shape is visually on top)
    draw_order = random.choice([(shape1, (x1, y1), 1), (shape2, (x2, y2), 2)])
    if draw_order[0] == shape1:
        # draw shape2 then shape1
        draw_shape(shape2, (x2, y2), color=colors[2])
        draw_shape(shape1, (x1, y1), color=colors[1])
        shape1_on_top = True
    else:
        # draw shape1 then shape2
        draw_shape(shape1, (x1, y1), color=colors[1])
        draw_shape(shape2, (x2, y2), color=colors[2])
        shape1_on_top = False

    # 4. Compute occlusion via bounding boxes
    box1 = get_shape_bbox(shape1, (x1, y1), radius)
    box2 = get_shape_bbox(shape2, (x2, y2), radius)
    overlap_ratio = overlap_area(box1, box2) / min(area(box1), area(box2))
    if overlap_ratio > 0.05:
        # Significant occlusion
        if shape1_on_top:
            relation_key = "in_front"
        else:
            relation_key = "behind"
    else:
        # 5. Geometric relation from relative centers
        dx = x1 - x2
        dy = y1 - y2
        tol = 5
        if abs(dx) <= tol:
            relation_key = "above" if dy < 0 else "below"
        elif abs(dy) <= tol:
            relation_key = "left" if dx < 0 else "right"
        else:
            if dx < 0 and dy < 0:
                relation_key = "upper_left"
            elif dx < 0 and dy > 0:
                relation_key = "lower_left"
            elif dx > 0 and dy < 0:
                relation_key = "upper_right"
            else:
                relation_key = "lower_right"
    # 6. Sample natural-language phrase for this relation
    phrase = random.choice(SPATIAL_PHRASES[relation_key])
    # Optionally include or drop color adjectives
    shape1_qual = random.choice([colors[1], ""])
    shape2_qual = random.choice([colors[2], ""])
    # 7. Compose caption: "<shape1> is <relation> <shape2>"
    caption = f"{shape1_qual} {shape1} is {phrase} {shape2_qual} {shape2}".strip()
    # 8. Return image and labels (shape ids, positions, relation, caption, etc.)
    labels = {
        "shape1": shape_to_idx(shape1),
        "shape2": shape_to_idx(shape2),
        "location1": (x1, y1),
        "location2": (x2, y2),
        "spatial_relationship": relation_key,
        "caption": caption,
    }
    return image, labels
\end{RoundedListing}

\subsection{Model Architecture and Training}\label{method:model_arch}

\subsubsection{Text Encoders}
In our main study, we experiment with two kinds of text encoders as conditioning of T2I model: (i) a pretrained T5 encoder and (ii) randomized embedding baselines (with and without positional encodings). 
All setups share a common tokenization. 
In appendix \ref{app_subsec:CLIP_DiT}, we also experimented with CLIP text encoder, which came with a different text tokenizer, and a different encoding dimension $D$. 

\paragraph{Tokenization.}
Captions are tokenized with a T5 tokenizer using a fixed maximum sequence length of $L=20$ tokens. For each caption $c$, we apply padding and truncation to obtain input IDs and attention masks:
\[
\texttt{tokenizer}(c, \texttt{max\_length}=L, \texttt{padding=``max\_length''}, \texttt{truncation=True}).
\]
We additionally construct an ``unconditional'' prompt by tokenizing the empty string $c = ""$, which is used as the null-conditioning input (e.g., for classifier-free guidance). 

\paragraph{Pretrained T5 Encoder.}
The default text encoder is a \texttt{T5EncoderModel} from the pretrained \texttt{t5-v1\_1-xxl} checkpoint. We load the encoder at mixed precision (bfloat16) and treat it as a \textit{frozen} feature extractor. For a tokenized prompt $(\texttt{input\_ids}, \texttt{attention\_mask})$, the encoder produces a sequence of contextual embeddings
$E \in \mathbb{R}^{L \times D}$,
where $D=4096$ is the hidden dimension for T5-XXL. 

\paragraph{Pretrained CLIP Encoder.}
As an alternative text encoder, we use the \texttt{CLIPTextModelWithProjection} from the pretrained \texttt{stabilityai/stable-diffusion-xl-base-1.0} checkpoint (\texttt{text\_encoder\_2} subfolder), which corresponds to the OpenCLIP ViT-bigG architecture. We load the encoder at half precision (float16) and treat it as a \textit{frozen} feature extractor. For a tokenized prompt $(\texttt{input\_ids}, \texttt{attention\_mask})$ with a maximum sequence length of $L=20$ tokens, the encoder produces a sequence of contextual embeddings
$E \in \mathbb{R}^{L \times D}$,
where $D = 1280$ is the hidden dimension of the ViT-bigG text encoder. 

\paragraph{Random Embedding Baselines.}
To disentangle the role of true linguistic structure from that of generic token-level conditioning, we also employ randomized text encoders that replace the T5 encoder with fixed random embeddings. Both variants operate on a learned-free dictionary
\[
\texttt{embedding\_dict} \in \mathbb{R}^{V' \times D},
\]
where $V'$ is the number of distinct token IDs appearing in the caption corpus and $D=4096$ matches the dimensionality of the T5-XXL encoder.

To construct this dictionary, we first gather all unique token IDs from the tokenized captions. Given the token matrix $\texttt{input\_ids\_tsr} \in \mathbb{N}^{N \times L}$, we compute
\[
\texttt{unique\_input\_ids} = \operatorname{Unique}(\texttt{input\_ids\_tsr}),
\]
and retain the inverse mapping needed to reconstruct the full caption tensor. For each unique token ID, we then sample a random embedding vector
\[
e_i \sim \mathcal{N}\!\left(0,\, \frac{7.5^2}{D} I_D\right), \qquad i = 1,\ldots,|\texttt{unique\_input\_ids}|.
\]
The scaling factor $7.5$ is chosen to approximately match the empirical $\ell_2$ norm of T5-XXL embeddings, ensuring that the resulting random encoders operate in a comparable magnitude regime and that the diffusion model receives conditioning vectors of the correct scale.

The resulting matrix
\[
\texttt{embedding\_dict} \in \mathbb{R}^{|\texttt{unique\_input\_ids}| \times D}
\]
is paired with two lookup tables,
\[
\texttt{input\_ids2dict\_ids} : \mathbb{N} \rightarrow \{0,\ldots,V'-1\}, \qquad
\texttt{dict\_ids2input\_ids} : \{0,\ldots,V'-1\} \rightarrow \mathbb{N},
\]
which implement a deterministic one-to-one mapping between tokenizer IDs and dictionary rows. These components are cached and used identically across all random-embedding text encoder variants. 

\textbf{RandomEmbeddingEncoder without Positional Encoding (RTE without pos).}
The first and simplest baseline, replaces the T5 encoder with a simple embedding lookup layer. Given token IDs $x \in \mathbb{N}^{L}$, we map each ID to its dictionary index and retrieve the corresponding embedding:
\[
E_{t} = \texttt{embedding\_dict}[\texttt{input\_ids2dict\_ids}(x_t)], \quad t=1,\ldots,L.
\]
No positional encodings or contextualization are applied; each token is represented by an independent random vector. The resulting embedding tensor has shape $L \times D$ and is used in place of T5 features.

\textbf{RandomEmbeddingEncoder with Positional Encodings (RTE).}
The second baseline, \texttt{RandomEmbeddingEncoder\_wPosEmb}, augments the random embeddings with sinusoidal positional encodings. We construct a standard transformer-style positional encoding
$\texttt{wpe} \in \mathbb{R}^{L \times D}$
using sine and cosine functions with logarithmically spaced frequencies:
\[
\texttt{wpe}_{t,2k} = \sin\left(\frac{t}{10000^{2k/D}}\right), \quad
\texttt{wpe}_{t,2k+1} = \cos\left(\frac{t}{10000^{2k/D}}\right),
\]
for positions $t=0,\ldots,L-1$ and feature indices $k$. The positional encodings are scaled by a factor of $1/6$ and added to the random embeddings:
\[
\tilde{E}_t = E_t + \alpha \cdot \texttt{wpe}_t,\quad \alpha = \tfrac{1}{6}.
\]
This yields position-aware random encoding with the same shape and dtype as the T5 embeddings, enabling a drop-in replacement while preserving the position information for each language token.


Together, these configurations---pretrained T5, pretrained CLIP, random embeddings without position (RTEwP), and random embeddings with sinusoidal positional encodings (RTE)---allow us to probe how much of the text2image generative model’s performance depends on rich linguistic structure versus generic token-level conditioning signals.

\subsubsection{Image Encoders}
Following the original PixArt framework~\cite{chen2023PixArtAlpha}, we use the pretrained variational autoencoder (VAE) from Stable Diffusion~\cite{rombach2022latentdiff} (\texttt{sd-vae-ft-ema}) as our image encoder. The VAE downsamples the spatial resolution by a factor of $8$ in each dimension and produces a 4-channel latent representation. 

\subsubsection{Diffusion Transformer}

We set the patch size of DiT as 2, which results in $8\times8=64$ image tokens per latent. 
We use the \texttt{PixArt} backbone with:
\begin{itemize}
    \item no windowed attention (\texttt{window\_size=0}, \texttt{window\_block\_indexes=[]}),
    \item no relative positional encoding (\texttt{use\_rel\_pos=False}),
    \item full-precision attention layers enabled (\texttt{fp32\_attention=True}),
    \item transformer context length of $20$ (\texttt{model\_max\_length=20}).
\end{itemize}
We train a family of \texttt{PixArt} backbones with different scales, that differ in transformer depth, hidden size, and number of attention heads (all use patch size 2). \texttt{PixArt\_B\_2} uses 12 layers with a 768-dimensional hidden size and 12 heads. \texttt{PixArt\_S\_2} retains the 12-layer depth but reduces the hidden size to 384 with 6 heads. The smaller \texttt{PixArt\_mini\_2} and \texttt{PixArt\_micro\_2} variants use 6 layers with hidden sizes of 384 and 192, respectively, each with 6 or 3 heads. The most compact model, \texttt{PixArt\_nano\_2}, uses 3 layers with a 192-dimensional hidden size and 3 heads. Together, these variants allow us to systematically examine how object–relation learning and circuit solutions scale with model capacity.

\subsubsection{Training}

We train the PixArt model on our synthetic dataset using the configuration shown below. Unless otherwise noted, we follow the standard PixArt training setup with several adjustments to accommodate our text encoders and small image resolution (128×128). The key hyperparameters are summarized below.

\paragraph{Data and Input Format.}
Images are encoded with the Stable Diffusion VAE, producing $4\times16\times16$ latent tensors. 
Captions are tokenized to a maximum length of $20$ tokens, and conditioning embeddings have dimensionality $4096$.\footnote{CLIP-PixArt version has conditioning embedding dimension $1280$.}
The caption embedding and image latents are pre-computed to reduce training overhead.

\paragraph{Optimization.}
The model is trained using AdamW with learning rate $1\times 10^{-4}$, weight decay $3\times 10^{-2}$, and $\epsilon=10^{-10}$. A constant learning-rate schedule with $500$ warmup steps is applied, and gradient clipping is set to $0.01$. 

\paragraph{Training Schedule and Batch Size.}
We train for up to $4000$ epochs with a global batch size of $256$, resulting in 160000 gradient steps. 
No gradient accumulation is used (\texttt{gradient\_accumulation\_steps=1}), and gradient checkpointing is enabled to reduce memory usage. 
During training, samples for a set of validation prompts are periodically generated to check the training status. 

Overall, the most critical hyperparameters for our experiments are the large batch size ($256$), the AdamW configuration (lr $1\mathrm{e}{-4}$, wd $3\mathrm{e}{-2}$), the short context length ($20$ tokens).

\subsection{Sampling}\label{method:sampling}
We generate images using a DPM-solver++\cite{lu2022dpm++} sampler with 14 inference steps for fast generation. We use classifier-free guidance \cite{ho2022classifierFreeGuidance} with a scale of 4.5 to strengthen the conditional signal. 

\subsection{Evaluation}\label{method:evaluation}
We evaluate models on their ability to (i) correctly generate object-level structure and (ii) satisfy a parametric scene description specifying shapes, colors, and spatial relations. To this end, we implement a deterministic analysis pipeline that parses each generated image into object hypotheses and compares them against the ground-truth scene specification.

\paragraph{Object Parsing and Classification.}
Given a generated image, we first convert it to a NumPy array and process each color channel independently using OpenCV. For each channel, we threshold the intensity at 180 to obtain a binary mask, extract connected components via contour detection, and approximate each contour using \texttt{cv2.approxPolyDP}. The resulting polygon is classified as a \emph{triangle}, \emph{square}, or \emph{circle} based on its number of vertices (3, 4, or $>4$, respectively). We discard small components whose area falls below a fixed threshold (100 pixels). For each surviving contour, we compute:
(i) the shape label,
(ii) the bounding box and center coordinates,
(iii) the contour area, and
(iv) the mean RGB color over the contour region, yielding a DataFrame of detected objects.

Color identities (\emph{red} vs.\ \emph{blue}) are derived by thresholding the mean RGB values: an object is tagged as red-dominant if its red channel is near 255 and the green and blue channels are near zero (within a margin of 25), and analogously for blue-dominant. This produces binary indicators \texttt{is\_red} and \texttt{is\_blue} used for subsequent matching.

\paragraph{Canonical Spatial Relation from Detections.}
To recover the spatial relation between two detected objects, we use their center coordinates $(x_1,y_1)$ and $(x_2,y_2)$. We define
$\Delta x = x_1 - x_2$ (positive when object 1 is to the right of object 2) and
$\Delta y = y_1 - y_2$ (positive when object 1 is lower in the image).
Using a $5$-pixel tolerance, we map $(\Delta x,\Delta y)$ to one of eight discrete relations:
\emph{above}, \emph{below}, \emph{left}, \emph{right}, \emph{upper\_left}, \emph{upper\_right}, \emph{lower\_left}, \emph{lower\_right}.
Vertical alignment ($|\Delta x| \le 5$) yields \emph{above} or \emph{below}, horizontal alignment ($|\Delta y| \le 5$) yields \emph{left} or \emph{right}, and all other cases are assigned to one of the four diagonal categories according to the signs of $(\Delta x,\Delta y)$.

\paragraph{Parametric Query Evaluation.}
Each scene is paired with a parametric description
\[
\texttt{scene\_info} = (\texttt{shape1}, \texttt{shape2}, \texttt{color1}, \texttt{color2}, \texttt{spatial\_relationship}),
\]
which encodes the intended relation ``shape1 of color1 is \texttt{spatial\_relationship} shape2 of color2.'' For a given detection DataFrame, we construct two candidate sets:
\begin{itemize}
    \item $O_1$: objects whose shape matches \texttt{shape1} and whose color satisfies \texttt{color1}
    \item $O_2$: analogously for \texttt{shape2} and \texttt{color2}.
\end{itemize}
We then compute the following indicators:
(i) \emph{shape correctness}: whether the two required shapes (if specified) exist in the detection set,
(ii) \emph{color correctness}: whether the two required colors (if specified) exist,
(iii) \emph{existence of a valid binding}: whether $O_1$ and $O_2$ are both non-empty, and
(iv) \emph{uniqueness}: whether there is exactly one object in each set ($|O_1| = |O_2| = 1$).

When a unique binding is present, we extract the centers of the matched objects, compute $(\Delta x,\Delta y)$, and infer the observed spatial relation using the deterministic mapping above. A strict spatial-relation score is obtained by checking whether the observed relation exactly matches \texttt{spatial\_relationship}.

\paragraph{Strict and Loose Spatial Criteria.}
In addition to the strict criterion, we define a looser spatial correctness measure that tolerates deviations within a larger margin and we relax the directional constraint by also accepting diagonally oriented configurations (e.g., above-right) as valid instances of ‘above’ or ‘right’. 
Specifically, we apply relation-specific inequalities on $(\Delta x,\Delta y)$ with threshold (default 8 pixels). For example, an ``above'' relation is considered correct if $\Delta y < -\texttt{threshold}$, while an ``upper\_right'' relation is satisfied if $\Delta x > \texttt{threshold}$ and $\Delta y < -\texttt{threshold}$. This yields a binary \texttt{spatial\_relationship\_loose} flag. 
In main text, the spatial relation accuracy is computed from this \texttt{spatial\_relationship\_loose} flag.

\paragraph{Metrics.}
For each image–scene pair, the evaluation function returns a set of diagnostic booleans:
\texttt{shape}, \texttt{color}, \texttt{exist\_binding}, \texttt{unique\_binding}, \texttt{spatial\_relationship}, and \texttt{spatial\_relationship\_loose}, as well as the overall scores
\[
\texttt{overall} = \texttt{shape} \land \texttt{color} \land \texttt{unique\_binding} \land \texttt{spatial\_relationship},
\]
\[
\texttt{overall\_loose} = \texttt{shape} \land \texttt{color} \land \texttt{unique\_binding} \land \texttt{spatial\_relationship\_loose}.
\]
We additionally record the raw offsets $(\Delta x,\Delta y)$ and the absolute coordinates of the two matched objects, which we use for diagnostic plots and ablations. These metrics allow us to separately quantify failures in object identity, color binding, and spatial configuration, as well as to report a single binary success indicator per prompt.

\subsection{Attention Map Synopsis}\label{method:attention_map_synopsis}

To understand how different PixArt models encode object–relation prompts, we perform a detailed analysis of cross-attention activations during generation. Our procedure instruments the PixArt transformer with custom attention processors that record all self-attention and cross-attention maps at every denoising step. For each prompt, we generate $N=49$ images using $T=14$ inference steps and collect the full tensor
\[
\mathbf{A}^{\text{cross}} \in 
\mathbb{R}^{L \times T \times (2N) \times H \times S \times W},
\]
where $L$ is the number of transformer blocks, $T$ is the number of denoising steps, $2N$ is the total number of samples, $H$ is the number of attention heads, and $S$ and $W$ denote image and text-token dimensions respectively. The first $N$ samples correspond to unconditional run (classifier-free guidance), and the latter $N$ correspond to conditional run; we analyze these separately.

\paragraph{Object-Based Image Masks.}
To relate attention patterns to generated content, we extract object masks for each image using a contour-based CV2 pipeline. Each image is decomposed into binary masks for \textit{square}, \textit{triangle}, \textit{circle}, and \textit{background} regions. Masks are resized to the attention resolution (typically $8 \times 8$) and normalized to obtain an image-token mask
$\mathbf{M}^{\text{img}} \in [0,1]^{N \times S}$. Note this mask differs per samples, since object appears on different locations. 
This allows us to pool attention contributions associated with specific generated objects.

\paragraph{Text-Token Masks.}
For each prompt, we tokenize the caption using the T5 tokenizer and construct multi-hot masks over text tokens for various semantic groups (e.g., \{\texttt{square}\}, \{\texttt{triangle}\}, \{\texttt{above}\}, \{\texttt{left}\}, \{\texttt{red}\}, \{\texttt{blue}\}, or function words \{\texttt{and, to, the, of}\}). This yields text-token masks $\mathbf{M}^{\text{text}} \in \{0,1\}^{W},$
which allow selective measurement of attention into any subset of words.

\paragraph{Template Similarity Scoring.}
Given an image-token mask and text-token mask, we construct an outer-product template
\[
\mathbf{T} = \mathbf{M}^{\text{img}} \otimes \mathbf{M}^{\text{text}} \in \mathbb{R}^{N \times S \times W},
\]
which highlights cross-attention positions connecting a specific visual region to a specific group of words. The cross-attention activation $\mathbf{A}^{\text{cross}}$ is then scored by computing
\[
\mathrm{Score}_{\ell,t,h} = 
\sum_{i,j} 
\mathbf{A}^{\text{cross}}_{\ell,t,h}(i,j)\,\mathbf{T}(i,j),
\]
yielding a tensor of size $L \times T \times H$. These scores are computed separately for the conditional and unconditional branches.

\paragraph{Layer-Head Synopsis and Visualization.}
For each prompt and each (image-mask, text-mask) pair, we summarize the scores by either:
(1) aggregating across inference steps (max or mean), or 
(2) selecting the maximum-activation step per head. Heatmaps over layers and heads visualize where the model most strongly aligns specific visual regions with specific words. We additionally report the top-$k$ heads per prompt, showing which attention heads—and at which layers—encode relations such as “\texttt{square} $\leftrightarrow$ \texttt{left}” or “\texttt{triangle} $\leftrightarrow$ \texttt{above}.”

\paragraph{Prompt Sweep.}
We evaluate all $96$ possible binary spatial-relation prompts (all color/shape combinations crossed with 8 relation types). For each prompt, we generate images, extract masks, compute cross-attention templates, visualize layer–head activation maps, and save summary statistics along with the best-performing heads. This produces a full synopsis of how relation semantics are distributed across layers, heads, and denoising steps for each trained model.

\clearpage
\subsection{Variance Partition and Factorization}\label{method:variance_partitioning}
\paragraph{Intuition.}
Each prompt is described by categorical factors such as the first shape, the spatial relationship, or the color--shape combination of the second object. We ask: \emph{how much of the variance in a representation space is explained by each factor?}

The idea is straightforward. For each factor, we group representations by their factor level (e.g.\ all prompts where shape1 = ``triangle'') and compute group means. If a factor matters, the group means spread apart; if not, they collapse near the grand mean. The ratio of between-group to total variance gives a multivariate $R^2$---the same quantity as in linear regression, applied to high-dimensional vectors rather than scalars.

Since each stimulus is labeled by multiple factors simultaneously, their contributions can overlap. We therefore report two quantities per factor: a \emph{marginal} $R^2$ (variance explained in isolation) and a \emph{partial} $R^2$ (unique variance after accounting for all other factors). The gap between the two reveals shared explanatory power.

Mechanically, this is a multivariate ANOVA: we regress the centered representations onto one-hot factor labels. For non-Euclidean metrics we replace the explicit regression with its kernel-space equivalent (distance-based redundancy analysis), keeping the same projector algebra.

\paragraph{Formulation.}
Let $\{\mathbf{x}_i\}_{i=1}^n$, $\mathbf{x}_i \in \mathbb{R}^d$, denote the representations of $n$ samples (e.g., embedding vectors) arranged in a matrix
\begin{equation}
    X = 
    \begin{bmatrix}
        \mathbf{x}_1^\top \\
        \vdots \\
        \mathbf{x}_n^\top
    \end{bmatrix}
    \in \mathbb{R}^{n \times d}.
\end{equation}
Each sample is annotated by a set of categorical factors (features) $f \in \mathcal{F}$, such as spatial relationship or shape1. For a given factor $f$, we denote the label of stimulus $i$ by $y_i^{(f)} \in \{1,\dots,L_f\}$, where $L_f$ is the number of levels of that factor.

\paragraph{Centered design matrices.}
For each categorical factor $f$ we construct a one-hot design matrix $Z^{(f)} \in \mathbb{R}^{n \times L_f}$,
\begin{equation}
    Z^{(f)}_{i\ell} = 
    \begin{cases}
        1 & \text{if } y_i^{(f)} = \ell, \\
        0 & \text{otherwise,}
    \end{cases}
\end{equation}
and remove the intercept by column-centering
\begin{equation}
    Z^{(f)} \leftarrow Z^{(f)} - \frac{1}{n}\mathbf{1}\mathbf{1}^\top Z^{(f)},
\end{equation}
where $\mathbf{1} \in \mathbb{R}^n$ is the all-ones vector.  
We then concatenate all factors into a single design matrix
\begin{equation}
    Z = \big[\,Z^{(f_1)} \;\big|\; Z^{(f_2)} \;\big|\; \dots \;\big|\; Z^{(f_{|\mathcal{F}|})}\,\big] \in \mathbb{R}^{n \times p}.
\end{equation}

\paragraph{Metric Gram matrix.}
We analyze variance in a (possibly nonlinear) feature space induced by a distance or similarity metric on $X$. Let
\begin{equation}
    J_n = I_n - \frac{1}{n}\mathbf{1}\mathbf{1}^\top
\end{equation}
be the centering matrix. Depending on the metric, we construct a centered Gram matrix $A \in \mathbb{R}^{n \times n}$:

\begin{itemize}
    \item \textbf{Euclidean metric.} For the standard Euclidean geometry, we center $X$ across samples and form
    \begin{equation}
        X_c = J_n X, \qquad
        A = X_c X_c^\top.
    \end{equation}
    This is equivalent to the standard linear kernel on centered features.
    
    
    \item \textbf{General dissimilarity metrics.} For a generic metric $\delta(\mathbf{x}_i,\mathbf{x}_j)$, we form the pairwise distance matrix $D \in \mathbb{R}^{n\times n}$, $D_{ij} = \delta(\mathbf{x}_i,\mathbf{x}_j)$, and its elementwise square $D^{\circ 2}$, then apply classical multidimensional scaling (MDS) centering:
    \begin{equation}
        A = -\frac{1}{2}\, J_n D^{\circ 2} J_n.
    \end{equation}
\end{itemize}

The matrix $A$ can be interpreted as the centered inner-product matrix in an implicit feature space associated with the chosen metric. Its total variance is
\begin{equation}
    SS_{\text{total}} = \operatorname{tr}(A).
\end{equation}

\paragraph{Projection operators and total explained variance.}
Given the full design matrix $Z$, we define the orthogonal projector onto its column space as
\begin{equation}
    P_{\text{all}} = Z\,(Z^\top Z)^+ Z^\top,
\end{equation}
where $(\cdot)^+$ denotes the Moore--Penrose pseudoinverse. In practice we compute $P_{\text{all}}$ via a rank-revealing QR decomposition and discard numerically degenerate directions for robustness.

The residual projector is $P_\perp = I_n - P_{\text{all}}$. We then define
\begin{align}
    SS_{\text{resid}} &= \operatorname{tr}(A P_\perp), \\
    SS_{\text{model}} &= SS_{\text{total}} - SS_{\text{resid}},
\end{align}
and the overall coefficient of determination
\begin{equation}
    R^2_{\text{total}} = \frac{SS_{\text{model}}}{SS_{\text{total}}}.
\end{equation}

\paragraph{Marginal and partial contributions of each factor.}
For each factor $f \in \mathcal{F}$ we construct:

\begin{itemize}
    \item The \emph{marginal} projector
    \begin{equation}
        P_f^{\text{marg}} = Z^{(f)} \big(Z^{(f)\top} Z^{(f)}\big)^+ Z^{(f)\top},
    \end{equation}
    which captures the variance explained by factor $f$ alone.
    
    \item The projector onto all \emph{other} factors,
    \begin{equation}
        Z_{-f} = \big[\,Z^{(g)} : g \in \mathcal{F},\, g \neq f\,\big], \qquad
        P_{-f} = Z_{-f} \big(Z_{-f}^\top Z_{-f}\big)^+ Z_{-f}^\top.
    \end{equation}
    
    \item The \emph{partial} projector of $f$ given the others,
    \begin{equation}
        P_f^{\text{part}} = P_{\text{all}} - P_{-f},
    \end{equation}
    analogous to a partial sum-of-squares in ANCOVA.
\end{itemize}

We then compute the marginal sum of squares, partial sum of squares, and the corresponding $R^2$ values,
\begin{align}
    SS_f^{\text{marg}} &= \operatorname{tr}\!\big(A P_f^{\text{marg}}\big), &
    R_f^{2,\text{marg}} &= \frac{SS_f^{\text{marg}}}{SS_{\text{total}}}, \\
    SS_f^{\text{part}} &= \operatorname{tr}\!\big(A P_f^{\text{part}}\big), &
    R_f^{2,\text{part}} &= \frac{SS_f^{\text{part}}}{SS_{\text{total}}}.
\end{align}
We additionally report the partial eta-squared effect size,
\begin{equation}
    \eta_{p,f}^2 = 
    \frac{SS_f^{\text{part}}}{SS_f^{\text{part}} + SS_{\text{resid}}}.
\end{equation}

\paragraph{Permutation-based significance testing.}
To assess whether the partial contribution of factor $f$ is greater than expected by chance, we perform a label-permutation test. Let $SS_f^{\text{part}}$ be the observed partial sum of squares for factor $f$. For each permutation $\pi$ of $\{1,\dots,n\}$, we:

\begin{enumerate}
    \item Shuffle the labels of factor $f$: $\tilde{y}_i^{(f)} = y_{\pi(i)}^{(f)}$.
    \item Reconstruct the permuted design $Z_\pi^{(f)}$ and the full design $Z_\pi = [Z_\pi^{(f)}, Z_{-f}]$.
    \item Form the permuted projector $P_{\text{all},\pi}$ and compute
    \begin{equation}
        SS_{f,\pi}^{\text{part}} = \operatorname{tr}\!\big(A \big(P_{\text{all},\pi} - P_{-f}\big)\big).
    \end{equation}
\end{enumerate}

Given $N_{\text{perm}}$ permutations, we define the permutation $p$-value as
\begin{equation}
    p_f = \frac{1 + \#\{\pi: SS_{f,\pi}^{\text{part}} \geq SS_f^{\text{part}}\}}{1 + N_{\text{perm}}}.
\end{equation}

\paragraph{Additive effect vectors in Euclidean space.}
When the metric is Euclidean, we can interpret variance partitioning in terms of an explicit linear model on the representation vectors. We write
\begin{equation}
    \mathbf{x}_i = \boldsymbol{\mu} + \sum_{f \in \mathcal{F}} \boldsymbol{\beta}^{(f)}_{y_i^{(f)}} + \boldsymbol{\varepsilon}_i,
    \label{eq:additive-model}
\end{equation}
where $\boldsymbol{\mu} \in \mathbb{R}^d$ is a global intercept, and $\boldsymbol{\beta}^{(f)}_\ell \in \mathbb{R}^d$ is the effect vector for level $\ell$ of factor $f$. To make the decomposition identifiable, we impose a \textit{sum-to-zero constraint per factor},
\begin{equation}\label{eq:varpart_zerosum_constr}
    \sum_{\ell=1}^{L_f} \boldsymbol{\beta}^{(f)}_\ell = \mathbf{0}, \quad \forall f \in \mathcal{F}.
\end{equation}

Let
\begin{equation}
    \boldsymbol{\mu} = \frac{1}{n} \sum_{i=1}^n \mathbf{x}_i, \qquad
    X_c = X - \mathbf{1}\boldsymbol{\mu}^\top
\end{equation}
be the sample mean and the centered representation matrix. The additive model
\eqref{eq:additive-model} can be written in matrix form as
\begin{equation}
    X_c = Z B + E,
\end{equation}
where $B \in \mathbb{R}^{p \times d}$ stacks all effect vectors as rows, and $E$ collects residuals. We estimate $B$ by multivariate least squares
\begin{equation}
    \hat{B} = \arg\min_{B} \lVert X_c - Z B \rVert_F^2
    = (Z^\top Z)^+ Z^\top X_c.
\end{equation}

The rows of $\hat{B}$ corresponding to factor $f$ yield its level-specific effect vectors
$\hat{\boldsymbol{\beta}}^{(f)}_\ell$. For numerical stability and to enforce the sum-to-zero constraint, we re-center these estimates per factor,
\begin{equation}
    \hat{\boldsymbol{\beta}}^{(f)}_\ell \leftarrow 
    \hat{\boldsymbol{\beta}}^{(f)}_\ell
    - \frac{1}{L_f} \sum_{\ell'=1}^{L_f} \hat{\boldsymbol{\beta}}^{(f)}_{\ell'}.
\end{equation}

Together, the variance-partitioning statistics $(R_f^{2,\text{marg}}, R_f^{2,\text{part}}, \eta_{p,f}^2, p_f)$ summarize how much of the representational variance is attributable to each factor, while the effect vectors $\hat{\boldsymbol{\beta}}^{(f)}_\ell$ describe the direction in representation space associated with each level of each factor.

\begin{table}[t]
\centering
\caption{\textbf{Example Variance partitioning results for representational factors of T5 contextual word vector}. 
The model achieves a total explained variance of $R^2_{\text{total}} = 0.7486$.}
\label{tab:varpart}
\begin{tabular}{lcccccccccc}
\toprule
\textbf{Feature} & \textbf{Levels} & \textbf{df$_\text{eff}$} & \textbf{df$_\text{res}$} &
\textbf{SS$_\text{tot}$} & \textbf{SSR$_\text{marg}$} & \textbf{R$^2_\text{marg}$} &
\textbf{SSR$_\text{part}$} & \textbf{R$^2_\text{part}$} & \boldmath$\eta^2_p$ &
$p_\text{perm}$ \\
\midrule
color2shape2          & 6 & 5 & 249 & 4291.0948 & 2479.8288 & 0.5779 & 1884.9433 & 0.4393 & 0.6360 & 0.0099 \\
spatial\_relationship & 8 & 7 & 249 & 4291.0948 &  517.8120 & 0.1207 &  517.8120 & 0.1207 & 0.3243 & 0.0099 \\
shape1                & 3 & 2 & 249 & 4291.0948 &  809.6197 & 0.1887 &  214.7342 & 0.0500 & 0.1660 & 0.0099 \\
\bottomrule
\end{tabular}
\end{table}

\paragraph{General remark}
In our study, we use the variance partitioning both as a statistical testing tool to examine how much linear explainable variance are there in each feature space; and as a feature discovery tool to find the feature vector corresponding to each factor and feature level. 
For the second purpose, we note that, the zero-sum constraint \cref{eq:varpart_zerosum_constr} may lead to bias in finding features: for examples, this constraint will enforce the factor vector for red and blue to be exactly negative to each other, and the eight relation factor vectors to be zero-sum. 
For a general feature space, this doesn't have to be true, but this constraint seems to work to some extent.

In general, we are solving a supervised feature finding problem, whereas commonly used Sparse Auto-encoder (SAE) solves the unsupervised feature finding problem. 
As a next step, we think a more general method the supervised feature finding can be helpful. 

\clearpage
\subsection{Spatial-Relationship Head Screening via Weight-Space Analysis}\label{method:head_screening}

The cross-attention mechanism in diffusion transformers couples text information to spatial image tokens. Each cross-attention head computes attention logits between spatial query tokens and text key tokens. We develop a screening procedure that identifies heads whose \emph{learned} query--key geometry implements a spatial gradient aligned with directional semantics---without generating any images, i.e. a weight space screening method. 

\paragraph{Cross-attention inner-product maps.}
Consider a transformer with $L$ layers and $H$ attention heads per layer, each of dimension $d_h = d_{\mathrm{model}} / H$. At layer $l$, the cross-attention head $h$ has learned projection matrices $W_Q^{(l)} \in \mathbb{R}^{d_{\mathrm{model}} \times d_{\mathrm{model}}}$ and $W_K^{(l)} \in \mathbb{R}^{d_{\mathrm{model}} \times d_{\mathrm{model}}}$, shared across heads, from which we extract the per-head slices
\begin{equation}
    W_{Q,h}^{(l)} = W_Q^{(l)}[\,:\,,\; h d_h : (h{+}1)d_h\,], \qquad
    W_{K,h}^{(l)} = W_K^{(l)}[\,:\,,\; h d_h : (h{+}1)d_h\,].
\end{equation}

The query input to cross-attention is the hidden state of image tokens which contains a fixed 2D sinusoidal positional embedding via residual connection. To first order approximation, we assume the sinusoidal positional embedding dominates the spatial structure of the query. Therefore, we use the positional embedding directly as a proxy for the query input\footnote{A more general version of our analysis could be using the learned spatial encoding, or the decoding vector for each position \citep{fel2025into} as the query.}. Let $\mathbf{p}_s \in \mathbb{R}^{d_{\mathrm{model}}}$ denote the 2D sinusoidal positional embedding for spatial position $s$ on an $S \times S$ grid ($S = $ \texttt{image\_size} / \texttt{patch\_size}), and let
\begin{equation}
    P = \begin{bmatrix} \mathbf{p}_1^\top \\ \vdots \\ \mathbf{p}_{S^2}^\top \end{bmatrix} \in \mathbb{R}^{S^2 \times d_{\mathrm{model}}}
\end{equation}
be the full positional embedding matrix. The projected query vectors for head $h$ at layer $l$ are
\begin{equation}
    Q_h^{(l)} = P \, W_{Q,h}^{(l)} \in \mathbb{R}^{S^2 \times d_h}.
\end{equation}

For the key input, we consider a set of $M$ text-derived feature vectors $\{\mathbf{v}_m\}_{m=1}^M$, $\mathbf{v}_m \in \mathbb{R}^{d_{\mathrm{model}}}$, which are projected through the caption projection MLP $\phi_{\mathrm{cap}}$ and then through the key matrix:
\begin{equation}
    K_h^{(l)} = V \, W_{K,h}^{(l)} \in \mathbb{R}^{M \times d_h},
\end{equation}
where $V = [\mathbf{v}_1, \dots, \mathbf{v}_M]^\top$. The \emph{cross-attention inner-product map} for head $h$ at layer $l$ is the matrix
\begin{equation}
    \Phi_h^{(l)} = Q_h^{(l)} \, K_h^{(l)\top} \in \mathbb{R}^{S^2 \times M},
    \label{eq:inner-product-map}
\end{equation}
whose $(s, m)$-th entry gives the (pre-softmax) attention logit between spatial position $s$ and text feature $m$. For each feature vector $m$, the column $\boldsymbol{\phi}_m = \Phi_h^{(l)}[:,m]$ can be reshaped into an $S \times S$ spatial map, which we denote $\Phi_m \in \mathbb{R}^{S \times S}$.

\paragraph{Choice of text feature vectors.}
The feature vectors $\{\mathbf{v}_m\}$ entering the key projection can be chosen in two ways, depending on the text encoder:

\begin{itemize}
    \item \textbf{Direct word embeddings.} For non-contextual encoders (e.g., random embeddings with positional encoding), we directly use the encoder output for individual spatial-relation words (``above'', ``below'', ``left'', ``right'', etc.) as the feature vectors, after projection through $\phi_{\mathrm{cap}}$.

    \item \textbf{Variance-partitioned effect vectors.} For contextual encoders (e.g., T5), the embedding of a spatial word depends on the surrounding prompt context. We therefore first extract the spatial-relationship effect vectors $\hat{\boldsymbol{\beta}}^{(\mathrm{rel})}_\ell$ from the variance partition procedure (Section~\ref{method:variance_partitioning}). Specifically, we collect the contextual embeddings at the object token positions (shape2 or shape1) across all prompts, project them through $\phi_{\mathrm{cap}}$, and apply variance partitioning with spatial relationship as one of the factors. The resulting relation effect vectors $V_{rel}$ isolate the spatial-relationship features from the full contextual embedding.
    \item More generally, we imagine other ways to discover spatial relation features should be fine, for example Sparse Auto-encoders. Variance partitioning is a linear supervised factorization method to identify such feature vectors; while SAEs are unsupervised ways to find features. 
\end{itemize}

\paragraph{Ramp template alignment.}
Each spatial relationship $m$ has an associated canonical direction vector $\mathbf{d}_m \in \mathbb{R}^2$:
\begin{equation}
\small
    \mathbf{d}_{\text{above}} = \begin{pmatrix} 0 \\ -1 \end{pmatrix}, \quad
    \mathbf{d}_{\text{below}} = \begin{pmatrix} 0 \\ 1 \end{pmatrix}, \quad
    \mathbf{d}_{\text{left}} = \begin{pmatrix} -1 \\ 0 \end{pmatrix}, \quad
    \mathbf{d}_{\text{right}} = \begin{pmatrix} 1 \\ 0 \end{pmatrix},
\end{equation}
and analogously for the four diagonal directions ($\mathbf{d}_{\text{upper\_left}} = (-1, -1)^\top / \sqrt{2}$, etc.).

For direction $\mathbf{d}_m$, we construct a \emph{ramp template} $T_m \in \mathbb{R}^{S \times S}$ on a coordinate grid $\{(x_j, y_k)\}$ with $x_j, y_k \in [-1, 1]$:
\begin{equation}
    T_m(j,k) = d_{m,x} \cdot x_j + d_{m,y} \cdot y_k,
\end{equation}
where $\mathbf{d}_m = (d_{m,x},\, d_{m,y})^\top$ is the unit direction vector. This template is then mean-centered:
\begin{equation}
    \bar{T}_m = T_m - \operatorname{mean}(T_m).
\end{equation}

Given the spatial attention map $\Phi_m \in \mathbb{R}^{S \times S}$ from Eq.~\eqref{eq:inner-product-map}, also mean-centered as $\bar{\Phi}_m = \Phi_m - \operatorname{mean}(\Phi_m)$, we compute three alignment metrics:

\begin{enumerate}
    \item \textbf{Ramp cosine similarity} (scale-free alignment):
    \begin{equation}
        \rho_m^{(l,h)} = \frac{\langle \bar{\Phi}_m,\, \bar{T}_m \rangle_F}{\lVert \bar{\Phi}_m \rVert_F \cdot \lVert \bar{T}_m \rVert_F},
        \label{eq:ramp-cosine}
    \end{equation}
    where $\langle \cdot, \cdot \rangle_F$ denotes the Frobenius inner product. This measures how well the attention map's spatial pattern correlates with the expected linear gradient, independent of scale. Values near $+1$ indicate strong alignment; near $-1$ indicate anti-alignment.

    \item \textbf{Ramp projection} (signed magnitude):
    \begin{equation}
        \pi_m^{(l,h)} = \frac{\langle \bar{\Phi}_m,\, \bar{T}_m \rangle_F}{\lVert \bar{T}_m \rVert_F},
    \end{equation}
    which captures both the alignment direction and the scale of the attention logits.

    \item \textbf{Map energy}:
    \begin{equation}
        E_m^{(l,h)} = \lVert \bar{\Phi}_m \rVert_F,
    \end{equation}
    which quantifies the overall magnitude of spatial variation in the attention logits (irrespective of direction).
\end{enumerate}

\paragraph{Head scoring and ranking.}
For each head $(l, h)$, we average the ramp cosine similarity across all $M$ spatial-relationship directions:
\begin{equation}
    \bar{\rho}^{(l,h)} = \frac{1}{M} \sum_{m=1}^{M} \rho_m^{(l,h)}.
    \label{eq:head-score}
\end{equation}
This yields a score matrix $\bar{\rho} \in \mathbb{R}^{L \times H}$ summarizing each head's average spatial alignment. Heads with high $\bar{\rho}$ consistently produce attention patterns that spatially grade in the direction expected by the corresponding text feature---indicating that their learned $W_Q$--$W_K$ geometry implements directional spatial positioning. Analogous summary matrices are computed for the projection $\bar{\pi}^{(l,h)}$ and energy $\bar{E}^{(l,h)}$.

\paragraph{Interpretation.}
This procedure can be understood as probing the \emph{static} geometry of each cross-attention head: how the fixed positional structure in the queries interacts with semantically meaningful directions in the keys. A head with high ramp alignment has learned key and query projections such that, when a spatial-relationship token (or its factored effect direction) appears in the text conditioning, the resulting attention logits form a monotonic spatial gradient that concentrates attention on the image region indicated by the relationship. Because the analysis operates entirely on model weights and fixed positional embeddings, it is orders of magnitude faster than empirical screening methods that require generating and evaluating images across many prompts. Whether the discovered relation-aligned heads are functionally relevant, however, must be validated through causal ablation experiments.

\paragraph{Algorithm summary.}
The full procedure is summarized in the pseudocode below.

\vspace{-15pt}
\begin{RoundedListing}[basicstyle=\fontsize{8}{9}\selectfont\ttfamily\color{atomForeground}]
def screen_spatial_heads(transformer, text_encoder, tokenizer,
                         direction_map, pos_embed, caption_proj):
    """Screen all cross-attention heads for spatial-relationship alignment.
    Returns three score matrices, each of shape (L, H): rho_bar, pi_bar, E_bar  """

    # --- Stage 1: Obtain text feature vectors V (M x d_model) ---
    if text_encoder.is_contextual:  # e.g. T5, CLIP
        # Encode balanced prompt set; extract embeddings at object token positions
        word_vecs = extract_object_token_embeddings(
            text_encoder, tokenizer, all_prompts)   # (n_prompts, d_encoder)
        word_vecs_proj = caption_proj(word_vecs)     # (n_prompts, d_model)
        # Variance partition: isolate spatial-relation effect vectors
        _, _, effect_vecs, levels, _ = variance_partition_with_effects(
            word_vecs_proj,
            factors={"spatial_relationship": rel_labels,
                     "shape": shape_labels, ...})
        V = effect_vecs["spatial_relationship"]      # (M, d_model)
    else:
        # Non-contextual encoder: encode relation words directly
        word_embeds = text_encoder(["above","below","left","right",...])
        V = caption_proj(word_embeds)                # (M, d_model)

    # --- Stage 2: Precompute ramp templates for each direction ---
    # direction_map: {"above": (0,-1), "below": (0,1), "left": (-1,0), ...}
    xs = linspace(-1, 1, S)  # S = image_size / patch_size
    ys = linspace(-1, 1, S)
    X_grid, Y_grid = meshgrid(xs, ys)
    templates = {}
    for rel_name, (dx, dy) in direction_map.items():
        T = dx * X_grid + dy * Y_grid   # (S, S) linear ramp
        templates[rel_name] = T - T.mean()  # mean-center

    # --- Stage 3: Screen all (layer, head) pairs ---
    rho_bar = zeros(L, H)   # cosine alignment
    pi_bar  = zeros(L, H)   # signed projection
    E_bar   = zeros(L, H)   # map energy
    for layer_idx in range(L):
        block = transformer.blocks[layer_idx]
        for head_idx in range(H):
            h_slice = slice(head_idx * d_h, (head_idx + 1) * d_h)
            # Project positional embeddings through this head's Q
            Q_h = block.cross_attn.to_q(pos_embed)[:, h_slice]  # (S^2, d_h)
            # Project text features through this head's K
            K_h = block.cross_attn.to_k(V)[:, h_slice]          # (M, d_h)
            # Cross-attention logit map: how each position attends to each feature
            Phi = Q_h @ K_h.T                                   # (S^2, M)

            # Score each direction by ramp alignment
            cosines, projections, energies = [], [], []
            for m, rel_name in enumerate(direction_map):
                attn_map = Phi[:, m].reshape(S, S)      # spatial attention map
                attn_centered = attn_map - attn_map.mean()
                T_bar = templates[rel_name]
                # Projection: signed magnitude of alignment with ramp
                proj = dot(attn_centered, T_bar) / norm(T_bar)
                # Energy: overall strength of spatial variation
                energy = norm(attn_centered)
                # Cosine: does this head's map grade in the right direction?
                rho = proj / energy
                cosines.append(rho)
                projections.append(proj)
                energies.append(energy)

            rho_bar[layer_idx, head_idx] = mean(cosines) # mean ramp cosine  (scale-free directional alignment)
            pi_bar[layer_idx, head_idx]  = mean(projections) # mean ramp projection (signed magnitude of alignment)
            E_bar[layer_idx, head_idx]   = mean(energies) # mean map energy   (overall spatial variation strength)

    return rho_bar, pi_bar, E_bar  # each (L, H)
\end{RoundedListing}

\clearpage

\section{Dataset and code availability}
\label{sec:code}
All code, configuration files, and datasets are publicly available at \url{https://github.com/Animadversio/DiT-Relation-Circuits}.

\section{LLM usage}
 
The usage of LLM is limited to research coding, language polishing, method generation, and literature search. We asked an LLM to suggest surface-level rewrites to improve clarity, grammar, and style for author-written passages. Edits were limited to phrasing and organization at the sentence/paragraph level. We also used an LLM to source papers, and produce brief literature summaries for writing references.


\end{document}